# Strong
# Neutrosophic Graphs and
# Subgraph Topological
# Subspaces

**W. B. Vasantha Kandasamy**
**Ilanthenral K**
**Florentin Smarandache**

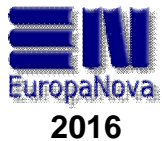

Peer reviewers:

Many books can be downloaded from the following Digital Library of Science:
http://www.gallup.unm.edu/ScienceLibrary.htm

# CONTENTS

# PREFACE

In this book authors for the first time introduce the notion of strong neutrosophic graphs. They are very different from the usual graphs and neutrosophic graphs.

Using these new structures special subgraph topological spaces are defined. Further special lattice graph of subgraphs of these graphs are defined and described.

Several interesting properties using subgraphs of a strong neutrosophic graph are obtained.

Several open conjectures are proposed. These new class of strong neutrosophic graphs will certainly find applications in NCMs, NRMs and NREs with appropriate modifications.

These new notions are interesting and researchers can find lots of applications where neutrosophic graphs find their applications. Apart from some open conjectures several problems at research level are also suggested for the readers.

We wish to acknowledge Dr. K Kandasamy for his sustained support and encouragement in the writing of this book.

W.B.VASANTHA KANDASAMY
ILANTHENRAL K
FLORENTIN SMARANDACHE

**Chapter One**

# INTRODUCTION

In this chapter we just indicate the books which are used for the basic notions used in this book. The concept of neutrosophy can be had from [6-8]. For the basic properties about neutrosophic graphs please refer [59].

The basics of graph theory can be had from [1]. The main notion dealt here is the notion of subset vertex graphs. We have defined in chapter II, two types of subset vertex graphs. However for a given graph G there exists one and only one special subset vertex graph of type I but there are many special subset vertex graphs of type II for a given the set of vertices. By this method one gets several subset vertex graphs of type II. This is elaborately discussed in chapter two of this book. Several nice properties associated with them are defined, described and developed in this chapter. Next for the first time we introduce the notion of strong neutrosophic graphs.

Infact one can say the net working of the brain is more close to strong neutrosophic graphs only. The concept of neutrosophic graphs can be had in [59] and so on. For a systematic analysis and study of neutrosophic graphs one can refer [59]. However strong neutrosophic graphs and specialty associated has been systematically studied in chapter III of this book.

Also associated with these graphs we define the special concept of neutrosophic complement and so on and discuss them in this chapter. Certainly we will use these new notions in several applications. These notions are very unique and will find lots of practical applications.

We prove several interesting features associated with them. Infact for a given set of n-vertices we have several strong neutrosophic graphs. We see even for a graph with two vertices we have the following graphs.

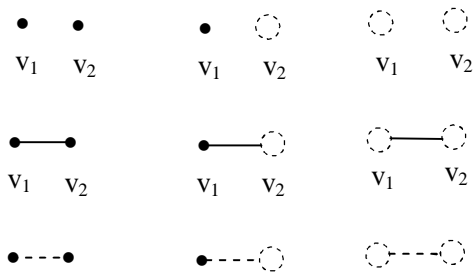

We have infact 9 such strong neutrosophic graphs with just two vertices. Thus this new notions gives one with abundant choices. The dotted circle and dotted lines denote the graphs with indeterminate vertices and edges respectively.

The fourth chapter is very different and innovative. For we get the collection of subgraphs of a graph G denote it by S(G) which includes also the empty graph ϕ we define two operation ∪ and ∩ and show S(G) is a topological space. Further we see these S(G) power set subgraph of the graph G always contains as a substructure a Boolean algebra of appropriate order.

Further we see if the neutrosophic graph is disjoint then the adjacency matrix associated with it will be a super diagonal matrix. Several interesting properties associated with them are discussed.

**Chapter Two**

# SPECIAL SUBSET VERTEX GRAPHS

In this chapter authors for the first time introduce the notion of subset vertex graphs. Subset vertex graphs are of two types. One given a graph G with V = {$v_1$, …, $v_n$} vertex sets obtain all the subsets of this vertex set V and only get edges which are in the graph G.

This will be known as type I subset vertex graph. Type II subset vertex graph is a graph obtained by collecting all subsets of graphs and obtaining graphs using these subsets of vertices. Type I graph of a graph G is unique.

However for type II vertex subset graph one gets many graphs for the given set of vertices.

First we will illustrate these situations by some examples.

***Example 2.1:*** Let G be the graph with only one vertex $v_1$ and zero edges.

We see {$v_1$} is a subset and {$v_1$} is also a point subset graph, $\bullet$ $v_1$.

When $V = \{v_1\}$ we have both the graph and the subset vertex graph are identical as subset vertex graph is also $\left\{ \begin{array}{c} \bullet \\ v_1 \end{array} \right\}$ .

It is interesting to note that in case the number of vertices is one both subset vertex graph of type I and type II are the same.

***Example 2.2:*** Let $G = \{\{v_1, v_2\} = V, e_1\}$ be the graph given in the following

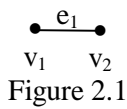

$$\begin{array}{cc} \underset{v_1}{\bullet} \overset{e_1}{\rule{1cm}{0.4pt}} \underset{v_2}{\bullet} \end{array}$$

Figure 2.1

The subset vertices of are $\{v_1, v_2\}$, $\{v_1\}$ and $\{v_2\}$. The subset vertex graph of G is as follows:

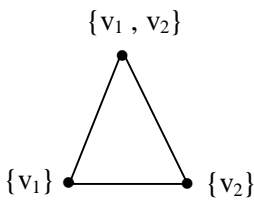

Figure 2.2

$G_v$ is the type I vertex subset graph of G.

Suppose G is just the graph $\underset{v_1}{\bullet}\ \underset{v_2}{\bullet}$

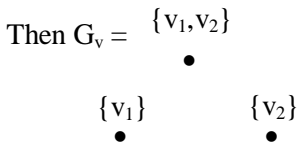

Figure 2.3

is the vertex subset type I graph.

On the other hand suppose we have $V = \{v_1, v_2\}$ a vertex set of order two and no graph is given.

To find all vertex subset graphs using V.

The subsets of V are $\{\{v_1, v_2\}, \{v_1\}, \{v_2\}\} = $ SP (V).

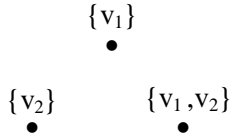

Figure 2.4

$G_1^v$ is a type II vertex subset graph.

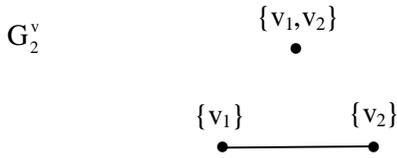

Figure 2.5

is another  type II vertex subset graph.

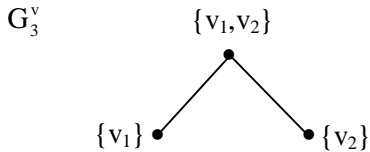

Figure 2.6

is a type II vertex subset graph.

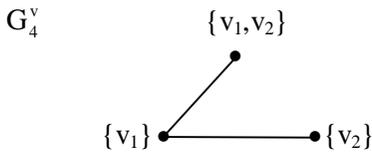

Figure 2.7

is a type II vertex subset graph.

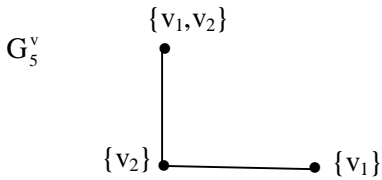

Figure 2.8

is a type II vertex subset graph.

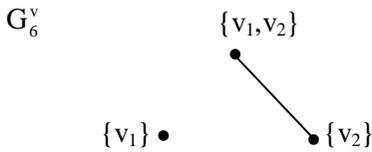

Figure 2.9

is a type II vertex subset graph.

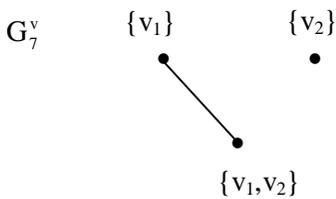

Figure 2.10

is a type II vertex subset graph.

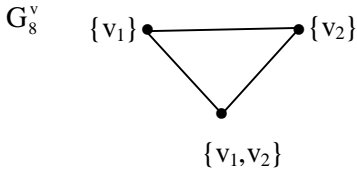

$$G_8^v$$

Figure 2.11

is a type II vertex subset graph.

Thus when no basic graph is given and we have only two vertices we can get eight different vertex subset type II graphs.

We now make the following official definition.

**DEFINITION 2.1**: *Let G be a graph with n vertices and m edges. Then the vertex subset graph of type I; $G_v$ will have $2^n - 1$ number of subset vertices and more than m-edges.*

Thus $G_v$ will denote all subset vertex graphs of type I.

We will first illustrate this situation by some examples.

***Example 2.3:*** Let G be the graph given in the following.

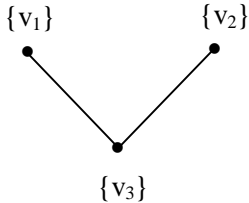

Figure 2.12

The subset vertex of G is

$\{\{v_1\}, \{v_2\}, \{v_3\}, \{v_1, v_2\}, \{v_1, v_3\}, \{v_2, v_3\}, \{v_1, v_2, v_3\}\}$.

The subset vertex graph $G_v$ of G of type I is as follows:

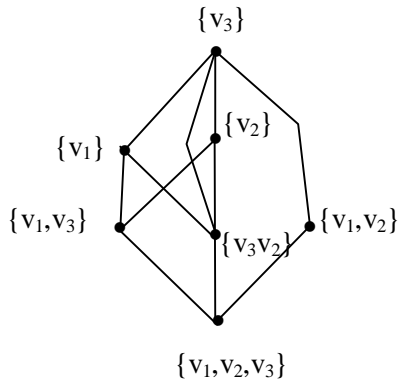

Figure 2.13

We have $G_v$ the one and only subset vertex graph of type I.

Now we proceed onto define the notion of vertex subset graphs $G_i$ of type II, $i \in N$.

However if we are not given any graph but only the three vertices $\{v_1, v_2, v_3\}$. Using the subsets of the vertex subsets we have the following graphs $G^v$ of type II.

**DEFINITION 2.2:** *Let V = {v₁, v₂, …, vₙ} be n vertices where the graph is not given. Clearly using V we have $2^n - 1$ number of subsets. Using these $2^n - 1$ subsets we can draw the graphs $G^v$ of type II defined as subset graphs of type II.*

When n = 1 we have only one subset vertex graph of type II. When n = 2 we have 8 subset vertex graphs of type II given by $G_1^v$, $G_2^v$, …, $G_8^v$.

Infact we leave it as a open conjecture to find the number of subset vertex graphs $G^v$ of type II when we have n vertices.

We see the number of subset vertex graphs $G^v$ of type II.

When n = 3,
$V = \{\{v_1\}, \{v_2\}, \{v_3\}, \{v_1, v_2\}, \{v_1, v_3\}, \{v_2, v_3\}, \{v_1, v_2, v_3\}\}$.

The graphs of subset vertex graphs of type II are very many even if o(V) = 3.

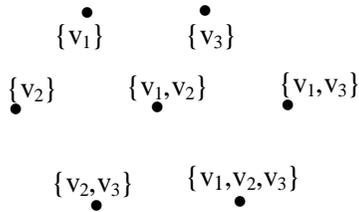

Figure 2.14

is a vertex subset graph $G_1^v$ with no edges.

We have 21 number of vertex subset graphs with only one edge.

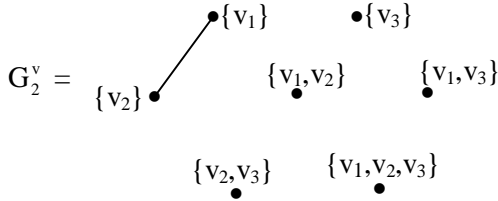

Figure 2.15

and so on.

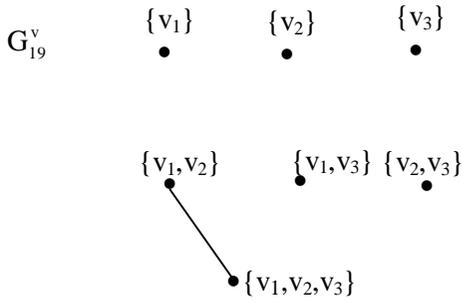

Figure 2.16

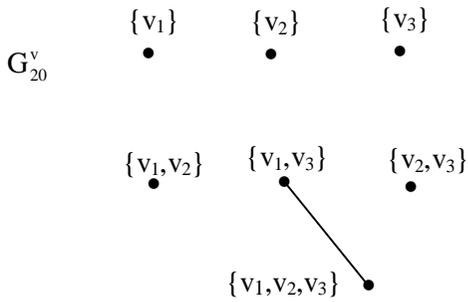

Figure 2.17

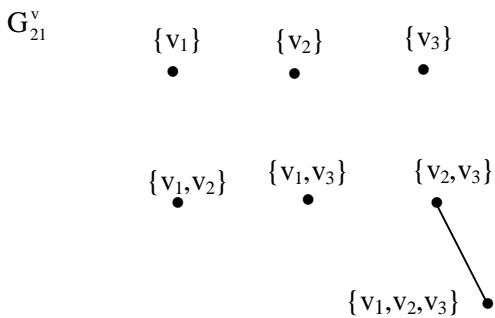

Figure 2.18

There are how many subset vertex graphs of type II?    This a open conjecture.

Number of subset vertex graphs of type II having two edges with only one edge adjacent to each vertex.

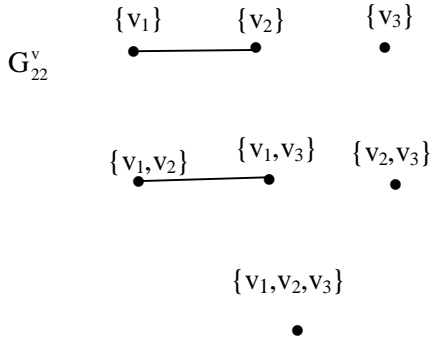

Figure 2.19

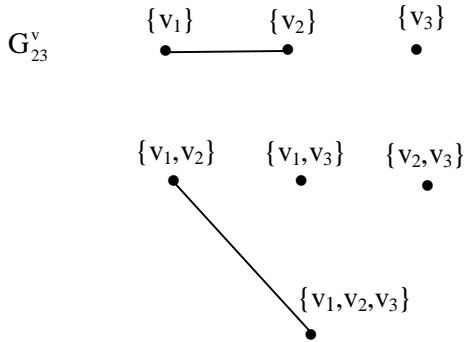

Figure 2.20

and so on.

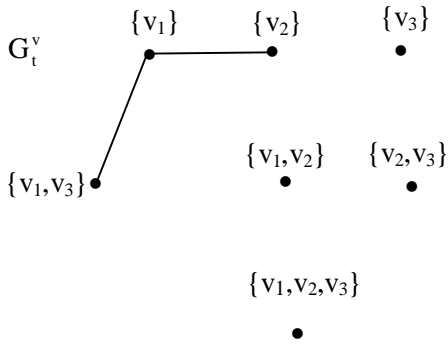

Figure 2.21

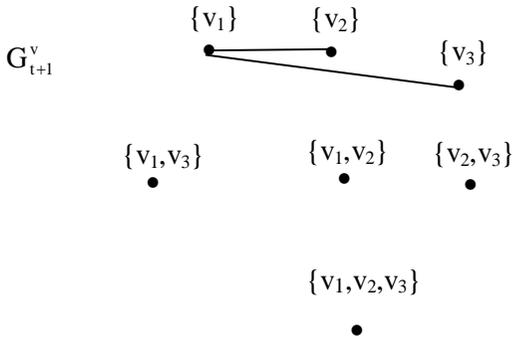

Figure 2.22

and so on.

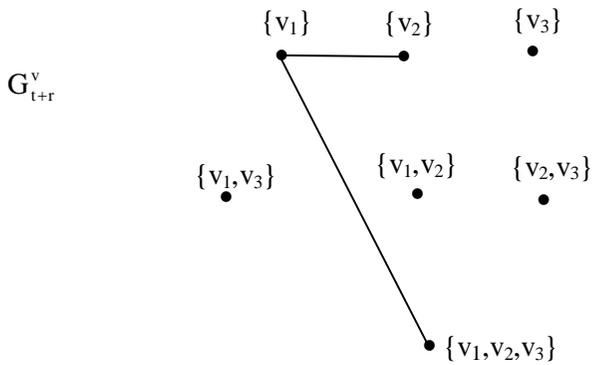

Figure 2.23

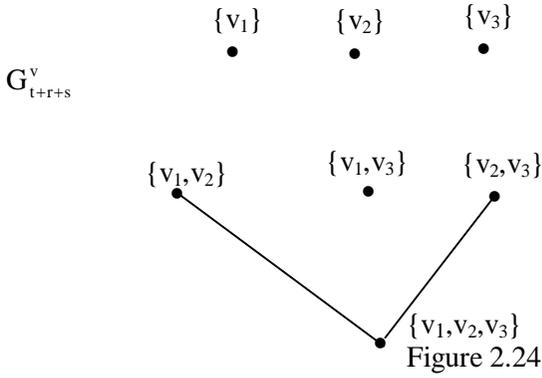

Figure 2.24

Next we can find 3 edges graphs using the seven vertices.

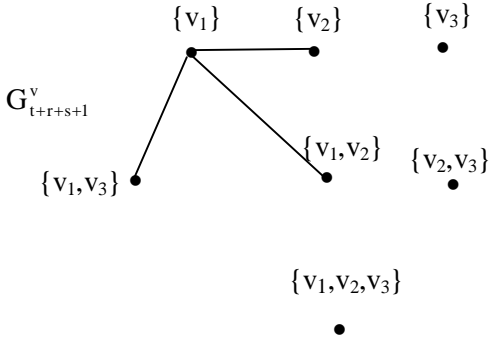

Figure 2.25

and so on.

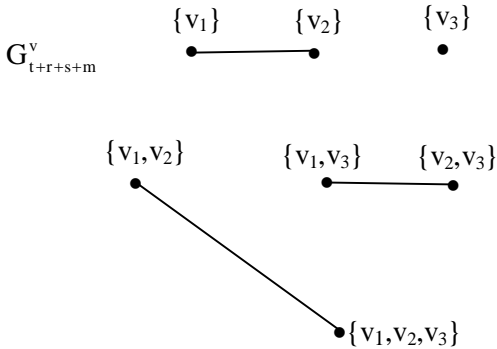

Figure 2.26

and so on.

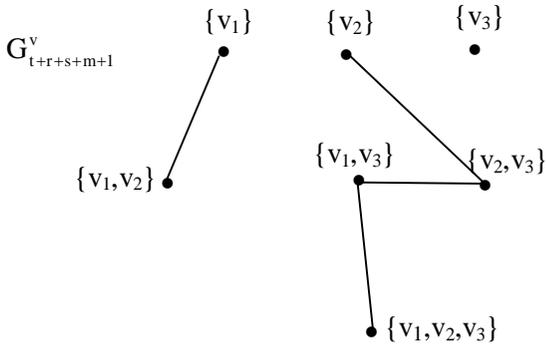

Figure 2.27

and so on.

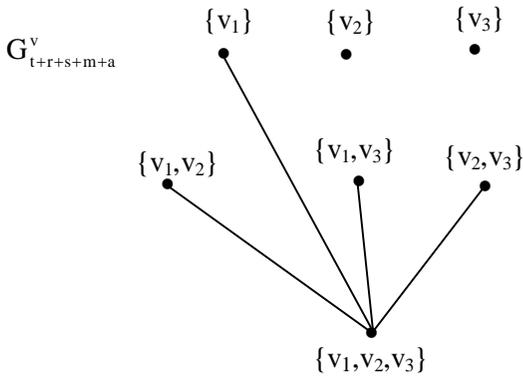

Figure 2.28

and so on.

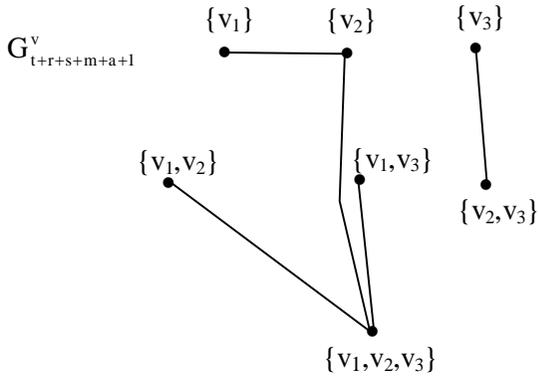

Figure 2.29

and so on.

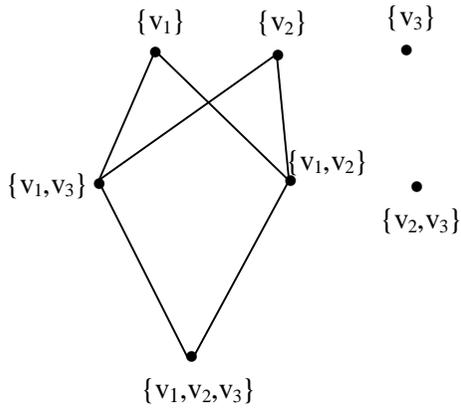

Figure 2.30

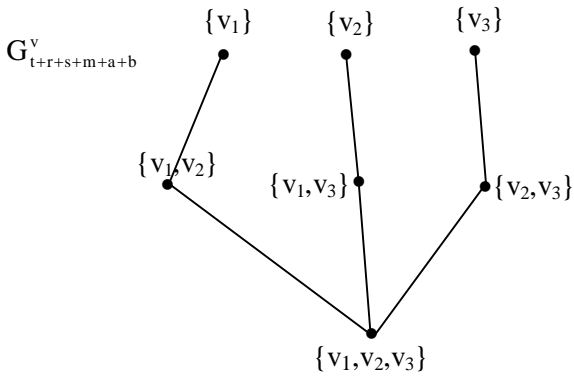

Figure 2.31

and so on.

We can have 7 edges, 8 edges, 9 edges and so on.

Now this study leads to the following conjectures.

**Conjecture 2.1:** Let $V = \{v_1, v_2, v_3\}$ be the set of vertices.

$SP(V) = \{\{v_1\}, \{v_2\}, \{v_3\}, \{v_1, v_2\}, \{v_1, v_3\}, \{v_2, v_3\}, \{v_1, v_2, v_3\}\}$ be the vertex subsets of type II.

How many distinct subset vertex graphs of type II can be constructed using $SP(V)$?

**Conjecture 2.2:** Let $V = \{v_1, v_2, \ldots, v_n\}$ be the subset vertex graph of type II.

How many distinct subset vertex graphs of type II can be constructed using $SP(V)$?

**Conjecture 2.3:** How many one edge subset vertex graphs of type II can be constructed using $2^n - 1$ subset vertices.

**Conjecture 2.4:** How many two edge subset vertex graphs of type II can be constructed using $2^n - 1$ subset vertices?

**Conjecture 2.5:** How many three edge subset vertex graphs of type II can be got using $2^n - 1$ subset vertices?

**Conjecture 2.6:** How many r edge subset vertex graphs of type II can be got using these $2^n - 1$ subset vertices.

**Conjecture 2.7:** Give the maximum number of edges that can be had for the subset vertex graphs of type II using $2^n - 1$ subset vertices.

***Example 2.4:*** Let $V = \{v_1, v_2, v_3\}$ be the vertices.

SP(V) = $\{\{v_1\}, \{v_2\}, \{v_3\}, \{v_1, v_2\}, \{v_1, v_3\}, \{v_1, v_2, v_3\}\}$ be the subset of vertices.

To find the maximum number of edges the subset vertex graph of type II can be got from

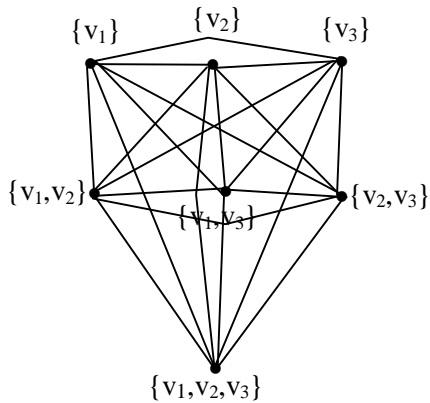

Figure 2.32

We can have several types.

However we can say the maximum number of edges adjacent to any of the subset vertex of a n number of vertices.

$V = \{v_1, v_2, \ldots, v_n\}$ have $2^n - 2$ edges for a subset vertex graph of type II. $SP(V) = 2^n - 1$.

The natural question is; can we have meaningful subset vertex graph type II trees?

For this first we have to fix the root. We always as a convention fix the whole set $V = \{v_1, v_2, \ldots, v_n\}$ as the root.

We will represent this by the following examples.

***Example 2.5:*** Let $V = \{v_1, v_2, v_3\}$ be the vertices.

$SP(V) = \{\{v_1\}, \{v_2\}, \{v_3\}, \{v_1, v_2\}, \{v_1, v_3\}, \{v_2, v_3\}, \{v_1, v_2, v_3\}\}$ be the subset vertices of V.

There can be many subset vertex trees of type II which will be described in the following.

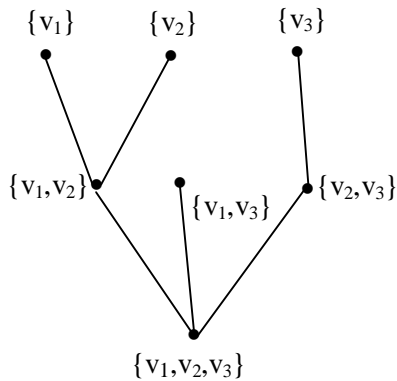

Figure 2.33

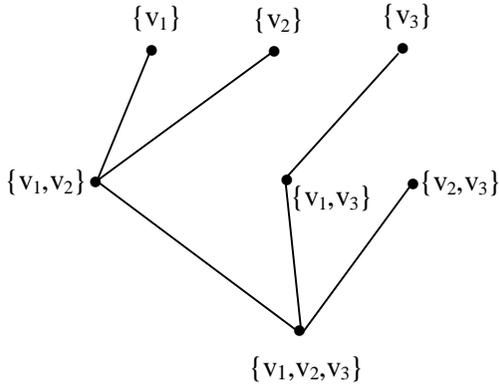

Figure 2.34

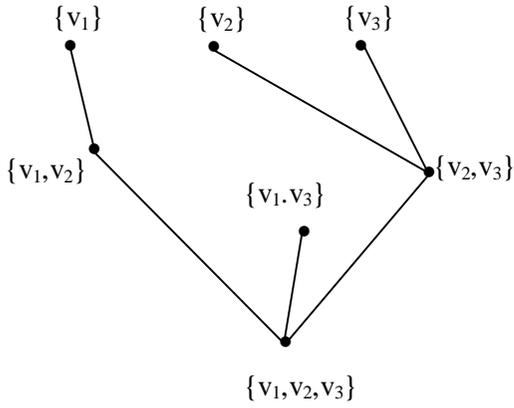

Figure 2.35

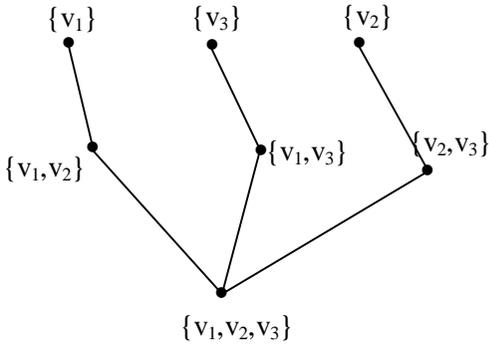

Figure 2.36

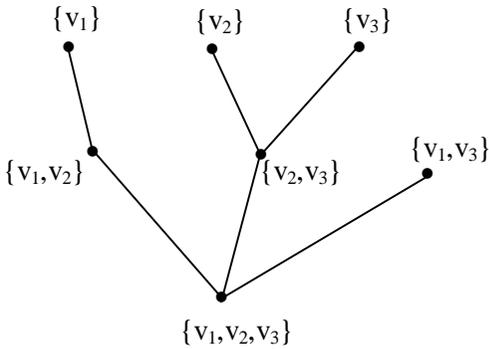

Figure 2.37

How many distinct meaningful subset vertex trees of type II can be constructed using

SP(V) = {{$v_1$}, {$v_2$}, {$v_3$}, {$v_1$,$v_2$}, {$v_1$, $v_3$}, {$v_2$, $v_3$}, {$v_1$, $v_2$, $v_3$}} where o(V) = 3?

If $V = \{v_1, v_2, \ldots, v_n\}$ find the number of subset vertex trees of type II using SP(V).

Using SP(V) of $V = \{v_1, v_2, \ldots, v_n\}$, how many complete subset vertex graph of type II can be obtained?

Next we proceed onto find the subset vertex graph of type I of a graph G which is a tree.

***Example 2.6:*** Let G be the tree graph given in the following.

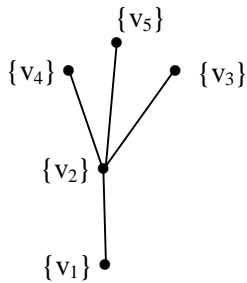

Figure 2.38

$V = \{v_1, v_2, v_3, v_4, v_5\}$ be the vertices of the tree.

$S(P(V)) = \{\{v_1\}, \{v_2\}, \{v_3\}, \{v_4\}, \{v_5\}, \{v_1, v_2\}, \{v_1, v_3\}, \{v_1, v_4\}, \{v_1, v_5\}, \{v_2, v_3\}, \{v_2, v_4\}, \{v_2, v_5\}, \{v_3, v_4\}, \{v_3, v_5\}, \{v_4, v_5\}, \{v_1, v_2, v_3\}, \{v_1, v_2, v_4\}, \{v_1, v_2, v_5\}, \{v_1, v_3, v_5\}, \{v_1, v_3, v_4\}, \{v_1, v_4, v_5\}, \{v_2, v_3, v_4\}, \{v_2, v_3, v_5\}, \{v_2, v_4, v_5\}, \{v_3, v_4, v_5\}, \{v_1, v_2, v_3, v_4\}, \{v_1, v_2, v_3, v_5\}, \{v_1, v_2, v_4, v_5\}, \{v_1, v_3, v_4, v_5\}, \{v_2, v_3, v_4, v_5\}, \{v_1, v_2, v_3, v_4, v_5\}\}$.

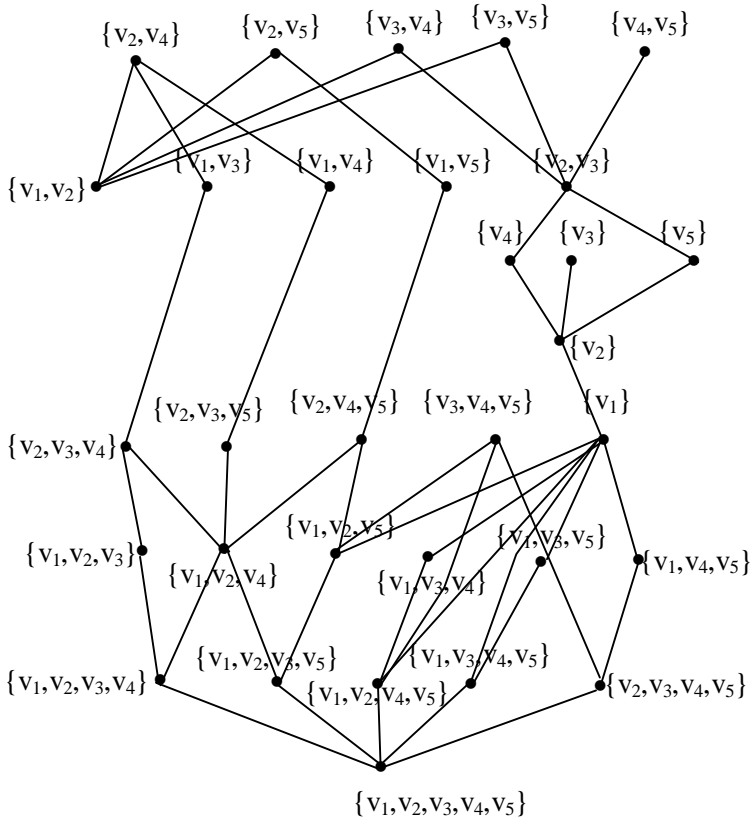

Figure 2.39

However the graph is not complete.

But infact it is very complicated.

However the subset vertex graph $G_v$ of type I is not a subset vertex graph tree even if the given graph G is a tree.

***Example 2.7:*** Let G be the tree

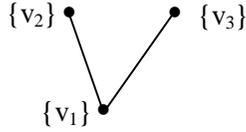

Figure 2.40

V = {$v_1$, $v_2$, $v_3$}, S(P(V)) = {{$v_1$}, {$v_2$}, {$v_3$}, {$v_1$, $v_2$}, {$v_1$, $v_3$}, {$v_2$, $v_3$}, {$v_1$, $v_2$, $v_3$}}.

We find the vertex subset graph (tree) of type I. $G_v$ is as follows:

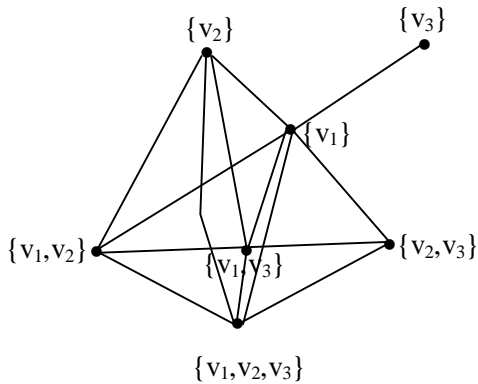

Figure 2.41

Clearly this not a subset graph tree of type I.

It is observed that in general a tree does not lead to a subset vertex tree of type I.

However one finds it difficult to get trees using subset vertices leading to graph $G_v$ of type I.

But if $\{v_1, v_2, \ldots, v_n\}$ are n vertices one can easily get many subset vertex tree of type II; $G^v$.

This is the main advantage of using type II subset vertex graph $G^v$.

Let $V = \{v_1, v_2, v_3\}$ be the vertex set.

$S(P(V)) = \{\{v_1\}, \{v_2\}, \{v_3\}, \{v_1, v_2\}, \{v_1, v_3\}, \{v_2, v_3\}, \{v_1, v_2, v_3\}\}$.

The following graph $G_1^v$ is a tree of type II.

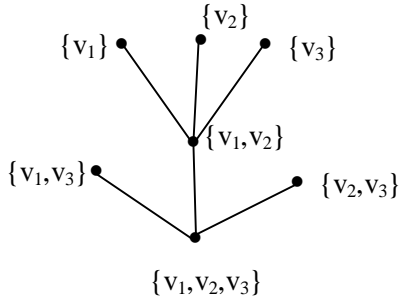

Figure 2.42

Already different graphs of type II tree are discussed.

Next we proceed onto use the set of vertices.

$V = \{v_1, v_2, v_3, v_4\}$.

$S(P(V)) = \{\{v_1\}, \{v_2\}, \{v_3\}, \{v_4\}, \{v_1, v_2\}, \{v_1, v_3\}, \{v_1, v_4\}, \{v_2, v_3\}, \{v_2, v_4\}, \{v_3, v_4\}, \{v_1, v_2, v_3\}, \{v_1, v_2, v_4\}, \{v_1, v_3, v_4\}, \{v_2, v_3, v_4\}, \{v_1, v_2, v_3, v_4\}\}$.

The following is one of the subset vertex tree of type II.

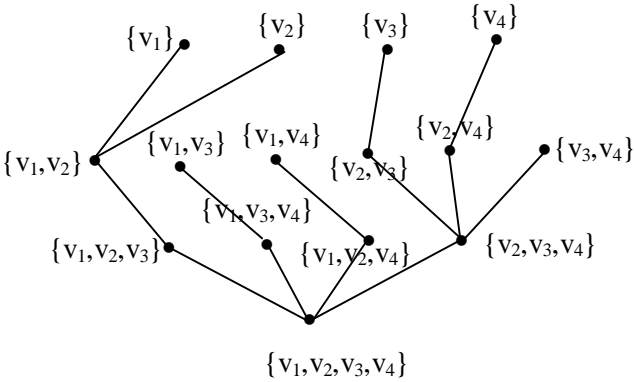

Figure 2.43

This is again another example of the subset vertex graph of type II which is a tree is as follows:

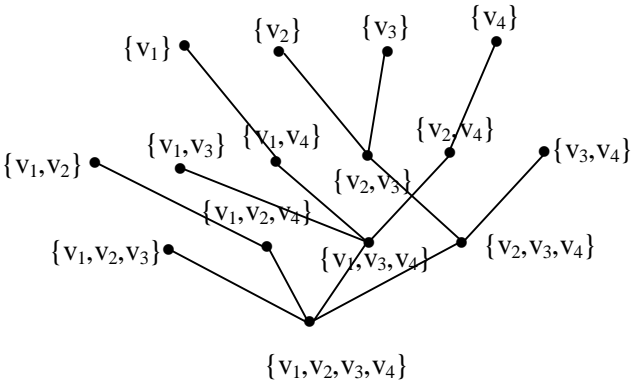

Figure 2.44

Next we proceed onto express a subset vertex graph of type II which is a tree is as follows using the set of vertices.

$V = \{v_1, v_2, v_3, v_4, v_5\}$, $SP(V)) = \{\{v_1\}, \{v_2\}, \{v_3\}, \{v_4\}, \{v_5\}, \{v_1,v_5\}, \{v_1,v_2\}, \{v_1,v_3\}, \{v_1,v_4\}, \{v_2,v_3\}, \{v_2,v_4\}, \{v_2,v_5\},$

$\{v_3,v_4\}$, $\{v_3,v_5\}$, $\{v_4,v_5\}$, $\{v_1,v_2,v_3\}$,$\{v_1,v_2,v_4\}$, $\{v_1,v_2,v_5\}$, $\{v_1,v_3,v_4\}$, $\{v_1,v_5,v_3\}$, $\{v_1,v_4,v_5\}$, $\{v_2,v_3,v_4\}$, $\{v_2,v_3,v_5\}$, $\{v_2,v_4,v_5\}$, $\{v_3,v_4,v_5\}$, $\{v_1,v_2,v_3, v_4\}$, $\{v_1,v_2,v_3, v_5\}$, $\{v_1,v_2,v_4, v_5\}$, $\{v_1,v_3,v_4, v_5\}$, $\{v_2,v_3,v_4, v_5\}$, $\{v_1,v_2,v_3,v_4, v_5\}\}$ is a subset vertex set.

We give the subset vertex graph of type II which is tree is as follows:

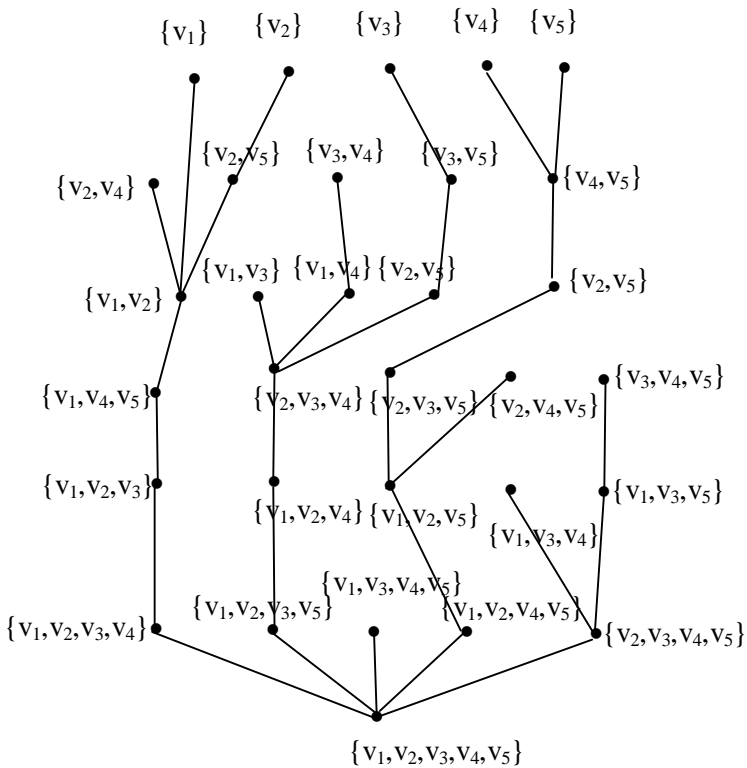

Figure 2.45

This is tree of type II.

Infact we can get several such trees of type II.

So for practical problems finding any suitable tree depending on the problem is at the choice of the researcher. For instance even preferable subsets can be selected from SP(V).

This is illustrated by the following example.

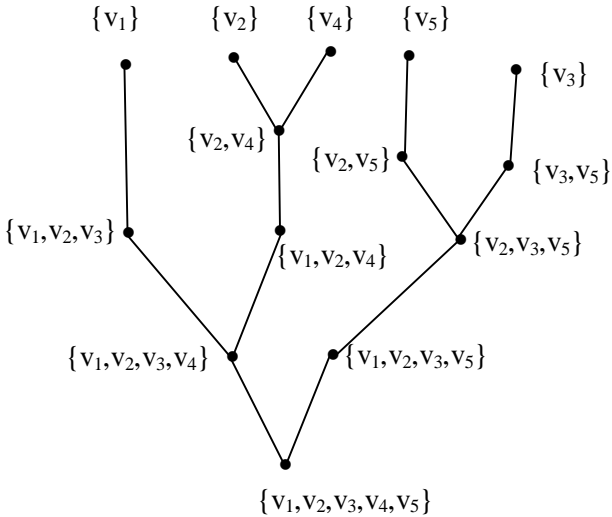

Figure 2.46

This can certainly be used in data mining.

For from a full collection a restricted collection can be taken and the goal is reached and getting the subset vertex tree. This is the way essential subset vertex tree of type II is carried out.

However one of the factors is can we the notion of merged graphs in case of subset vertex graphs of both type I and type II.

Here we do merging not of the whole subset vertex graph but only subgraphs of the graphs.

To this end we first proceed onto study the subgraphs of subset vertex graphs of type I and type II.

It is infact important to keep on record that these subgraphs may be just graphs in some cases and in most cases subset vertex subgraphs of type I or type II according as $G_v$ or $G^v$ is taken.

First we will illustrate this situation by some examples.

***Example 2.8:*** Let $G_v$ be the vertex subset graph of type I of the graph G given by the following figure.

G =

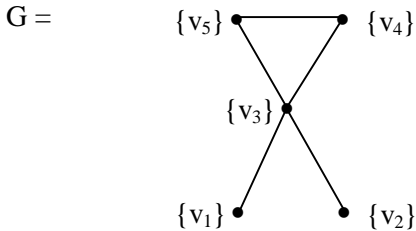

Figure 2.47

Let $S(P(V)) = \{\{v_1\}, \{v_2\}, \{v_3\}, \{v_4\}, \{v_5\}, \{v_1,v_2\}, \{v_1,v_3\}, \ldots, \{v_4,v_5\}, \{v_1,v_2,v_3\}, \{v_1,v_2,v_4\}, \ldots, \{v_3,v_4,v_5\}, \{v_1,v_2,v_3,v_4\}, \{v_1,v_2,v_3,v_5\}, \ldots, \{v_2,v_3,v_4,v_5\}, \{v_1,v_2,v_3,v_4, v_5\}\}$.

Clearly $o(S(P(V)) = 2^5 - 1$.

We see we can have several such vertex subset graphs of type I using the subset vertex set S(P(V)).

We give some of the subset vertex subgraphs of $G_v$.

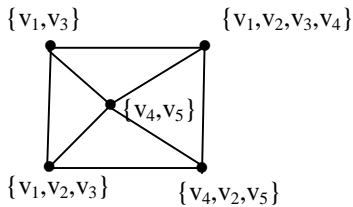

Figure 2.48

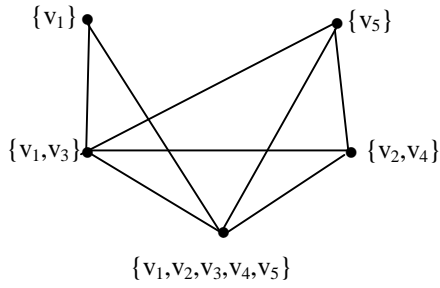

Figure 2.49

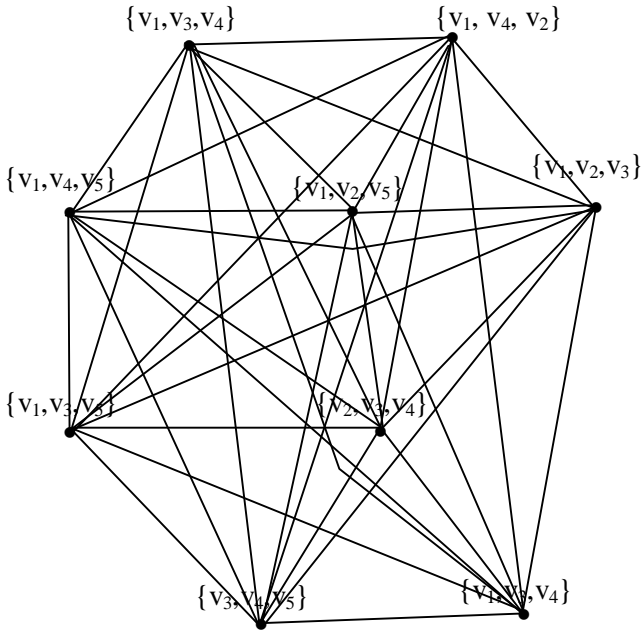

Figure 2.50

It is interesting to note that this subset subgraph of $G_v$ is unique for it a complete graph.

Each subset vertex is adjacent with 8 of the vertices. We next describe the vertex subset subgraph of $G_v$.

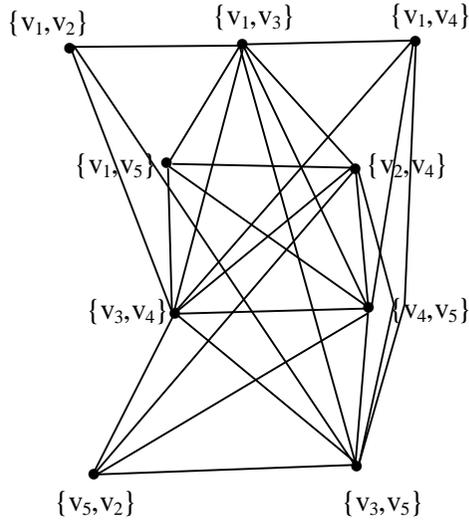

Figure 2.51

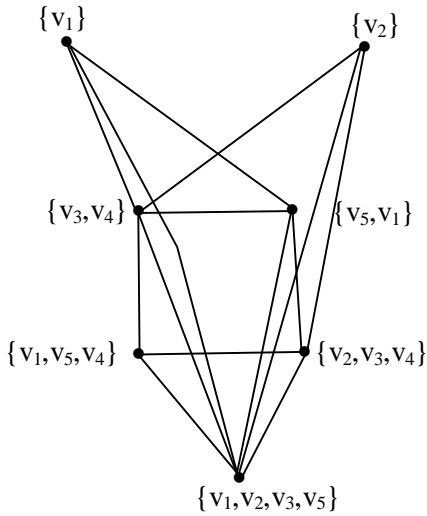

Figure 2.52

Each of the subset vertex has different sets of edges adjacent to it.

The vertex set $\{v_1, v_2, v_3, v_4\}$ has six edges adjacent to it.

The vertex $\{v_2, v_3, v_4\}$ has only four edges adjacent to it. Clearly the vertex $\{v_1\}$ and $\{v_2\}$ have only three edges adjacent to it.

Next the vertex $\{v_3, v_4\}$ is also important for 5 edges are adjacent to it.

Next consider the vertex subset subgraph with vertices $\{v_3\}$ $\{v_2, v_1\}$ and $\{v_4, v_5\}$;

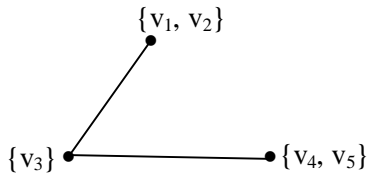

Figure 2.53

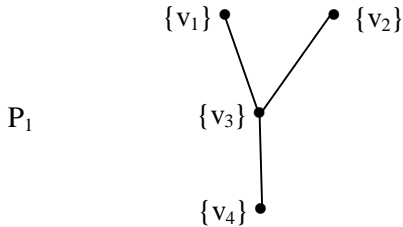

Figure 2.54

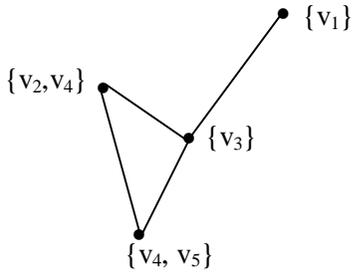

Figure 2.55

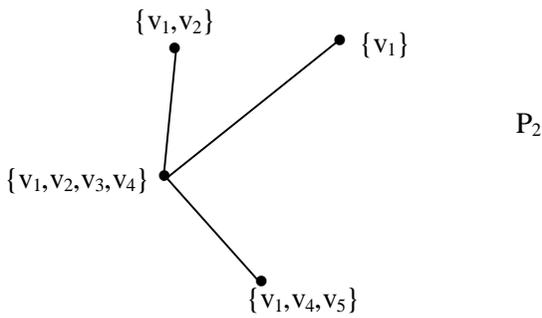

Figure 2.56

Now we see the two subset vertex graphs which are trees.

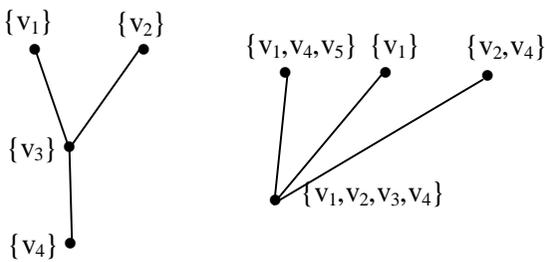

Figure 2.57

We try to merge these subset vertex trees.

$M_1 =$

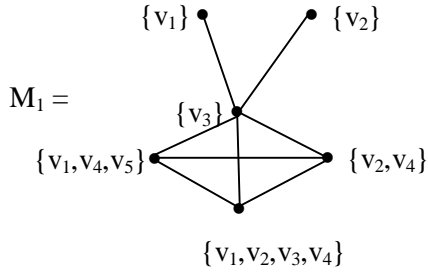

Figure 2.58

We see the merged graph of $P_1$ and $P_2$ is not a tree.

We call this the meshed tree for we get a very different structure which is not a tree.

Consider

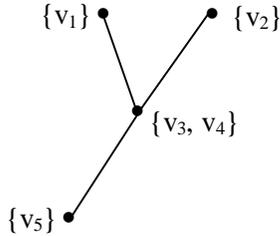

Figure 2.59

is again a subset vertex tree.

Let
$T_2 =$

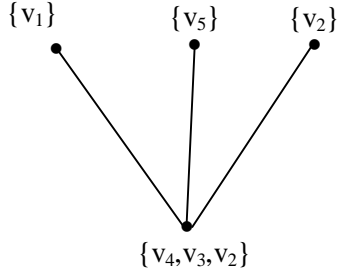

Figure 2.60

be another subset vertex tree.

We merge both $T_1$ and $T_2$

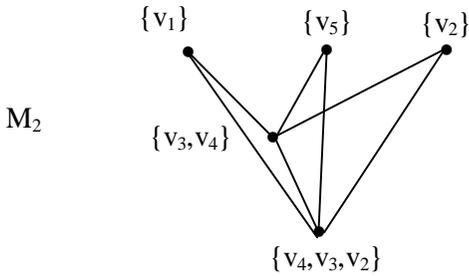

Figure 2.61

This is not a tree not even a meshed tree.

So $M_1$ and $M_2$ are distinctly different.

**Chapter Three**

# STRONG NEUTROSOPHIC GRAPHS

In this chapter we define yet a new class of graphs called strong neutrosophic graphs. The notion of neutrosophic graphs was recently introduced by the authors in [59].

In case of neutrosophic graphs we had the vertices to be real and not indeterminates. But in case of strong neutrosophic graphs we take some of the vertices to be also indeterminates that is neutrosophic.

We now proceed onto define the notion of strong neutrosophic graphs.

**DEFINITION 3.1:** *Let G = (V, E) be a graph with n number of vertices V = {$v_1$, $v_2$, ..., $v_n$} where k of the vertices are real and n–k of the vertices are neutrosophic or indeterminates (k ≥ 1); t of the edges are usual and p–t of the edges are neutrosophic (t ≥ 1). We define the graph G = (V, E) to be the strong neutrosophic graph.*

We will illustrate this situation by some examples.

***Example 3.1:*** Let G be a strong neutrosophic graph which is as follows:

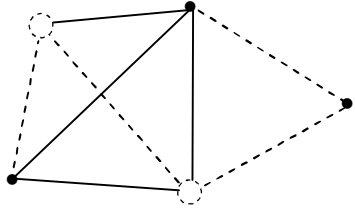

Figure 3.1

Clearly $\bigcirc$ denotes the neutrosophic vertex and $\bullet$ denotes the usual vertex.

We know

——— denotes usual edge and

- - - - - - denotes the neutrosophic edge.

Figure 3.2

***Example 3.2:*** We give strong neutrosophic graph G = (V, E) where o(V) = 3.

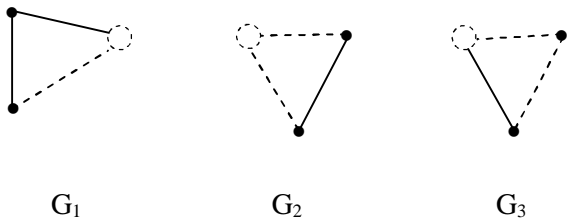

$G_1$ $G_2$ $G_3$

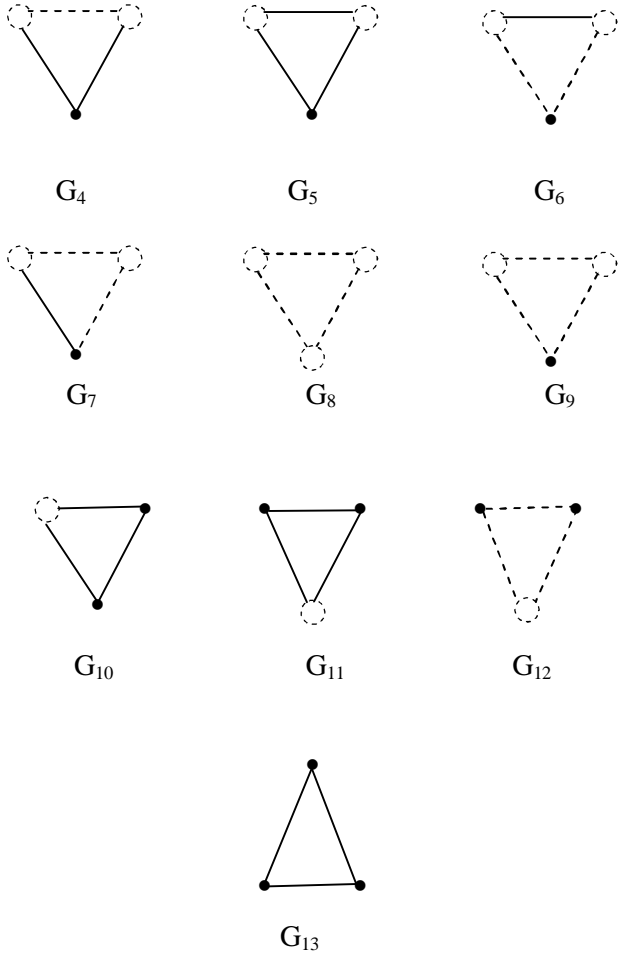

Figure 3.3

We see $G_5$ is not a strong neutrosophic graph.

We call edge weak neutrosophic graph. $G_8$ is a strong pure neutrosophic graph $G_{11}$ is also a edge weak neutrosophic graph.

$G_9$ and $G_{12}$ are complements of each other cannot be called as strong neutrosophic graphs.

***Example 3.3:*** We see G = (V, K), V = {$v_0$, $v_1$} and K = {e}.

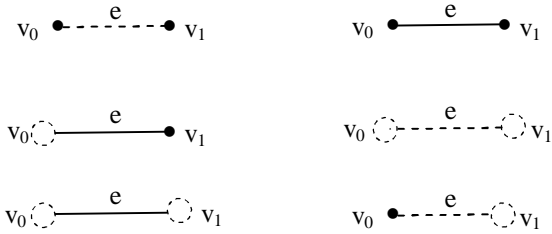

Figure 3.4

We see we have 6 different graphs with two vertices and one edge.

***Example 3.4:*** Let us consider G = {V, E} where V = {$v_0$, $v_1$, $v_2$, $v_3$} and E = {$e_0$, $e_2$, $e_1$, $e_3$}

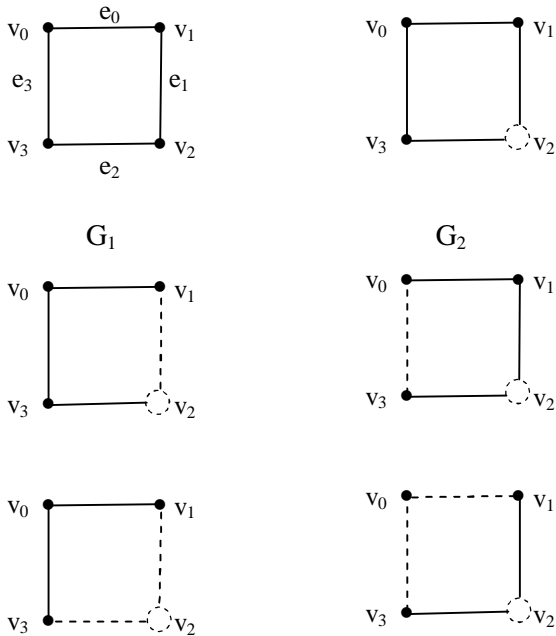

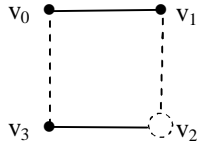

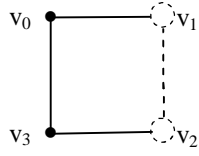

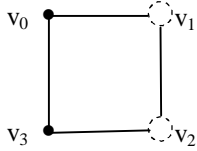

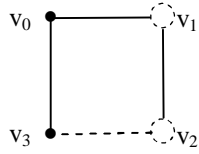

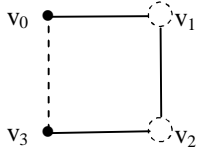

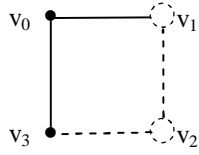

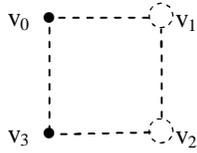

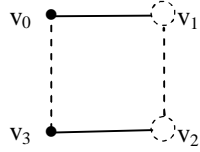

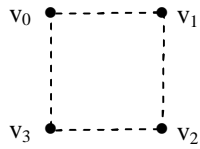

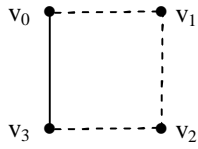

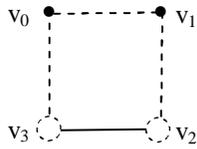

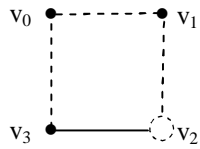

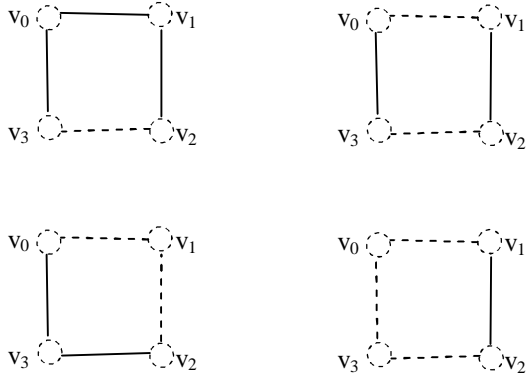

Figure 3.5

We see by including the usual graph, pure neutrosophic graph, semi strong neutrosophic, strong neutrosophic graphs and pure strong neutrosophic graphs we get a very large collection.

***Example 3.5:*** Let us consider the graphs which are all types not connected and planar with four vertices. Just G = {V, E}.

V is not connected or non planar V = {$v_0$, $v_1$, $v_2$, $v_3$} and E can have edges less than or equal to four.

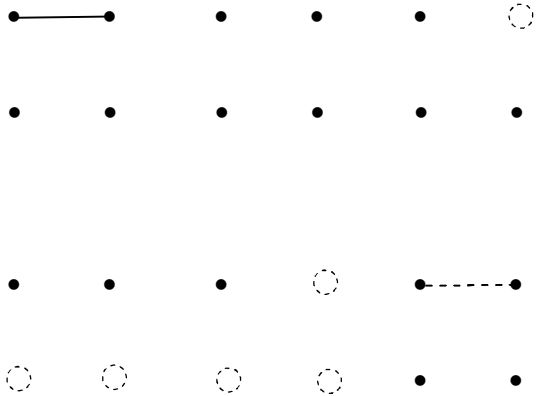

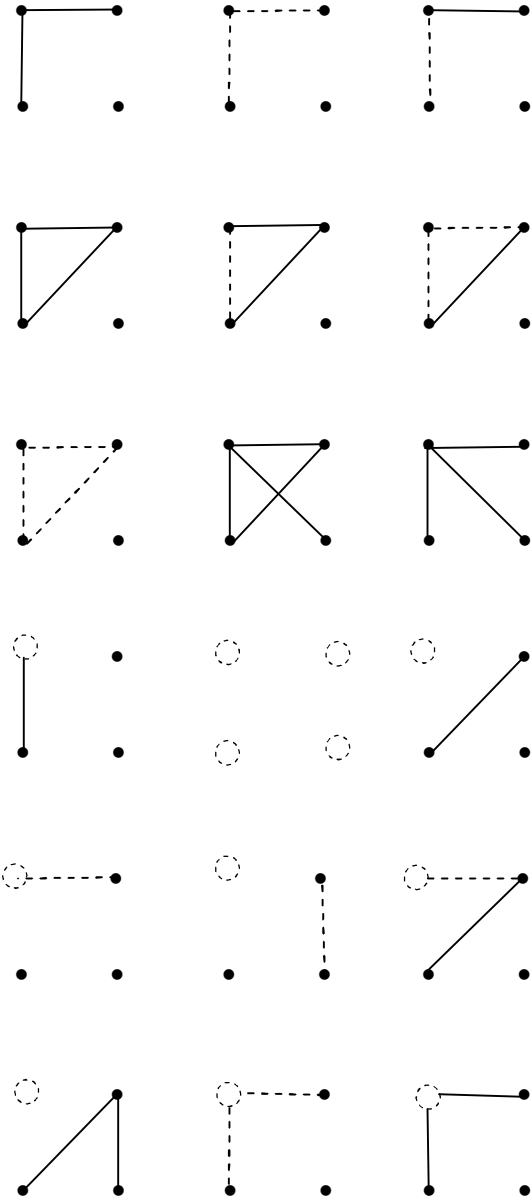

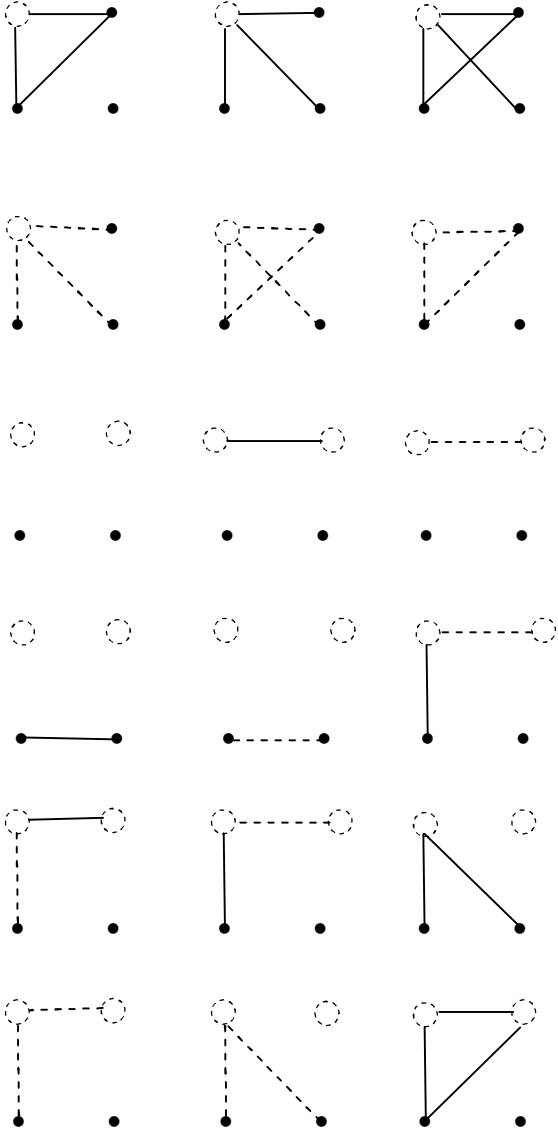

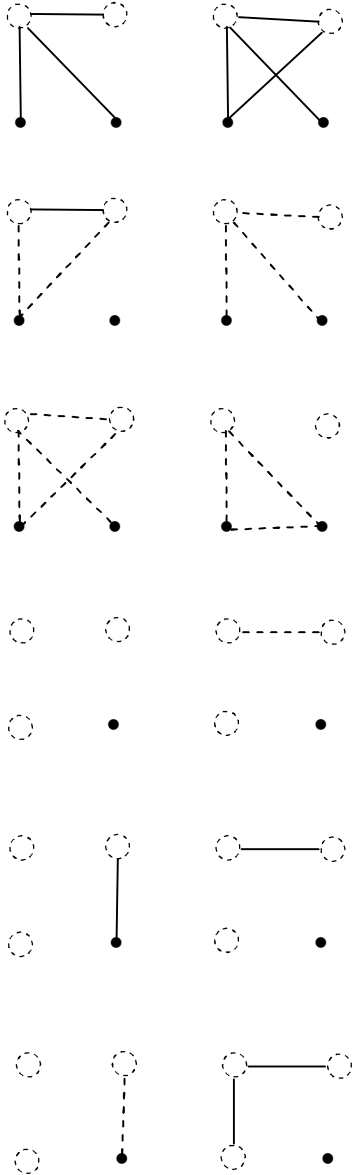

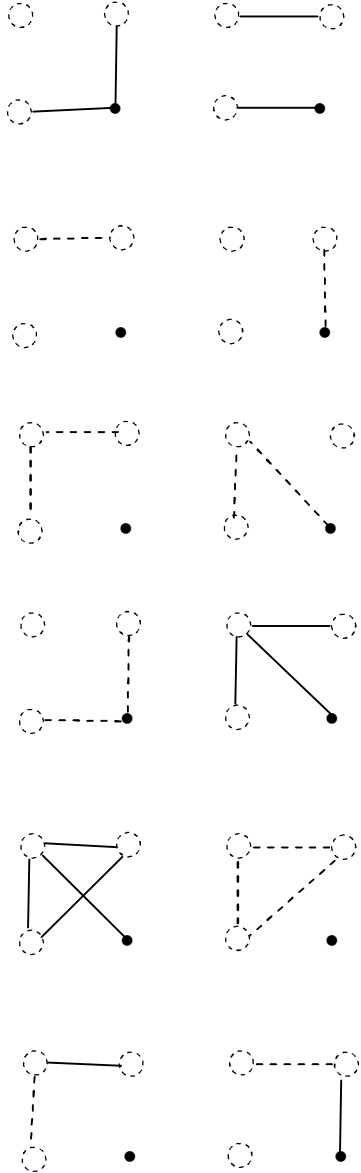

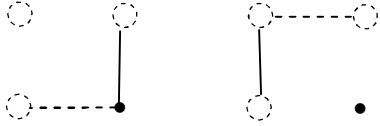

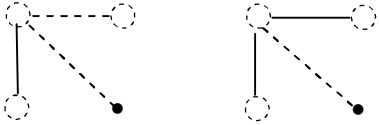

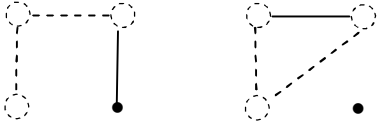

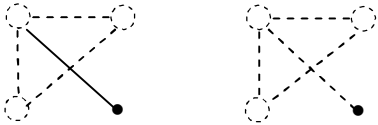

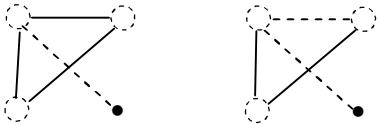

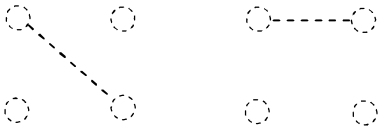

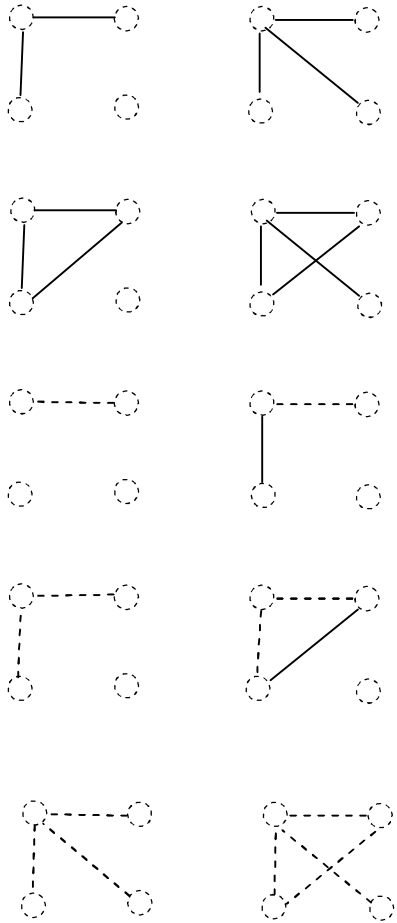

and so on.

Figure 3.6

Now we propose a few open problems.

**Problem 3.1:** Let G = (V, E) be a neutrosophic graph with n edges and n vertices such that G is planar.

- (a) Find the total number of neutrosophic graphs.
- (b) Find the total number of strong neutrosophic graphs.

(c)     Find the total number of semistrong edge neutrosophic graphs.

(d)     Find the total number of semistrong vertex neutrosophic graphs.

**Problem 3.2:**  Let G = (V, E) be a connected non planar graph with n vertices and p edges.

1. Find the total number of complete graphs which are usual, pure neutrosophic, neutrosophic, strong neutrosophic semistrong edge neutrosophic and semistrong vertex neutrosophic.

ii.     How many graphs are strong neutrosophic?

iii.    How many graphs are neutrosophic?

iv.     Find the number of graphs which are semi strong edge neutrosophic.

v.      Find the number of graphs which are semi strong vertex neutrosophic.

We give some example of strong neutrosophic trees and adjoint strong neutrosophic graphs.

***Example 3.6:***  Let G be a strong neutrosophic graph which is disjoint.

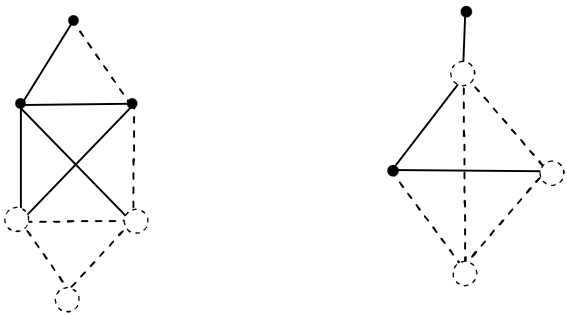

Figure 3.7

*Example 3.7:* Let G be a strong neutrosophic graph which is as follows:

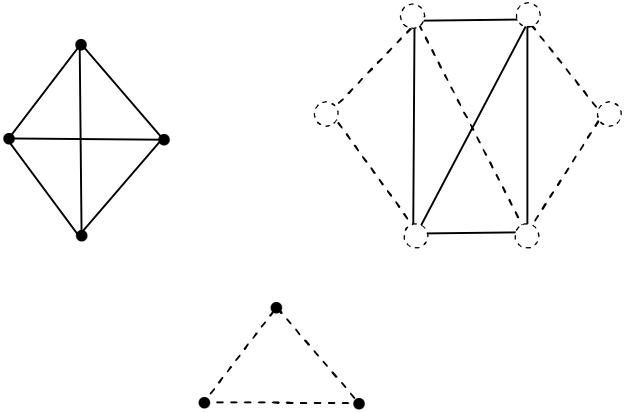

Figure 3.8

*Example 3.8:* Let G be a strong neutrosophic graph which is as follows:

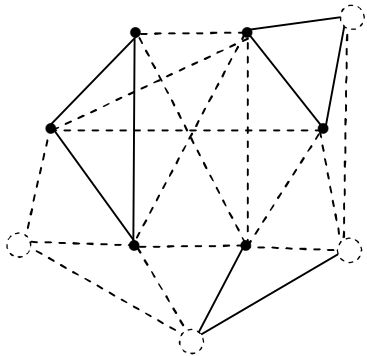

Figure 3.9

***Example 3.9:*** Let G be a strong neutrosophic graph which is as follows:

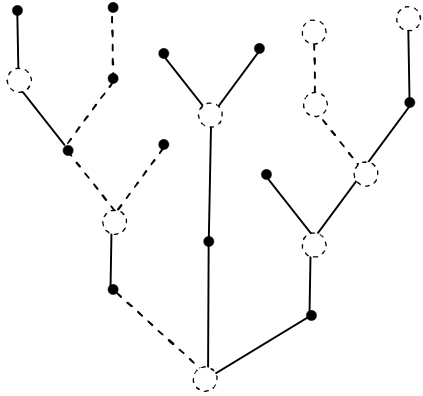

Figure 3.10

***Example 3.10:*** Let G be a strong neutrosophic graph which is as follows:

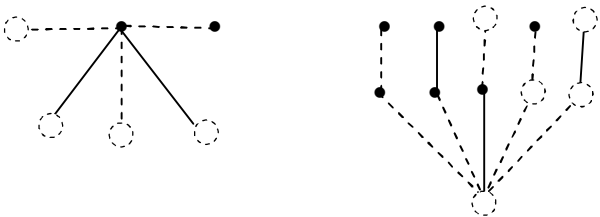

Figure 3.11

***Example 3.11:*** Let G be a strong neutrosophic graph which is as follows:

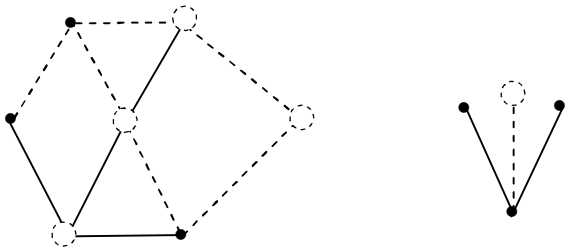

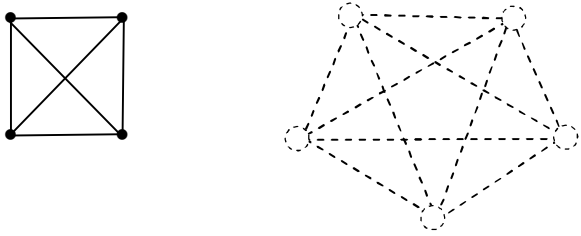

Figure 3.12

Clearly G is a strong neutrosophic graph which is disjoint.

**Example 3.12 :** Let G be a strong neutrosophic graph which is as follows:

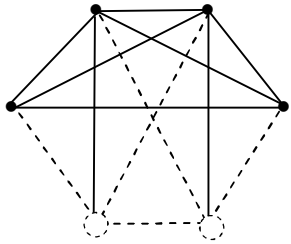

Figure 3.13

**Example 3.13:** Let G be a strong neutrosophic graph given in the following.

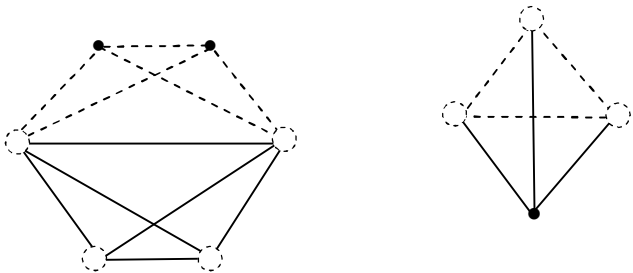

Figure 3.14

Clearly G is disjoint.

***Example 3.14:*** Let G be a strong neutrosophic graph which is as follows:

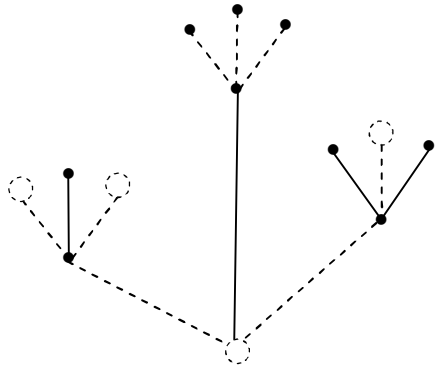

Figure 3.15

***Example 3.15:*** Let G be a strong neutrosophic graph which is as follows:

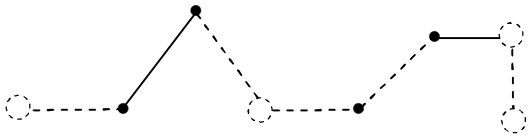

Figure 3.16

We have several examples some of them connected some a strong neutrosophic path some disjoint and so on.

Now we see by introducing the indeterminate vertices and indeterminate edges we get many graphs.

We get six strong graphs with two vertices and one edge.

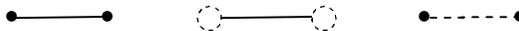

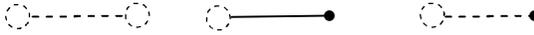

Only two neutrosophic graphs

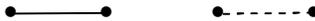

Figure 3.17

One may be curious to know why the authors have introduced the notion of indeterminate vertex and already indeterminate edges exist [59].

When some one works with a problem be it scientific or medical or social it is natural we at times are not in the position to spell about a vertex in such cases these graphs will play a vital role.

For instance in a social problem we many think about the existence of a node or not.

For if we are interested in studying the students problems among the concepts his performance is an indeterminate if the teacher is a partial teacher, many a times the node is indeterminate.

The concept of indeterminate vertex will be vital in case of network.

Even our brain cell at times keeps certain nodes to be indeterminate while associating them. So if we try to give the map by a graph we have the only appropriate graph is the strong neutrosophic graph. All most all the working of the brain is only in the form of graphs for any sort of it only strong neutrosophic graphs can serve the real purpose. For all the stimulai of the brain cannot and do not act simultaneously, several of them are in the indeterminate state only.

Thus authors feel, be it neutral networks or artificial intelligence it is strong neutrosophic graphs which is going to play a vital role.

For we are in the popular computer age where we view brain = computer (Kosko).

As rightly said by Kosko
"We shall explore machine intelligence from a dynamical systems view point brain = dynamical system."

Thus it is doubly confirmed brain can be appropriately described only by strong neutrosophic graphs where at a given time for a particular problem or situation  the brain synapse may be a usual graph or a neutrosophic graph or a strong neutrosophic graph.

All the while we have been only working using a usual graph.  It is time to get sensitive results, one needs to work with neutrosophic graphs [59] and strong neutrosophic graphs.

We have already used the concepts of neutrosophic graphs in NCMs, NRMs and NREs[25, 26].

We are yet to use the concept of strong neutrosophic graphs. Soon we shall construct many such in due course of time.

Now we see yet other type of neutrosophic graphs.

We may have the following type of graphs.

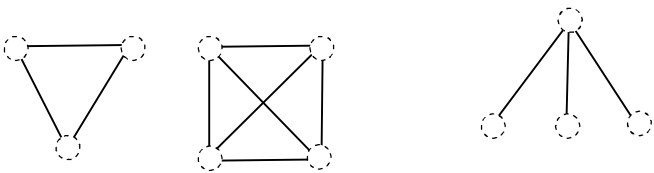

G =

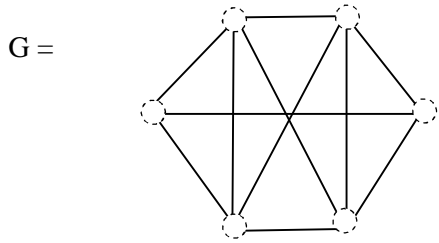

Figure 3.18

and so on. We see only vertices are neutrosophic but all the edges are real.

Thus this may so happen in case all the vertices are indeterminate at one time.

This graph will be called or defined as only vertex strong neutrosophic graph.

***Example 3.16:***

Consider

H =

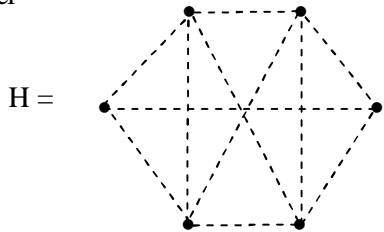

Figure 3.19

H is a pure neutrosophic graph. Clearly H is not strong neutrosophic graph as none of its vertices are neutrosophic.

We further observe that H is the quasi strong complement of G and G is the complement of H.

**Example 3.17:** Let G be the strong neutrosophic graph

G = 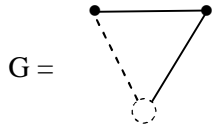

The strong complement of G is

H = 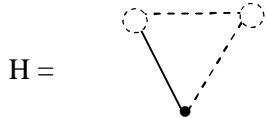

Figure 3.20

Clearly H and G are quasi strong complements. Infact G has no quasi strong complements.

**Example 3.18:** Let G be a strong neutrosophic graph the complement of G is H which is as follows:

G = 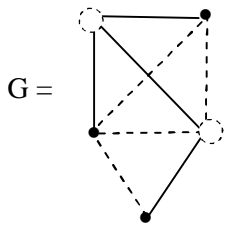

Figure 3.21

Now the strong complement of H is as follows:

H = 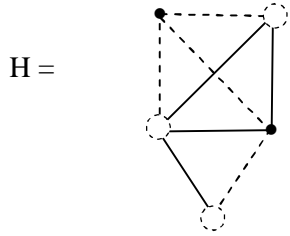

Figure 3.22

Clearly H is not a quasi strong complement of G. Infact G has no quasi strong complements.

***Example 3.19:*** Let G be a strong neutrosophic graph which is as follows.

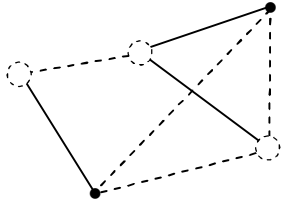

Figure 3.23

The strong neutrosophic complement of G is H which is as follows:

H = 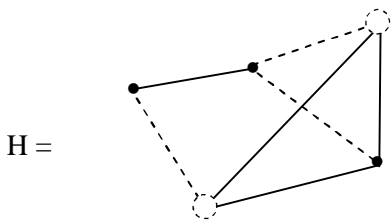

Figure 3.24

Clearly H is a strong neutrosophic graph.

***Example 3.20:*** Let G be a strong edge neutrosophic graph which is as follows:

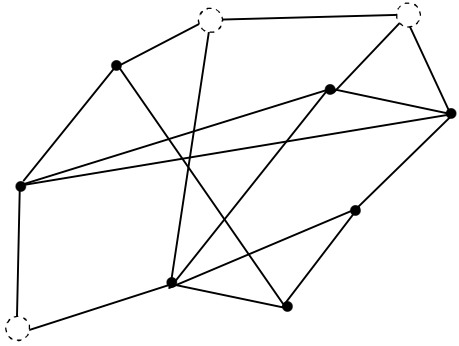

Figure 3.25

Let H be the strong complement of the neutrosophic graph H which is as follows:

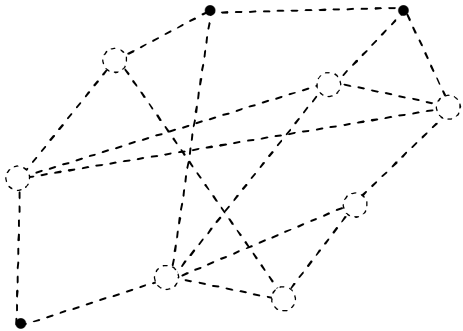

Figure 3.26

We see both G and H are neither neutrosophic nor strong neutrosophic graphs. For we see in G all the edges are real but some vertices are neutrosophic and some vertices are real.

Similarly we see H is neither neutrosophic nor a strong neutrosophic graph.

However G is the complement of H and vice versa.

We call the graphs of this type as quasi strong vertices neutrosophic graph G and H as quasi strong pure neutrosophic edge graphs.

We have several types of graphs which are illustrated.

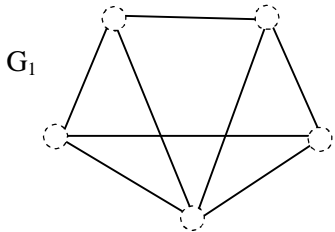

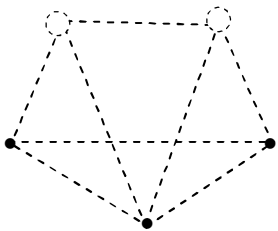

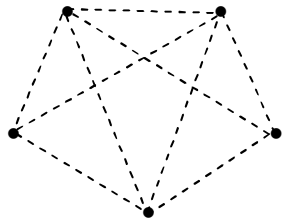

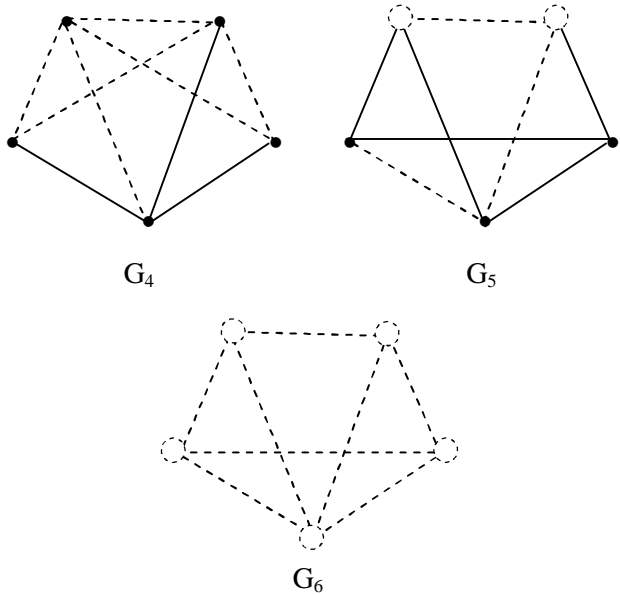

Figure 3.27

G$_1$ is neither neutrosophic nor pure neutrosophic or strong neutrosophic graph.

G$_2$ is also neither neutrosophic nor pure neutrosophic or a strong neutrosophic graph.

G$_3$ is pure neutrosophic.

G$_5$ is strong neutrosophic.

G$_4$ is also not neutrosophic or strong neutrosophic nor pure neutrosophic.

G$_6$ is strong pure neutrosophic. Thus some of them are quasi strong vertex (edge) neutrosophic.

We will give some more examples.

***Example 3.21:*** Let G be the graph which is as follows:

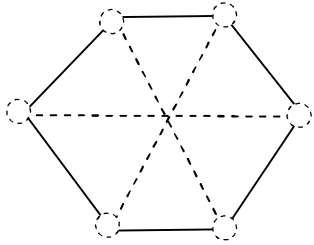

Figure 3.28

This graph is not neutrosophic and not strongly neutrosophic.

The strong complement of G is as follows:

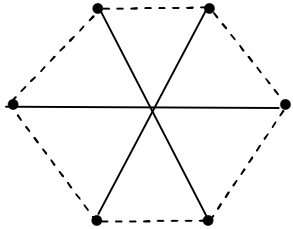

Figure 3.29

Clearly H is a neutrosophic graph and not a strong neutrosophic graph.

***Example 3.22:*** Let G be a strong neutrosophic graph.

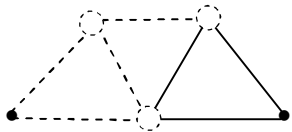

Figure 3.30

The complement of G is as follows

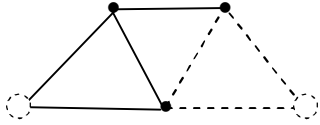

Figure 3.31

***Example 3.23:*** Let G be a strong neutrosophic graph.

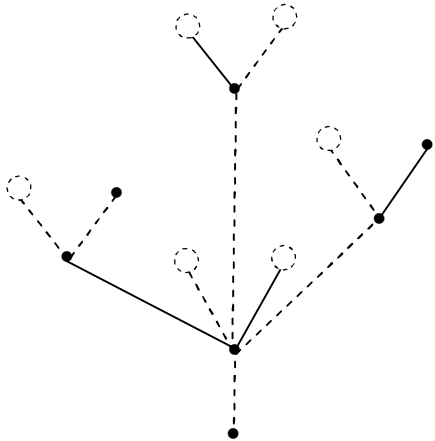

Figure 3.32

The complement of strong neutrosophic graph is as follows.

H be the complement of G which is as follows:

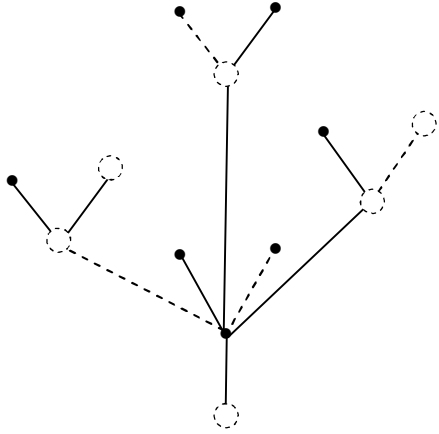

Figure 3.33

This tree is also a strong neutrosophic graph.

***Example 3.24:*** Let G be a strong neutrosophic graph which is as follows:

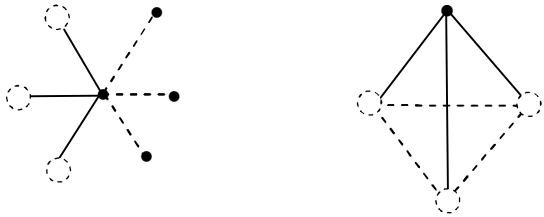

Figure 3.34

The strong neutrosophic complement of G be H.

H is as follows:

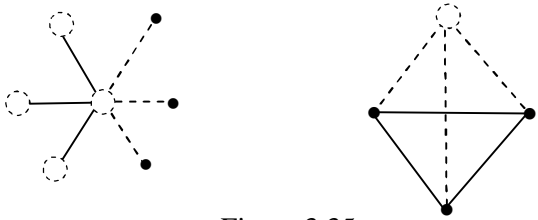

Figure 3.35

**Example 3.25:** Let G be a strong neutrosophic graph which is as follows:

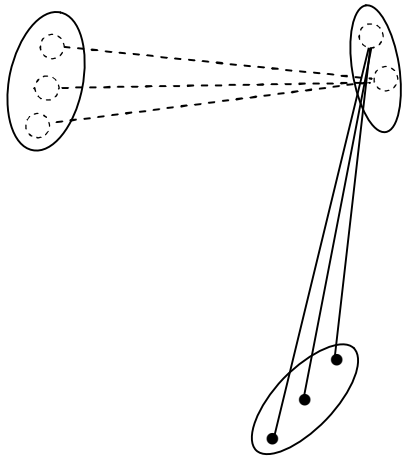

Figure 3.36

The strong neutrosophic complement of G is as follows:

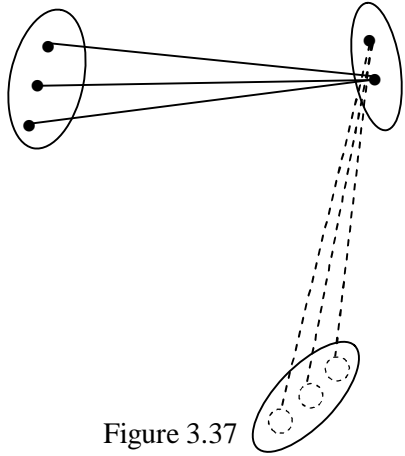

Figure 3.37

We see H and G are identical. So we call G to be a self complemented strong neutrosophic graph.

We can have several self complemented strong neutrosophic graphs.

**Example 3.26:** Let G be strong neutrosophic graph which is as follows:

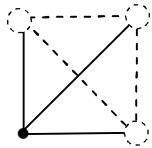

Figure 3.38

Let H be the complement of G which is as follows:

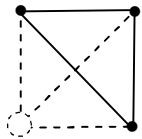

Figure 3.39

We see G and H are identical.

Thus G is a self complemented strong neutrosophic graph.

***Example 3.27:*** Let G be a strong complemented graph which is as follows:

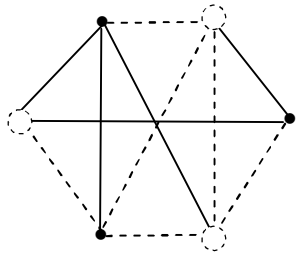

Figure 3.40

The complement H of G is as follows:

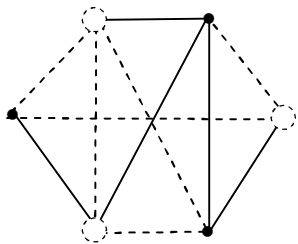

Figure 3.41

Clearly G is a self complement strong neutrosophic graph.

Now we define a new notion.

**DEFINITION 3.2:** *Let G be a neutrosophic graph. We have H the neutrosophic complement of G. We define quasi vertex*

*strong neutrosophic complement of G in which all the vertices are indeterminate.*

We will show this by some examples.

***Example 3.28:*** Let G be a neutrosophic graph which is as follows:

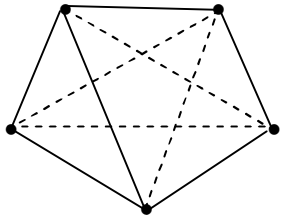

The complement H of G is as follows:

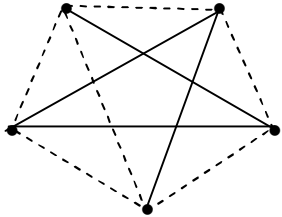

Figure 3.42

The strong complement K of G is as follows:

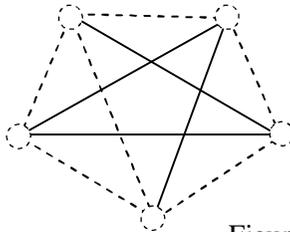

Figure 3.43

Clearly K is not a strong neutrosophic graph.

***Example 3.29:*** Let G be a neutrosophic graph which is follows:

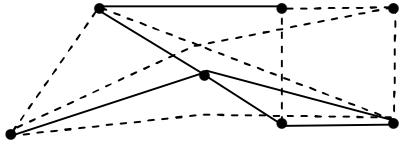

Figure 3.44

The neutrosophic complement K of G is as follows:

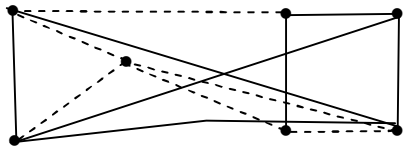

Figure 3.45

The strong neutrosophic complement H of G is as follows:

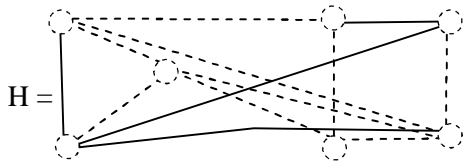

Figure 3.46

Clearly H is not a strong neutrosophic graph.

We see for a given neutrosophic graph we can have two distinct complements.

***Example 3.30:*** Let G be a pure neutrosophic graph which is as follows:

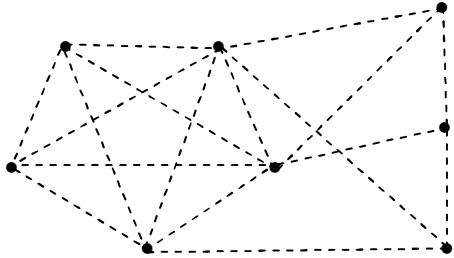

Figure 3.47

The complement of G is the usual graph H which is as follows:

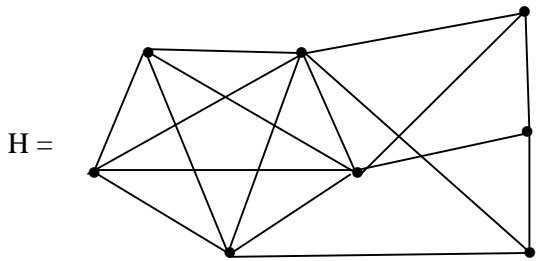

Figure 3.48

We see the complement of pure neutrosophic graph is always a usual graph.

However the strong neutrosophic complement P of G is as follows:

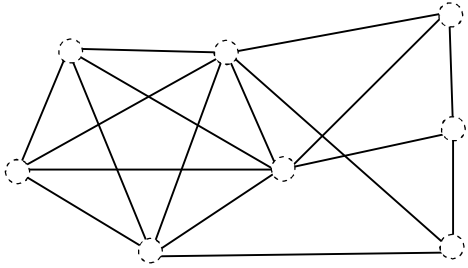

Figure 3.49

***Example 3.31:*** Let G be a neutrosophic graph which is as follows:

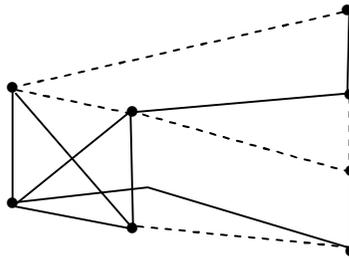

Figure 3.50

The complement H of G is as follows:

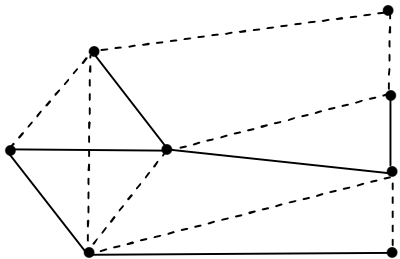

Figure 3.51

Clearly H is also a neutrosophic graph.  The strong neutrosophic complement of G be K.  K is as follows:

K =
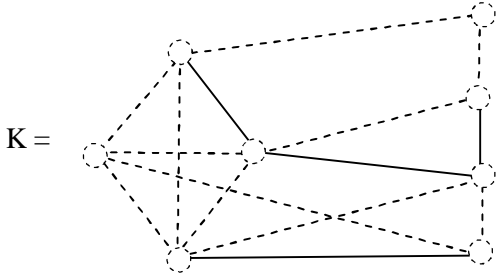

Figure 3.52

Clearly K is not a strong neutrosophic graph.

We see the complement of the usual graph is a pure strong doubly neutrosophic graph.

We see the pure strong doubly neutrosophic graph is not in the class of strong neutrosophic graph.  We see G is a usual graph.

G =
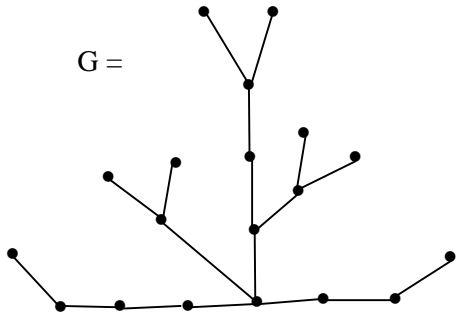

Figure 3.53

The neutrosophic graph H is as follows:

H =

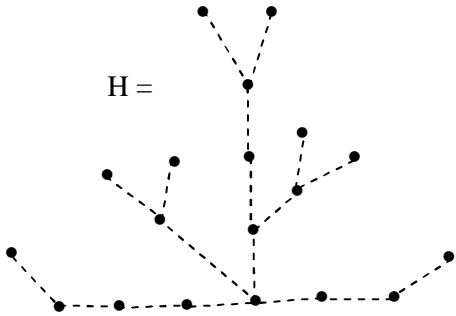

Figure 3.54

Clearly H is a pure neutrosophic graph.

Now we find the strong neutrosophic complement K of graph G which is as follows:

K =

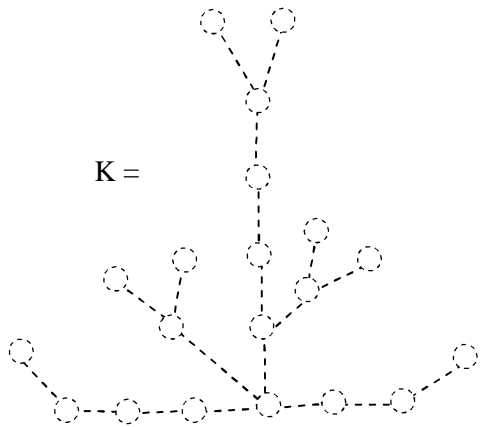

Figure 3.55

We see K is not a strong neutrosophic pure graph.

We see under complements we cannot at all times accept the graph is in the collection.

So to this end we define super strong neutrosophic graphs which contains the usual graph, pure neutrosophic graph semi strong vertex neutrosophic graph and semi strong edge neutrosophic graph.

This for a given number of points and edges is analysed.

For a given number of vertices and given number of edges we have one and only one usual graph and one and only one pure neutrosophic graph and one, only one semi strong vertex neutrosophic graph, one and only graph which semi strong edge neutrosophic graph and finally one and only pure strong neutrosophic graph.

However we have several neutrosophic graphs and several strong neutrosophic graph.

We will first illustrate this situation by some examples.

G = 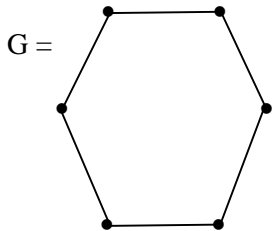

Figure 3.56

G is a usual graph.

$G_1$ 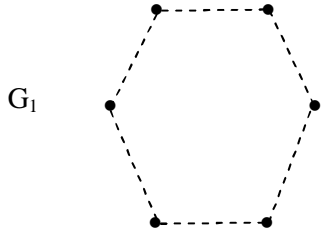

Figure 3.57

$G_1$ is the pure neutrosophic graph.

$G_2$ is the strong pure neutrosophic graph which is as follows

$G_2 =$ 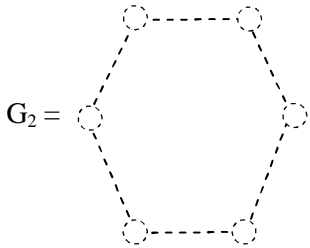

Figure 3.58

Let $G_3$ be the semi strong neutrosophic vertex graph which is as follows:

$G_3$ 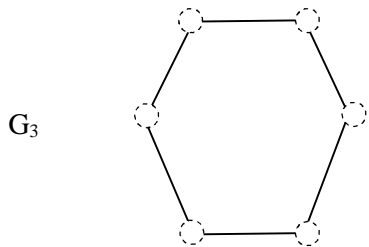

G₄ 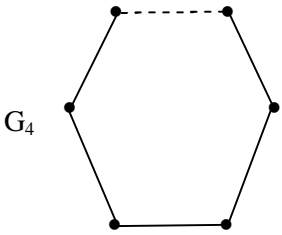

G₅ 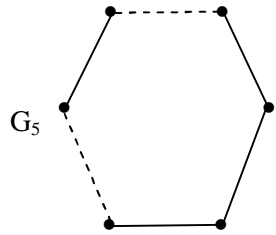

G₆ 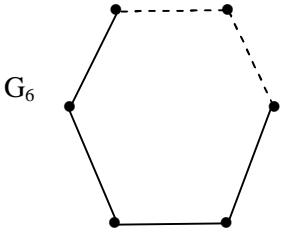

G₇ 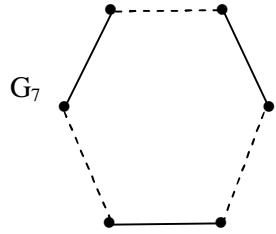

G₈ 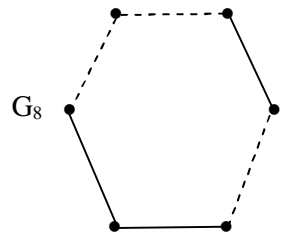

G₉ 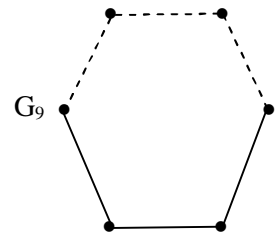

G₁₀ 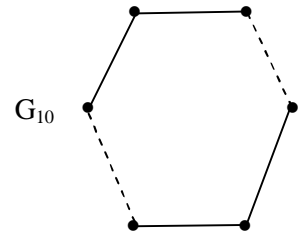

G₁₁ 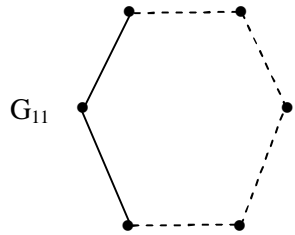

$G_{12}$ 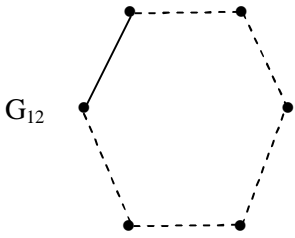

$G_{13}$ 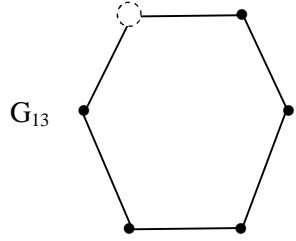

$G_{14}$ 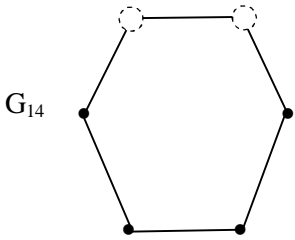

$G_{15}$ 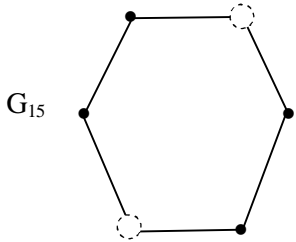

$G_{16}$ 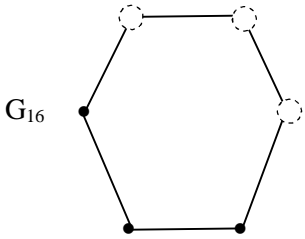

$G_{17}$ 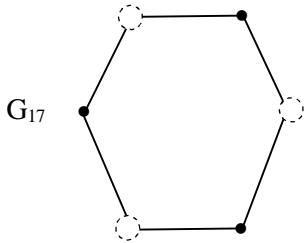

$G_{18}$ 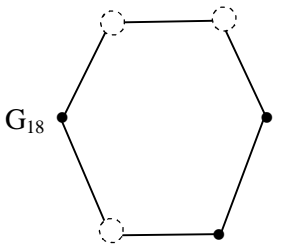

$G_{19}$ 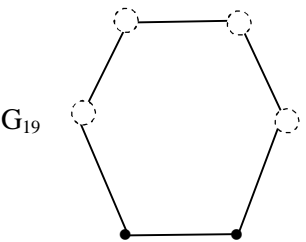

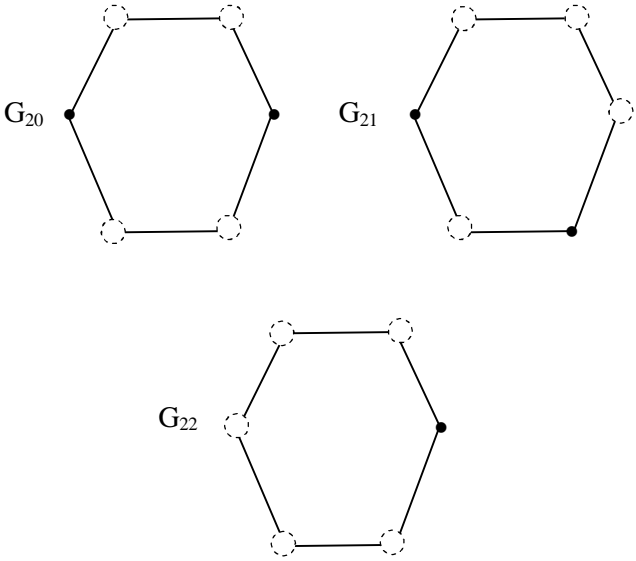

Figure 3.59

and so on.

We see the class of super strong neutrosophic graphs has over 22 of them.

***Example 3.32:*** We now give an example with three vertices and three edges.

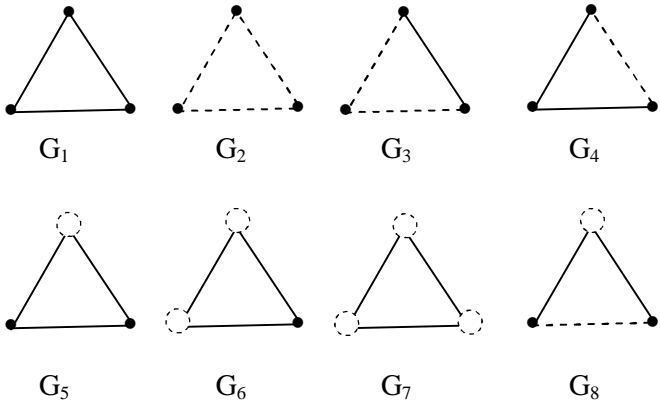

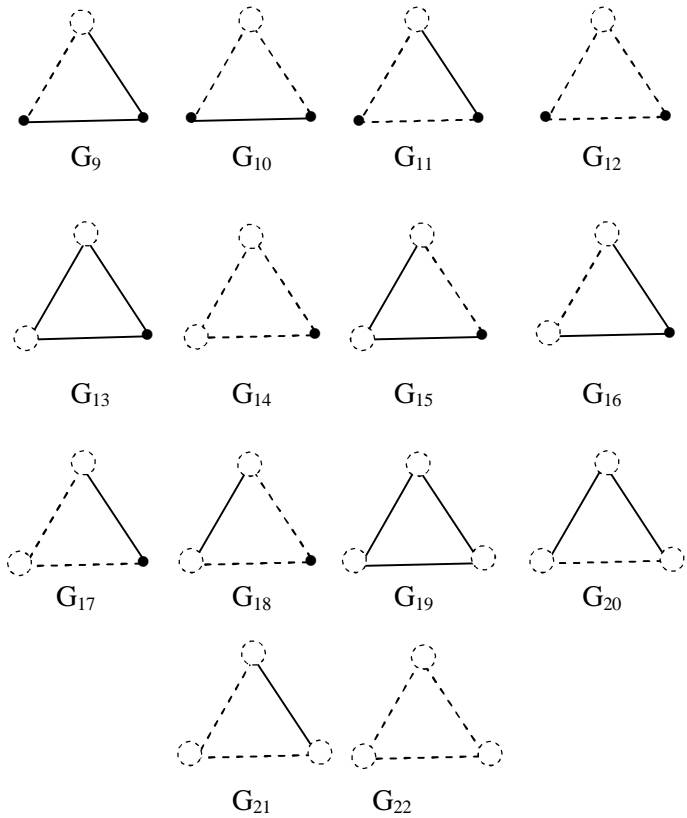

Figure 3.60

We have 22 graphs in class of super strong neutrosophic graphs.

***Example 3.33:*** We now study the graph G with two vertices and two edges.

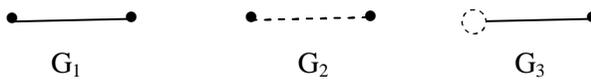

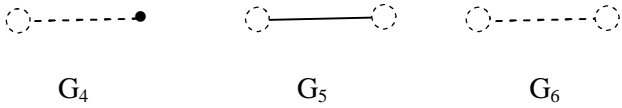

G₄        G₅        G₆

Figure 3.61

We have only 6 super strong neutrosophic graphs.

***Example 3.34:*** Let us consider the strong neutrosophic graph G.

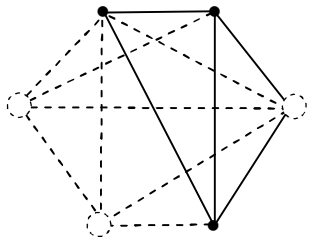

Figure 3.62

We find the subgraphs of G.

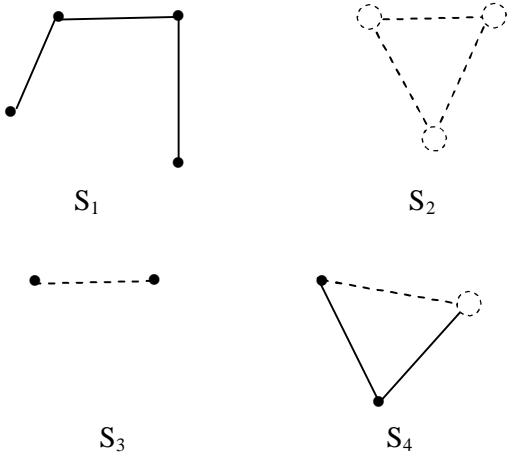

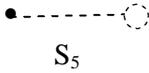

$S_5$

Figure 3.63

$S_1$ is a usual subgraph,
$S_3$ is a pure neutrosophic subgraph,
$S_5$ is not strong neutrosophic subgraph,
$S_4$ is a strong neutrosophic subgraph and
$S_5$ is not a strong neutrosophic subgraph.

Thus $S_1$, $S_2$, …, $S_5$ belong to the class of super strong neutrosophic graphs.

***Example 3.35:*** Let G be a strong neutrosophic graph. The subgraphs of G need not be a strong neutrosophic subgraphs.

However subgraphs of G is in the class of super strong neutrosophic subgraphs.

We want to illustrate this by some examples.

***Example 3.36:*** Let G be a strong neutrosophic graph which is as follows:

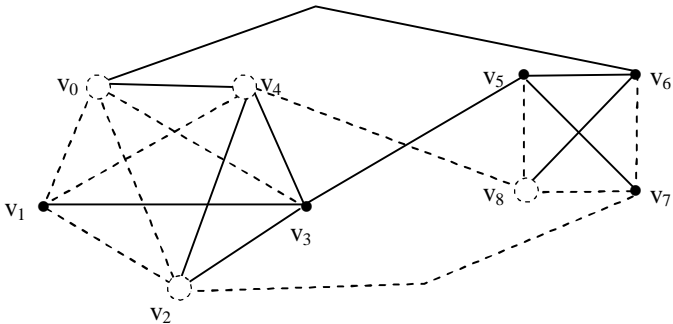

Figure 3.64

We see this strong neutrosophic graphs has subgraphs which are not strong neutrosophic graphs.

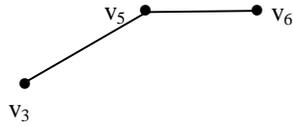

$S_1$

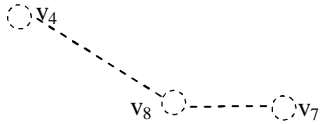

$S_2$

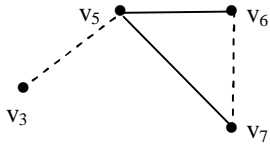

$S_3$

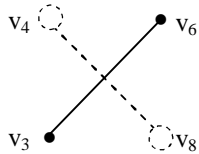

$S_4$

Figure 3.65

Of the four subgraphs only $S_4$ is a strong neutrosophic graph.

$S_1$ is a usual graph.
$S_2$ is a pure neutrosophic graph.
$S_3$ is a neutrosophic graph.

$S_3$ has no neutrosophic vertices.

So we see the subgraphs  general need not enjoy the properties of the graph.

So when we build structures on the collection of all subgraphs we accept this and make the class a larger one.

We can define connected path and walk in case of strong neutrosophic graphs also.

We will illustrate this situation by some examples.

Let G be a strong neutrosophic graph which is as follows:

We define a strong neutrosophic walk is defined as a finite alternating sequence of vertices and edges which some of the vertices must be neutrosophic and some must necessarily be neutrosophic edges.

In a neutrosophic walk we have a sequence of vertices and edges where vertices are real and necessarily some of the edges are neutrosophic.

We will illustrate both the situations before we proceed to define more concepts in strong neutrosophic graphs.

***Example 3.37:*** Let G be a strong neutrosophic graph.

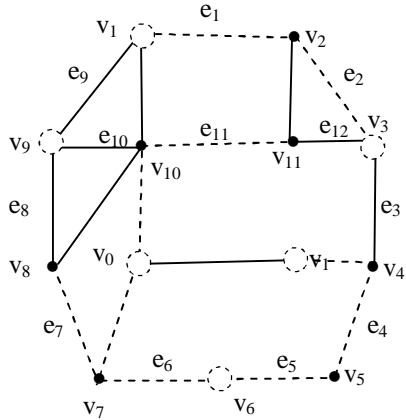

Figure 3.66

A walk is $v_1 \ e_1 \ v_2 \ e_2 \ v_3 \ e_3 \ v_4 \ e_4 \ v_5 \ e_5 \ v_6 \ e_6 \ v_7 \ e_7 \ v_8 \ e_8$ is a strong neutrosophic walk.

$v_6 \ e_6 \ v_7 \ e_7$ is a strong neutrosophic walk.

$v_4 e_4 v_5$ is only a neutrosophic walk not a strong neutrosophic walk.

$v_8 \ e_{10} \ v_{10} \ e_{11} \ v_{11}$ is a usual walk.

Thus a strong neutrosophic graph can have all types of graph.

***Example 3.38:*** Let G be a strong neutrosophic graph which is as follows:

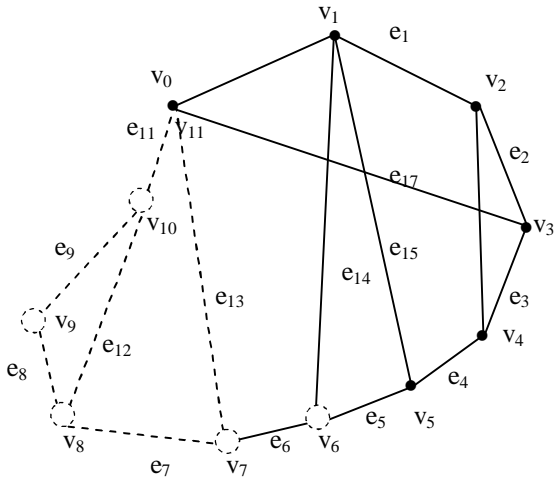

Figure 3.67

We see this strong neutrosophic graph has all types of types.

$v_0\ e_0\ v_1\ e_1\ v_2\ e_2\ v_3\ e_3\ v_4\ e_4\ v_5$ is a usual walk.

$v_{10}\ e_{10}\ v_{11}\ e_{11}\ v_0\ e_0\ v_1$ is a strong neutrosophic walk.

$v_7\ e_7\ v_8\ e_8\ v_9\ e_9\ v_{10}\ e_{10}$ is a pure strong neutrosophic graph.

This strong neutrosophic graph has no neutrosophic walk.

**Example 3.39:** Let G be a strong neutrosophic graph which is as follows:

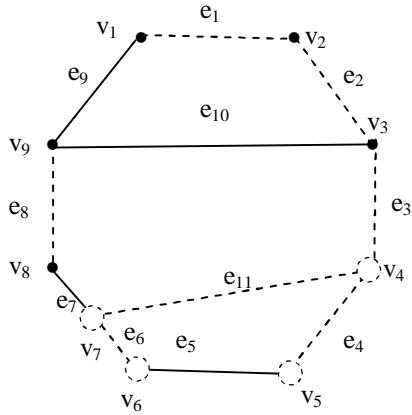

Figure 3.68

Consider $v_2$ $e_2$ $v_3$ $e_3$ $e_{10}$ $v_9$ $e_9$ $v_1$ is a usual walk.

Consider $v_3$ $e_3$ $v_4$ $e_4$ $e_{11}$ $v_7$ is a strong pure neutrosophic walk.

$v_9$ $e_9$ $v_1$ $e_1$ $v_2$ $e_2$ $v_3$ is a neutrosophic walk.

$v_3$ $e_3$ $v_4$ $e_4$ $v_5$ $e_5$ $v_6$ $e_6$ $v_7$ $e_7$ $v_8$ is a strong neutrosophic graph.

Thus we have seen all types of walks is a strong neutrosophic graph.

A open walk in which no vertex appears more than once is a path.

We all several types of paths.

Infact we have as many as number of distinct walks as distinct paths.

We will illustrate this by some examples.

**Example 3.40:** Let G be a strong neutrosophic graph which is as follows:

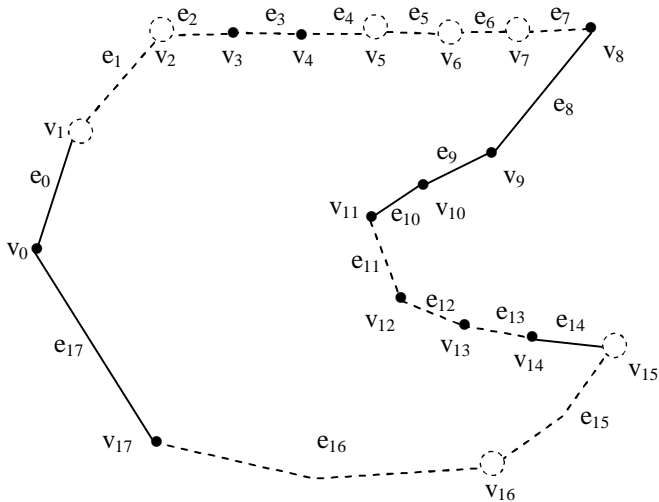

Figure 3.69

Consider $p_1$ - $v_{17}$ $e_{17}$ $v_0$ $e_0$ $v_1$ $e_1$ $v_2$ $e_2$ $v_3$ $e_3$ $v_4$ .

$p_1$ is a open strong neutrosophic walk and hence a open strong neutrosophic path.

Consider $p_2$ : $v_5$ $e_5$ $v_6$ is a pure strong neutrosophic open walk hence $p_2$ is a pure strong neutrosophic path.

Let $p_3$ : $v_8$ $e_8$ $v_9$ $e_9$ $v_{10}$; $p_3$ is a open usual walk which is also a usual path.

Let $p_4$ : $v_{11}$ $e_{11}$ $v_{12}$ $e_{12}$ $v_{13}$ $e_{13}$ $v_{14}$ $e_{14}$ $v_{15}$ is a open neutrosophic walk which is also a neutrosophic path.

**Example 3.41:** Let G be a strong neutrosophic graph which also follows:

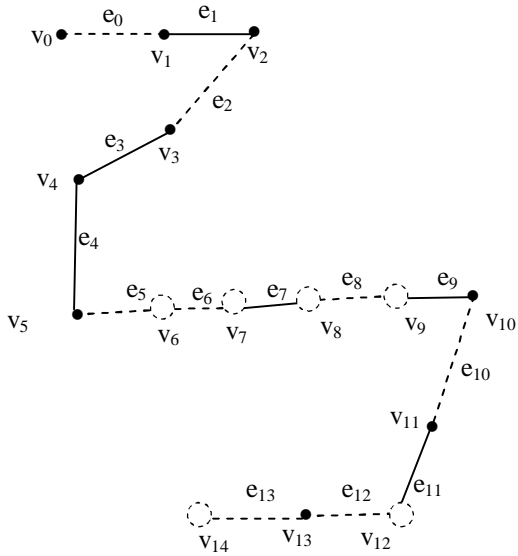

Figure 3.70

Let $p_1$ : $v_0$ $e_0$ $v_1$ $e_1$ $v_2$ $e_2$ $v_3$ $e_3$ $v_4$ is a open neutrosophic walk which is also a neutrosophic path.

Clearly $p_1$ is not a strong neutrosophic(open walk) path.

Consider $p_2$ : $v_4$ $e_4$ $v_5$ $v_6$ $e_6$ $v_7$ $e_7$ $v_8$ $e_8$ $v_9$ is a open strong neutrosophic walk which is also a strong neutrosophic path.

Consider $p_3$ : $v_1$ $e_1$ $v_2$ is also a open usual walk hence the usual path.

Let $p_4$ : $v_8$ $e_8$ $v_9$ is a pure neutrosophic open walk which is a pure neutrosophic path.

The number of edges correspond to the length of a path.

In case of both neutrosophic and strong neutrosophic path we have some edges to be usual and some edges to be neutrosophic.

Suppose we have n edges in a strong neutrosophic graph or a neutrosophic graph then certainly we must have n = s + t where s ≥ 1 and t ≥ 1 with s corresponding to neutrosophic edges and t corresponding to the usual edges.

In case of pure neutrosophic path or strong pure neutrosophic path of length n we have all the n-edges to be neutrosophic.

We will first illustrate this by some examples.

**Example 3.42:** Let G be a strong neutrosophic graph which is as follows:

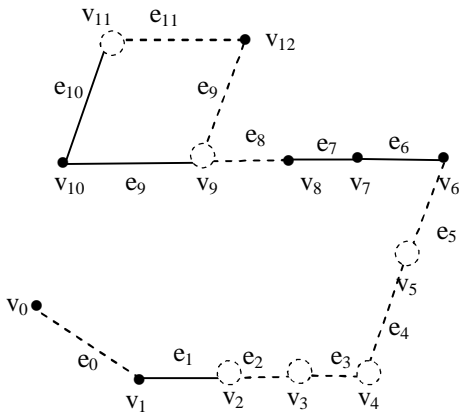

Figure 3.71

Consider $p_1 : v_{10} \, e_9 \, v_9 \, e_8 \, v_8 \, e_7 \, v_7 \, e_6 \, v_6$; $p_1$ is the usual path that is a open walk.

Length of the path is four.

Let $p_2 : v_{10}\ e_{10}\ v_{11}\ e_{11}\ v_{12}$, $p_2$ is a strong neutrosophic path or a strong neutrosophic open walk two $e_{10}$ is real and $e_{11}$ is neutrosphic.

$p_3 : v_8\ e_7\ v_7\ e_6\ v_6\ e_5\ v_5$ is a neutrosophic path or neutrosophic open walk of length three of which two edges are reals and one edge is neutrosophic.

Let $p_4 : v_2\ e_2\ v_3\ e_3\ v_4$ be a strong pure neutrosophic path or a pure neutrosophic open walk of length two both the edges are neutrosophic.

Recall a closed walk in which no vertex repeats is a circuit.

We have four types of circuits.

We will describe this by some examples.

**Example 3.43:** Let G be a strong neutrosophic path

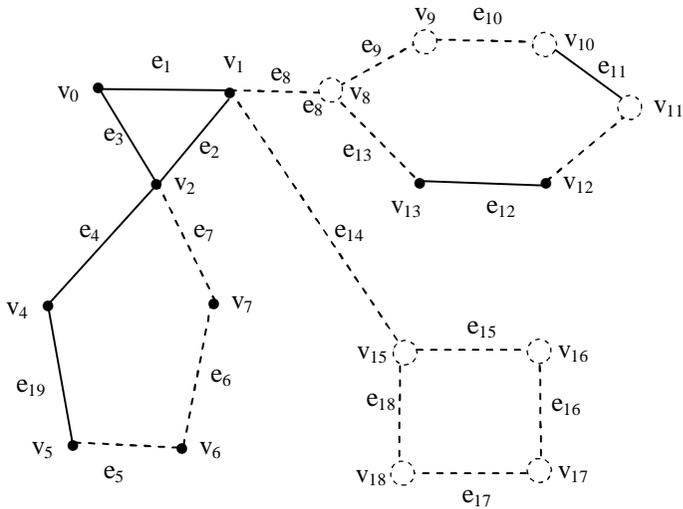

Figure 3.72

$p_1 : v_0\ e_1\ v_1\ e_2\ v_2\ e_3\ v_0$

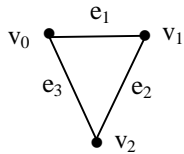

Figure 3.73

$p_1$ is a circuit which is neither neutrosophic nor pure neutrosophic nor strong neutrosophic but only usual circuit.

$p_2$ : $v_8$ $e_9$ $v_9$ $e_{10}$ $e_{11}$ $v_{11}$ $e_{12}$ $v_{12}$ $e_{13}$ $v_{13}$ $e_8$

$p_2$ is a circuit which is a strong neutrosophic circuit.

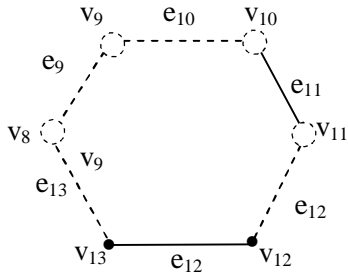

Figure 3.74

$p_3$ : $v_2$ $e_7$ $v_7$ $e_6$ $v_0$ $e_5$ $v_5$ $e_{19}$ $v_4$ $e_4$ $v_2$.

$p_3$ is a neutrosophic circuit which is not a strong neutrosophic circuit.

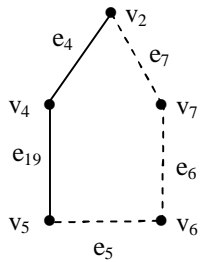

Figure 3.75

Let $p_4$ :

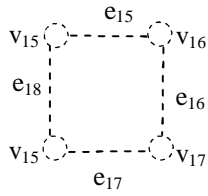

Figure 3.76

$p_4$ is a pure neutrosophic circuit.

$p_4 : v_{15} \ e_{15} \ v_{16} \ e_{16} \ v_{17} \ e_{17} \ v_{18} \ e_{18} \ v_{15}.$

***Example 3.44:*** Let G be a strong neutrosophic graphs which is as follows:

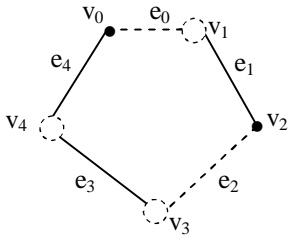

Figure 3.77

We see G has no proper circuits.

The only circuit is G itself which is a strong neutrosophic circuit.

***Example 3.45:*** Let G be a strong neutrosophic path.

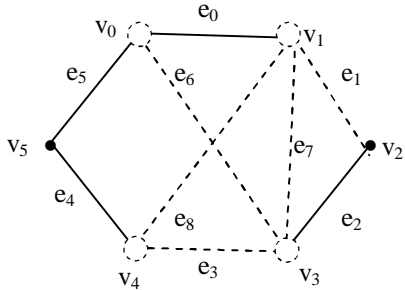

Figure 3.78

Clearly G has no closed usual circuits.

$p_1$ : $v_0$ $e_0$ $v_1$ $e_1$ $v_2$ $e_2$ $v_3$ $e_6$ $v_0$ strong neutrosophic circuit which as follows:

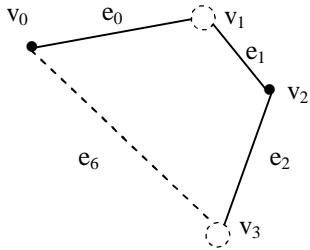

Figure 3.79

Consider $p_2$ : $v_1$ $e_7$ $v_3$ $e_3$ $v_4$ $e_8$ $v_1$;
a pure neutrosophic circuit.

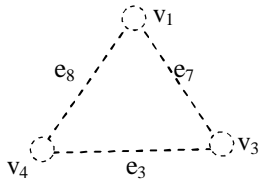

Figure 3.80

Clearly G has no neutrosophic circuit.

Thus we see a given strong neutrosophic graph can in general need not have all the four types of circuits.

***Example 3.46:*** Let G be a strong neutrosophic graph which is as follows.

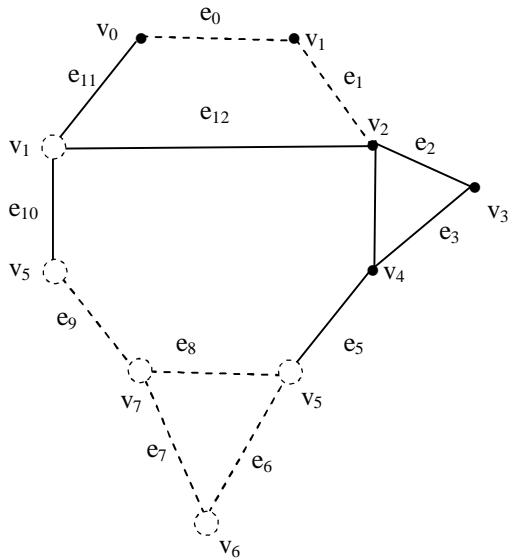

Figure 3.81

Let $p_1$ :

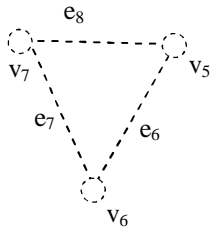

Figure 3.82

a pure neutrosophic circuit $p_1 : v_7 \, e_8 \, v_5 \, e_6 \, v_6 \, e_7 \, v_7$

$p_2 : v_2 \, e_2 \, v_3 \, e_3 \, v_4 \, e_4 \, v_2$ be a usual circuit which is as follows:

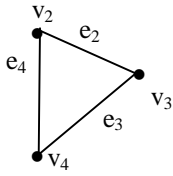

Figure 3.83

$p_3 : v_9 \, e_{11} \quad v_0 \, e_0 \, v_1 \, e_1 \, v_2 \, e_{12} \, v_9$ be a strong neutrosophic circuit which is as follows:

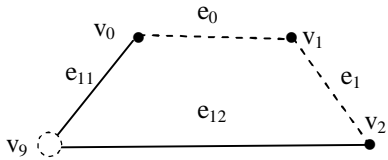

Figure 3.84

Clearly this graph has no neutrosophic circuit.

***Example 3.47 :*** Let G be a strong neutrosophic graph.

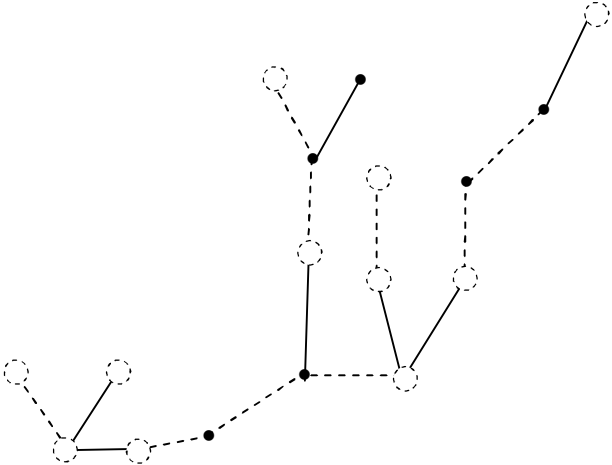

Figure 3.85

This strong neutrosophic graph has no circuits.

***Example 3.48:*** Let G be a strong neutrosophic graph which is as follows:

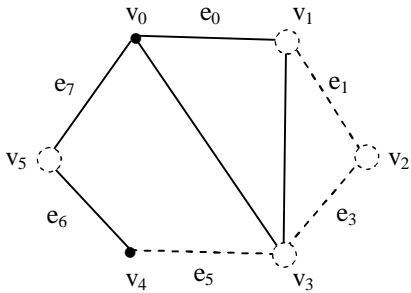

Figure 3.86

This is a strong neutrosophic unicursal graph.

We can have several examples unicursal pure neutrosophic graphs and neutrosophic graphs also exist.

***Example 3.49:*** Let G be a neutrosophic graph which is as follows:

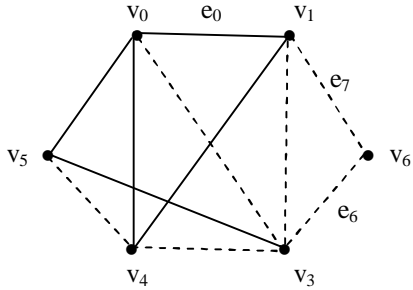

Figure 3.87

Check G is a unicursal graph.

Now we can as in case of usual graphs have disconnected strong neutrosophic graph.

We will give examples of them.

***Example 3.50:*** Let G be a strong neutrosophic graph which is as follows:

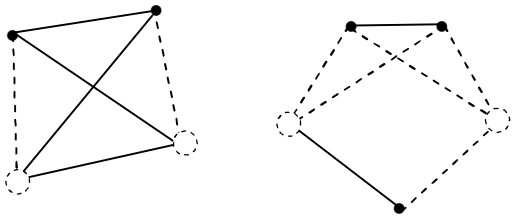

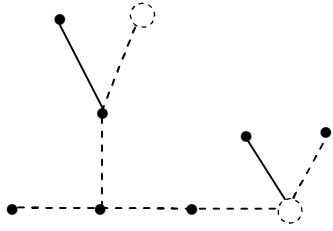

Figure 3.88

Clearly G is a disconnected.

***Example 3.51:*** Let G be a strong neutrosophic graph.

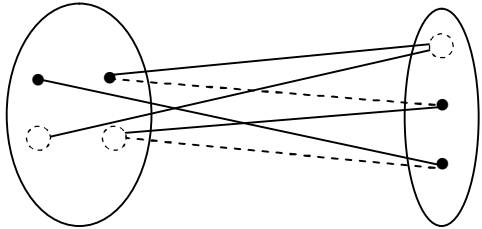

Figure 3.89

Suppose we have a strong neutrosophic graph G of the form

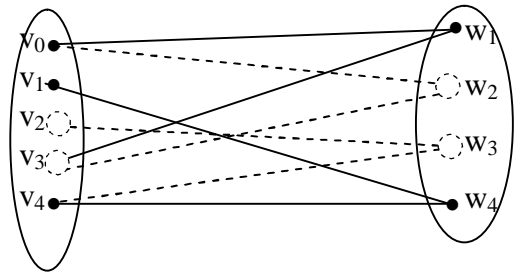

Figure 3.90

We see G is a strong neutrosophic bipartite graph. These graphs can be used in strong Neutrosophic Relational Models and Strong Neutrosophic Relation Equations.

***Example 3.52:*** Let G be a strong neutrosophic graph which is as follows:

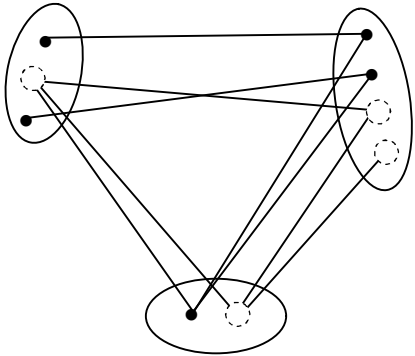

Figure 3.91

G is clearly a tripartite strong neutrosophic graph.

***Example 3.53:*** Let G be a 4 partite graph which is strong neutrosophic.

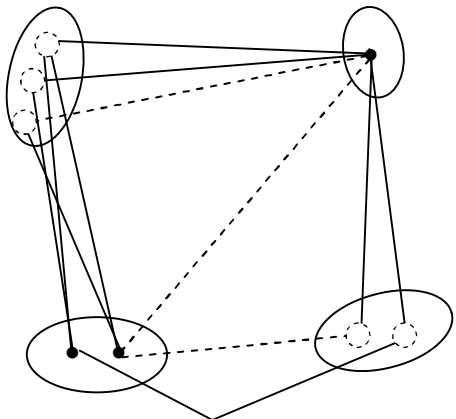

Figure 3.92

This strong neutrosophic graph is a four partite graph.

We call a strong neutrosophic graph as k-neutrosophic partite if the vertex set $\{v_1, \ldots, v_n\}$ is partitioned into k subsets that no vertices joints in subsets and each of the sets contain either only real vertices or only neutrosophic vertices, that is thee is no mixed vertices in the subsets.

We will illustrate this situation by some examples.

***Example 3.54:*** Let G be a strong neutrosophic graph which is as follows:

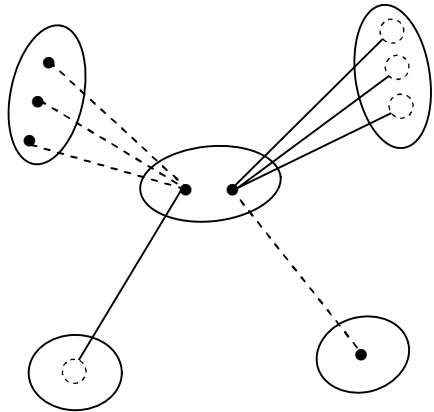

Figure 3.93

Clearly G is a strong 5-partite neutrosophic graph.

***Example 3.55:*** Let G be a strong neutrosophic graph which is as follows:

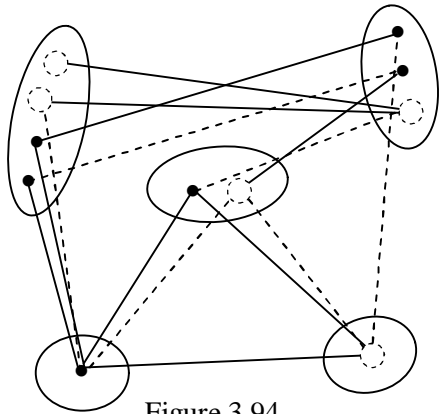

Figure 3.94

Clearly G is 5-partite graph which is strong neutrosophic. But G is not a strong 5-neutrosophic partite graph.

Thus from these two graphs one easily sees the difference.

***Example 3.56:*** Let G be a 3-partite strong neutrosophic graph which is as follows:

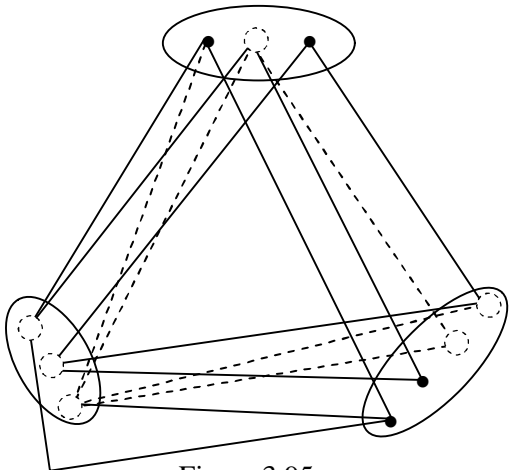

Figure 3.95

Interested reader can obtain any number of such k-partite strong neutrosophic graph.

**Chapter Four**

# SPECIAL SUBGRAPH TOPOLOGICAL SPACES

In this chapter we for the first time define the notion of special subgraph topological spaces.

We know given a set of vertices $\{v_1, v_2\}$ we can have the graphs $\bullet\, v_1, \bullet\, v_2$ $\left\{ \begin{matrix} v_1 & v_2 \\ \bullet & \bullet \end{matrix} \right\}$ and $v_1 \bullet\!\!\!\text{————}\!\!\!\bullet\, v_2$ .

Figure 4.1

Consider the set $S(G) = \{\phi,\ \left\{ \begin{matrix} v_1 \\ \bullet \end{matrix} \right\}, \left\{ \begin{matrix} v_2 \\ \bullet \end{matrix} \right\},\ \{v_1 \bullet\quad v_2 \bullet\},$

$v_1 \bullet\!\!\!\text{————}\!\!\!\bullet\ v_2$ } we call the empty set to be the least element and a complete graph with two vertices as the greatest element of G.

We with these gradations we make $S(G)$ a topological space.

Suppose we have G to be a complete graph with three vertices $v_1$, $v_2$ and $v_3$ to find S(G) of $v_1$, $v_2$ and $v_3$.

$$S(G) = \{\phi, \left\{\begin{matrix}\bullet \\ v_1\end{matrix}\right\}, \left\{\begin{matrix}\bullet \\ v_2\end{matrix}\right\}, \left\{\begin{matrix}\bullet \\ v_3\end{matrix}\right\}, \left\{\begin{matrix}\bullet & & \bullet \\ v_1 & & v_2\end{matrix}\right\}, \left\{\begin{matrix}\bullet & & \bullet \\ v_1 & & v_3\end{matrix}\right\},$$

$$\left\{\begin{matrix}\bullet & & \bullet \\ v_2 & & v_3\end{matrix}\right\}, \left\{\begin{matrix}\bullet\!\!-\!\!\bullet \\ v_1 & v_2\end{matrix}\right\}, \left\{\begin{matrix}\bullet\!\!-\!\!\bullet \\ v_2 & v_3\end{matrix}\right\}, \left\{\begin{matrix}\bullet\!\!-\!\!\bullet \\ v_3 & v_1\end{matrix}\right\},$$

$$\{\bullet\ v_1, v_2\ \bullet \quad \bullet\ v_3\}, \left\{\begin{matrix}\bullet\!\!-\!\!\bullet & v_3 \\ v_1 & v_2 & \bullet\end{matrix}\right\}, \left\{\begin{matrix}\bullet & \bullet\!\!-\!\!\bullet \\ v_1 & v_2 & v_3\end{matrix}\right\},$$

$$\left\{\begin{matrix}\bullet & \bullet\!\!-\!\!\bullet \\ v_2 & v_1 & v_3\end{matrix}\right\}, \left\{\begin{matrix}v_1 \\ \\ v_2 \quad v_3\end{matrix}\right\}, \left\{\begin{matrix}v_1 \\ \\ v_2 \quad v_3\end{matrix}\right\}, \left\{\begin{matrix}v_1 \\ \\ v_2 \quad v_3\end{matrix}\right\},$$

$$\left\{\begin{matrix}v_1 \\ \triangle \\ v_2 \quad v_3\end{matrix}\right\}\} \}.$$

Figure 4.2

We see $\phi$ is the least element of S(G) and 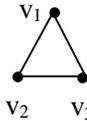

Figure 4.3

is the greatest element.

However the number of elements in S(G) is 18.

Let S(G) be a collection of all subgraphs of the graph G

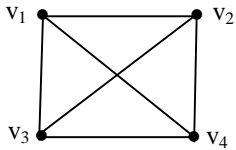

Figure 4.4

with vertices $\{v_1, v_2, v_3, v_4\}$;

$$S(G) = \{\phi, \{\overset{\bullet}{v_1}\}, \{\overset{\bullet}{v_2}\}, \{\overset{\bullet}{v_3}\}, \{\overset{\bullet}{v_4}\}, \{\overset{\bullet}{v_1} \; \overset{\bullet}{v_2}\},$$

$$\{\overset{\bullet}{v_1} \; \overset{\bullet}{v_3}\}, \{\overset{\bullet}{v_1} \; \overset{\bullet}{v_4}\}, \{\overset{\bullet}{v_2} \; \overset{\bullet}{v_3}\}, \{\overset{\bullet}{v_2} \; \overset{\bullet}{v_4}\},$$

$$\{\overset{\bullet}{v_3} \; \overset{\bullet}{v_4}\}, \quad \{\overset{\bullet}{v_1} \; \overset{\bullet}{v_2} \; \overset{\bullet}{v_3}\}, \{\overset{\bullet}{v_1} \; \overset{\bullet}{v_2} \; \overset{\bullet}{v_4}\},$$

$$\{\overset{\bullet}{v_1} \; \overset{\bullet}{v_3} \; \overset{\bullet}{v_4}\}, \{\overset{\bullet}{v_2} \; \overset{\bullet}{v_3} \; \overset{\bullet}{v_4}\}, \{\overset{\bullet}{v_1} \!-\! \overset{\bullet}{v_2} \; \overset{\bullet}{v_3}\},$$

$$\{\overset{\bullet}{v_1} \!-\! \overset{\bullet}{v_2} \; \overset{\bullet}{v_4}\}, \{\overset{\bullet}{v_2} \!-\! \overset{\bullet}{v_3} \; \overset{\bullet}{v_1}\}, \{\overset{\bullet}{v_1} \; \overset{\bullet}{v_2} \; \overset{\bullet}{v_3} \; \overset{\bullet}{v_4}\},$$

$$\{\overset{\bullet}{v_1} \!-\! \overset{\bullet}{v_2} \; \overset{\bullet}{v_3} \; \overset{\bullet}{v_4}\}, \{\overset{\bullet}{v_1} \!-\! \overset{\bullet}{v_3} \; \overset{\bullet}{v_2} \; \overset{\bullet}{v_4}\}, \{\overset{\bullet}{v_2} \!-\! \overset{\bullet}{v_3} \; \overset{\bullet}{v_4}\},$$

$$\{\overset{\bullet}{v_1} \!-\! \overset{\bullet}{v_4} \; \overset{\bullet}{v_2} \; \overset{\bullet}{v_3}\}, \{\overset{\bullet}{v_2} \!-\! \overset{\bullet}{v_3} \; \overset{\bullet}{v_1} \; \overset{\bullet}{v_4}\},$$

$$\{\overset{\bullet}{v_2} \!-\! \overset{\bullet}{v_4} \; \overset{\bullet}{v_1} \; \overset{\bullet}{v_3}\}, \{\overset{\bullet}{v_3} \!-\! \overset{\bullet}{v_4} \; \overset{\bullet}{v_1} \; \overset{\bullet}{v_2}\},$$

$$\{\overset{\bullet}{v_3} \!-\! \overset{\bullet}{v_4} \; \overset{\bullet}{v_1}\}, \{\overset{\bullet}{v_1} \!-\! \overset{\bullet}{v_3} \; \overset{\bullet}{v_2}\}, \{\overset{\bullet}{v_1} \!-\! \overset{\bullet}{v_3} \; \overset{\bullet}{v_4}\},$$

$$\{\overset{\bullet}{v_1} \!-\! \overset{\bullet}{v_2} \; \overset{\bullet}{v_3} \!-\! \overset{\bullet}{v_4}\}, \{\overset{\bullet}{v_1} \!-\! \overset{\bullet}{v_3} \; \overset{\bullet}{v_2} \!-\! \overset{\bullet}{v_4}\},$$

$$\{\overset{\bullet}{v_1} \!-\! \overset{\bullet}{v_4} \; \overset{\bullet}{v_2} \!-\! \overset{\bullet}{v_3}\}, \{\overset{\bullet}{v_1} \!-\! \overset{\bullet}{v_4} \; \overset{\bullet}{v_3}\}, \{\overset{\bullet}{v_1} \!-\! \overset{\bullet}{v_4} \; \overset{\bullet}{v_2}\},$$

$$\{\overset{\bullet}{v_3} \!-\! \overset{\bullet}{v_4} \; \overset{\bullet}{v_2}\}, \{\overset{\bullet}{v_2} \!-\! \overset{\bullet}{v_4} \; \overset{\bullet}{v_3}\}, \{\overset{\bullet}{v_2} \!-\! \overset{\bullet}{v_4} \; \overset{\bullet}{v_1}\},$$

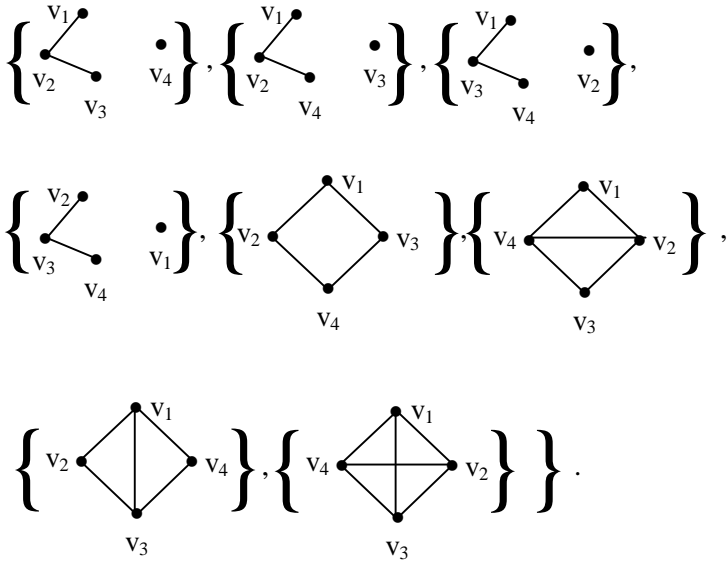

Figure 4.5

The greatest element of S(G) is 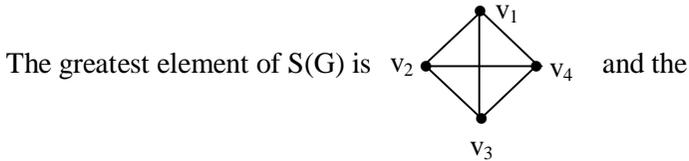 and the

Figure 4.6

least element is $\phi$. Number of elements in S(G) is 46.

Suppose S(G) is a collection of all subgraphs with five vertices $\{v_1, v_2, v_3, v_4, v_5\}$. To find S(G).

$$S(G) = \{\phi, \left\{\begin{matrix} \bullet \\ v_1 \end{matrix}\right\}, \left\{\begin{matrix} \bullet \\ v_2 \end{matrix}\right\}, \left\{\begin{matrix} \bullet \\ v_3 \end{matrix}\right\}, \left\{\begin{matrix} \bullet \\ v_4 \end{matrix}\right\}, \left\{\begin{matrix} \bullet \\ v_5 \end{matrix}\right\}, \left\{\begin{matrix} \bullet & \bullet \\ v_1 & v_2 \end{matrix}\right\},$$

$$\left\{ \begin{matrix} v_1 & v_3 \\ \bullet & \bullet \end{matrix} \right\}, \left\{ \begin{matrix} v_1 & v_4 \\ \bullet & \bullet \end{matrix} \right\}, \left\{ \begin{matrix} v_1 & v_5 \\ \bullet & \bullet \end{matrix} \right\}, \left\{ \begin{matrix} v_2 & v_3 \\ \bullet & \bullet \end{matrix} \right\},$$

$$\left\{ \begin{matrix} \bullet & \bullet \\ v_2 & v_4 \end{matrix} \right\}, \left\{ \begin{matrix} \bullet & \bullet \\ v_3 & v_4 \end{matrix} \right\}, \left\{ \begin{matrix} v_2 & v_5 \\ \bullet & \bullet \end{matrix} \right\}, \left\{ \begin{matrix} \bullet & \bullet \\ v_3 & v_5 \end{matrix} \right\},$$

$$\left\{ \begin{matrix} \bullet & \bullet \\ v_4 & v_5 \end{matrix} \right\}, \left\{ \begin{matrix} \bullet & \bullet & \bullet \\ v_1 & v_2 & v_3 \end{matrix} \right\}, \left\{ \begin{matrix} \bullet & \bullet & \bullet \\ v_1 & v_2 & v_4 \end{matrix} \right\}$$

$$\left\{ \begin{matrix} \bullet & \bullet & \bullet \\ v_1 & v_2 & v_5 \end{matrix} \right\}, \left\{ \begin{matrix} \bullet & \bullet & \bullet \\ v_2 & v_3 & v_4 \end{matrix} \right\}, \left\{ \begin{matrix} \bullet & \bullet & \bullet \\ v_2 & v_3 & v_5 \end{matrix} \right\},$$

$$\left\{ \begin{matrix} \bullet & \bullet & \bullet \\ v_3 & v_4 & v_5 \end{matrix} \right\}, \left\{ \begin{matrix} \bullet & \bullet & \bullet \\ v_1 & v_3 & v_4 \end{matrix} \right\}, \left\{ \begin{matrix} \bullet & \bullet & \bullet \\ v_1 & v_4 & v_5 \end{matrix} \right\},$$

$$\left\{ \begin{matrix} \bullet & \bullet & \bullet \\ v_1 & v_3 & v_5 \end{matrix} \right\}, \left\{ \begin{matrix} \bullet & \bullet & \bullet \\ v_2 & v_4 & v_5 \end{matrix} \right\}, \left\{ \begin{matrix} \bullet & \bullet & \bullet & \bullet \\ v_1 & v_2 & v_3 & v_4 \end{matrix} \right\},$$

$$\left\{ \begin{matrix} \bullet & \bullet & \bullet & \bullet \\ v_1 & v_2 & v_3 & v_5 \end{matrix} \right\}, \left\{ \begin{matrix} \bullet & \bullet & \bullet & \bullet \\ v_1 & v_2 & v_4 & v_5 \end{matrix} \right\},$$

$$\left\{ \begin{matrix} \bullet & \bullet & \bullet & \bullet \\ v_1 & v_3 & v_4 & v_5 \end{matrix} \right\}, \left\{ \begin{matrix} \bullet & \bullet & \bullet & \bullet \\ v_2 & v_3 & v_4 & v_5 \end{matrix} \right\},$$

$$\left\{ \begin{matrix} \bullet & \bullet & \bullet & \bullet & \bullet \\ v_1 & v_2 & v_3 & v_4 & v_5 \end{matrix} \right\}, \left\{ \begin{matrix} \bullet\!-\!\bullet & \bullet \\ v_1 & v_2 & v_3 \end{matrix} \right\}, \left\{ \begin{matrix} \bullet\!-\!\bullet & \bullet \\ v_1 & v_2 & v_4 \end{matrix} \right\},$$

$$\left\{ \begin{matrix} \bullet\!-\!\bullet & \bullet \\ v_1 & v_2 & v_5 \end{matrix} \right\}, \left\{ \begin{matrix} \bullet\!-\!\bullet & \bullet \\ v_1 & v_3 & v_2 \end{matrix} \right\}, \left\{ \begin{matrix} \bullet\!-\!\bullet & \bullet \\ v_1 & v_4 & v_3 \end{matrix} \right\},$$

$$\left\{ \begin{matrix} \bullet\!-\!\bullet & \bullet \\ v_1 & v_5 & v_2 \end{matrix} \right\}, \left\{ \begin{matrix} \bullet\!-\!\bullet & \bullet \\ v_1 & v_3 & v_4 \end{matrix} \right\}, \left\{ \begin{matrix} \bullet\!-\!\bullet & \bullet \\ v_1 & v_3 & v_5 \end{matrix} \right\},$$

$$\left\{ \begin{matrix} \bullet\!-\!\bullet & \bullet \\ v_1 & v_4 & v_2 \end{matrix} \right\}, \left\{ \begin{matrix} \bullet\!-\!\bullet & \bullet \\ v_1 & v_4 & v_5 \end{matrix} \right\}, \left\{ \begin{matrix} \bullet\!-\!\bullet & \bullet \\ v_1 & v_5 & v_3 \end{matrix} \right\},$$

$$\left\{ \overset{\bullet}{v_1} \!-\! \overset{\bullet}{v_5} \quad \overset{\bullet}{v_4} \right\}, \left\{ \overset{\bullet}{v_2} \!-\! \overset{\bullet}{v_3} \quad \overset{\bullet}{v_1} \right\}, \left\{ \overset{\bullet}{v_2} \!-\! \overset{\bullet}{v_4} \quad \overset{\bullet}{v_3} \right\},$$

$$\left\{ \overset{\bullet}{v_2} \!-\! \overset{\bullet}{v_5} \quad \overset{\bullet}{v_4} \right\}, \left\{ \overset{\bullet}{v_3} \!-\! \overset{\bullet}{v_4} \quad \overset{\bullet}{v_1} \right\}, \left\{ \overset{\bullet}{v_3} \!-\! \overset{\bullet}{v_4} \quad \overset{\bullet}{v_2} \right\},$$

$$\left\{ \overset{\bullet}{v_3} \!-\! \overset{\bullet}{v_4} \quad \overset{\bullet}{v_5} \right\}, \left\{ \overset{\bullet}{v_2} \!-\! \overset{\bullet}{v_3} \quad \overset{\bullet}{v_4} \right\}, \left\{ \overset{\bullet}{v_2} \!-\! \overset{\bullet}{v_3} \quad \overset{\bullet}{v_5} \right\},$$

$$\left\{ \overset{\bullet}{v_2} \!-\! \overset{\bullet}{v_4} \quad \overset{\bullet}{v_1} \right\}, \left\{ \overset{\bullet}{v_2} \!-\! \overset{\bullet}{v_4} \quad \overset{\bullet}{v_5} \right\}, \left\{ \overset{\bullet}{v_2} \!-\! \overset{\bullet}{v_5} \quad \overset{\bullet}{v_1} \right\},$$

$$\left\{ \overset{\bullet}{v_2} \!-\! \overset{\bullet}{v_5} \quad \overset{\bullet}{v_3} \right\}, \left\{ \overset{\bullet}{v_3} \!-\! \overset{\bullet}{v_5} \quad \overset{\bullet}{v_4} \right\}, \left\{ \overset{\bullet}{v_3} \!-\! \overset{\bullet}{v_5} \quad \overset{\bullet}{v_1} \right\},$$

$$\left\{ \overset{\bullet}{v_3} \!-\! \overset{\bullet}{v_5} \quad \overset{\bullet}{v_2} \right\}, \left\{ \overset{\bullet}{v_4} \!-\! \overset{\bullet}{v_5} \quad \overset{\bullet}{v_1} \right\}, \left\{ \overset{\bullet}{v_4} \!-\! \overset{\bullet}{v_5} \quad \overset{\bullet}{v_2} \right\},$$

$$\left\{ \overset{\bullet}{v_4} \!-\! \overset{\bullet}{v_5} \quad \overset{\bullet}{v_3} \right\}, \left\{ \overset{\bullet}{v_1} \!-\! \overset{\bullet}{v_2} \quad \overset{\bullet}{v_3} \quad \overset{\bullet}{v_4} \right\}, \left\{ \overset{\bullet}{v_1} \!-\! \overset{\bullet}{v_2} \quad \overset{\bullet}{v_3} \quad \overset{\bullet}{v_5} \right\},$$

$$\left\{ \overset{\bullet}{v_1} \!-\! \overset{\bullet}{v_2} \quad \overset{\bullet}{v_4} \quad \overset{\bullet}{v_5} \right\}, \left\{ \overset{\bullet}{v_1} \!-\! \overset{\bullet}{v_3} \quad \overset{\bullet}{v_2} \quad \overset{\bullet}{v_4} \right\},$$

$$\left\{ \overset{\bullet}{v_1} \!-\! \overset{\bullet}{v_3} \quad \overset{\bullet}{v_2} \quad \overset{\bullet}{v_5} \right\}, \left\{ \overset{\bullet}{v_1} \!-\! \overset{\bullet}{v_3} \quad \overset{\bullet}{v_4} \quad \overset{\bullet}{v_5} \right\},$$

$$\left\{ \overset{\bullet}{v_1} \!-\! \overset{\bullet}{v_4} \quad \overset{\bullet}{v_2} \quad \overset{\bullet}{v_5} \right\}, \left\{ \overset{\bullet}{v_1} \!-\! \overset{\bullet}{v_4} \quad \overset{\bullet}{v_3} \quad \overset{\bullet}{v_5} \right\},$$

$$\left\{ \overset{\bullet}{v_1} \!-\! \overset{\bullet}{v_4} \quad \overset{\bullet}{v_2} \quad \overset{\bullet}{v_3} \right\}, \left\{ \overset{\bullet}{v_1} \!-\! \overset{\bullet}{v_5} \quad \overset{\bullet}{v_2} \quad \overset{\bullet}{v_3} \right\}$$

$$\left\{ \overset{\bullet}{v_1} \!-\! \overset{\bullet}{v_5} \quad \overset{\bullet}{v_2} \quad \overset{\bullet}{v_4} \right\}, \left\{ \overset{\bullet}{v_1} \!-\! \overset{\bullet}{v_5} \quad \overset{\bullet}{v_3} \quad \overset{\bullet}{v_3} \right\},$$

$$\left\{ \overset{\bullet}{v_2} \!-\! \overset{\bullet}{v_3} \quad \overset{\bullet}{v_4} \quad \overset{\bullet}{v_5} \right\}, \left\{ \overset{\bullet}{v_2} \!-\! \overset{\bullet}{v_3} \quad \overset{\bullet}{v_1} \quad \overset{\bullet}{v_4} \right\},$$

$\left\{ \underset{v_2 \quad v_3}{\bullet\!-\!\bullet} \quad \underset{v_1}{\bullet} \quad \underset{v_5}{\bullet} \right\}, \left\{ \underset{v_2 \quad v_4}{\bullet\!-\!\bullet} \quad \underset{v_1}{\bullet} \quad \underset{v_3}{\bullet} \right\},$

$\left\{ \underset{v_2 \quad v_4}{\bullet\!-\!\bullet} \quad \underset{v_1}{\bullet} \quad \underset{v_5}{\bullet} \right\}, \left\{ \underset{v_2 \quad v_4}{\bullet\!-\!\bullet} \quad \underset{v_3}{\bullet} \quad \underset{v_5}{\bullet} \right\},$

$\left\{ \underset{v_2 \quad v_5}{\bullet\!-\!\bullet} \quad \underset{v_1}{\bullet} \quad \underset{v_3}{\bullet} \right\}, \left\{ \underset{v_2 \quad v_5}{\bullet\!-\!\bullet} \quad \underset{v_1}{\bullet} \quad \underset{v_4}{\bullet} \right\},$

$\left\{ \underset{v_2 \quad v_5}{\bullet\!-\!\bullet} \quad \underset{v_3}{\bullet} \quad \underset{v_4}{\bullet} \right\}, \left\{ \underset{v_3 \quad v_4}{\bullet\!-\!\bullet} \quad \underset{v_1}{\bullet} \quad \underset{v_2}{\bullet} \right\},$

$\left\{ \underset{v_3 \quad v_4}{\bullet\!-\!\bullet} \quad \underset{v_1}{\bullet} \quad \underset{v_5}{\bullet} \right\}, \left\{ \underset{v_3 \quad v_4}{\bullet\!-\!\bullet} \quad \underset{v_2}{\bullet} \quad \underset{v_5}{\bullet} \right\},$

$\left\{ \underset{v_4 \quad v_5}{\bullet\!-\!\bullet} \quad \underset{v_1}{\bullet} \quad \underset{v_2}{\bullet} \right\}, \left\{ \underset{v_4 \quad v_5}{\bullet\!-\!\bullet} \quad \underset{v_1}{\bullet} \quad \underset{v_3}{\bullet} \right\},$

$\left\{ \underset{v_4 \quad v_5}{\bullet\!-\!\bullet} \quad \underset{v_2}{\bullet} \quad \underset{v_3}{\bullet} \right\}, \left\{ \underset{v_3 \quad v_5}{\bullet\!-\!\bullet} \quad \underset{v_1}{\bullet} \quad \underset{v_2}{\bullet} \right\},$

$\left\{ \underset{v_3 \quad v_5}{\bullet\!-\!\bullet} \quad \underset{v_1}{\bullet} \quad \underset{v_4}{\bullet} \right\}, \left\{ \underset{v_3 \quad v_5}{\bullet\!-\!\bullet} \quad \underset{v_2}{\bullet} \quad \underset{v_4}{\bullet} \right\},$

$\left\{ \underset{v_1 \quad v_2}{\bullet\!-\!\bullet} \quad \underset{v_3 \quad v_4}{\bullet\!-\!\bullet} \quad \underset{v_5}{\bullet} \right\} \left\{ \underset{v_1 \quad v_2}{\bullet\!-\!\bullet} \quad \underset{v_4 \quad v_5}{\bullet\!-\!\bullet} \quad \underset{v_3}{\bullet} \right\},$

$\left\{ \underset{v_1 \quad v_2}{\bullet\!-\!\bullet} \quad \underset{v_3 \quad v_5}{\bullet\!-\!\bullet} \quad \underset{v_4}{\bullet} \right\}, \left\{ \underset{v_1 \quad v_3}{\bullet\!-\!\bullet} \quad \underset{v_2 \quad v_4}{\bullet\!-\!\bullet} \quad \underset{v_5}{\bullet} \right\},$

$\left\{ \underset{v_1 \quad v_3}{\bullet\!-\!\bullet} \quad \underset{v_2 \quad v_5}{\bullet\!-\!\bullet} \quad \underset{v_4}{\bullet} \right\} \left\{ \underset{v_1 \quad v_3}{\bullet\!-\!\bullet} \quad \underset{v_4 \quad v_5}{\bullet\!-\!\bullet} \quad \underset{v_2}{\bullet} \right\},$

$\left\{ \underset{v_1 \quad v_4}{\bullet\!-\!\bullet} \quad \underset{v_2 \quad v_3}{\bullet\!-\!\bullet} \quad \underset{v_5}{\bullet} \right\}, \left\{ \underset{v_1 \quad v_4}{\bullet\!-\!\bullet} \quad \underset{v_2 \quad v_5}{\bullet\!-\!\bullet} \quad \underset{v_3}{\bullet} \right\},$

$\left\{ \underset{v_1 \quad v_4}{\bullet\!-\!\bullet} \quad \underset{v_3 \quad v_5}{\bullet\!-\!\bullet} \quad \underset{v_5}{\bullet} \right\}, \left\{ \underset{v_1 \quad v_5}{\bullet\!-\!\bullet} \quad \underset{v_2 \quad v_3}{\bullet\!-\!\bullet} \quad \underset{v_4}{\bullet} \right\},$

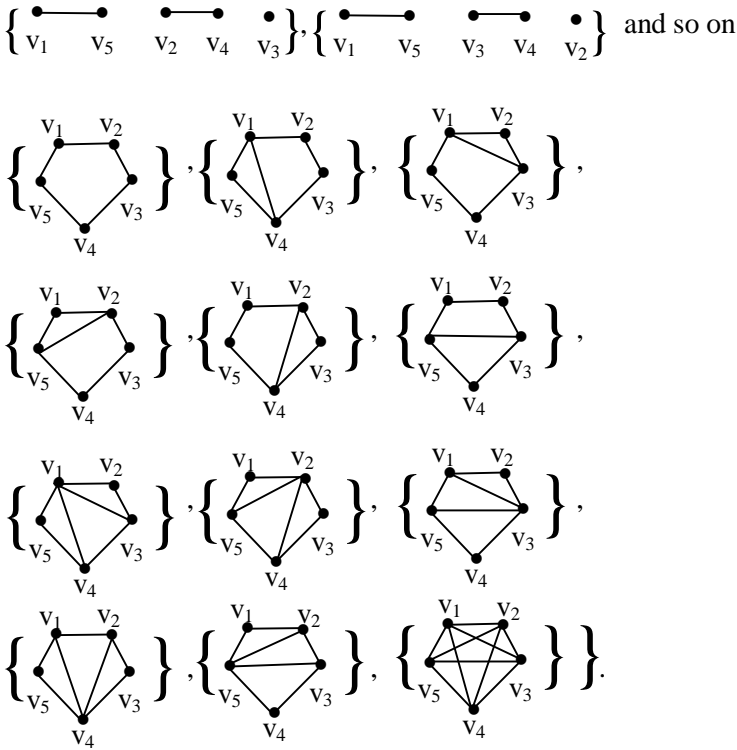

Figure 4.7

It is left for the reader as an open problem to find the number of elements in S(G).

Thus we leave it as a open conjecture.

**Conjecture 4.1:** Let V = {$v_1$, ..., $v_n$}, be n vertices. Let S(G) be the collection of all graphs constructed using V including {$\phi$} and the complete graph with n vertices. Find the number of elements in S(G).

Now we reformulate or redefine S(G) as follows.

Let G be any complete graph.

S(G) = {Collection of all subgraphs of G together with φ and G}.

If G has n vertices find the cardinality of S(G).

Thus we see S(G) can be defined in two ways and both are equivalent.

Now we can relax the condition for any graph G. Let G be any graph

S(G) = {Collection of all subgraphs of G including φ and G}.

We give examples of this situation.

***Example 4.1:*** Let G be a graph with 4 vertices.

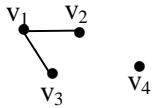

Figure 4.8

$$S(G) = \{\phi, \left\{ {\bullet \atop v_1} \right\}, \left\{ {\bullet \atop v_2} \right\}, \left\{ {\bullet \atop v_3} \right\}, \left\{ {\bullet \atop v_4} \right\}, \left\{ {\bullet \atop v_1} \quad {\bullet \atop v_2} \right\},$$

$$\left\{ {\bullet \atop v_1} \quad {\bullet \atop v_3} \right\}, \left\{ {\bullet \atop v_1} \quad {\bullet \atop v_4} \right\}\left\{ {\bullet \atop v_2} \quad {\bullet \atop v_3} \right\}, \left\{ {\bullet \atop v_2} \quad {\bullet \atop v_4} \right\},$$

$$\left\{ {\bullet \atop v_2} \quad {\bullet \atop v_3} \right\}, \left\{ {\bullet\!\!-\!\!\bullet \atop v_1 \quad v_2} \right\}, \left\{ {\bullet\!\!-\!\!\bullet \atop v_1 \quad v_3} \right\},$$

$$\left\{ {\bullet \atop v_1} \quad {\bullet \atop v_2} \quad {\bullet \atop v_3} \right\}, \left\{ {\bullet \atop v_1} \quad {\bullet \atop v_2} \quad {\bullet \atop v_4} \right\}, \left\{ {\bullet \atop v_1} \quad {\bullet \atop v_3} \quad {\bullet \atop v_4} \right\},$$

$$\left\{ {\bullet\!\!-\!\!\bullet \atop v_1 \quad v_3} \quad {\bullet \atop v_4} \right\}, \left\{ {\bullet \atop v_2} \quad {\bullet \atop v_3} \quad {\bullet \atop v_4} \right\}, \left\{ {\bullet\!\!-\!\!\bullet \atop v_1 \quad v_2} \quad {\bullet \atop v_4} \right\},$$

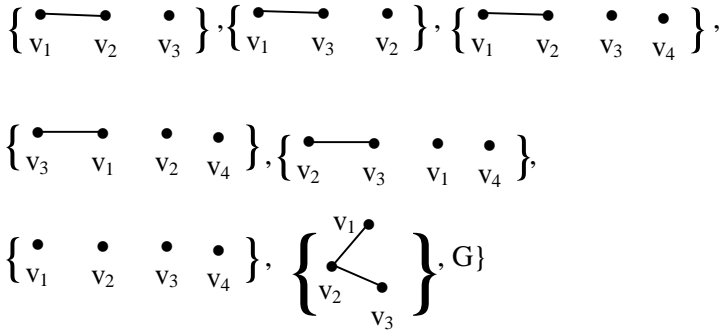

Figure 4.9

Number of elements in S(G) is 26.

***Example 4.2:*** Let G be the graph 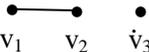        .

Figure 4.10

To find S(G) = {Collection of all subgraphs of G}

$$= \{\phi, \left\{\begin{smallmatrix} \bullet \\ v_1 \end{smallmatrix}\right\}, \left\{\begin{smallmatrix} \bullet \\ v_2 \end{smallmatrix}\right\}, \left\{\begin{smallmatrix} \bullet \\ v_3 \end{smallmatrix}\right\}, \left\{\begin{smallmatrix} \bullet & \bullet \\ v_1 & v_2 \end{smallmatrix}\right\}, \left\{\begin{smallmatrix} \bullet & \bullet \\ v_1 & v_3 \end{smallmatrix}\right\}, \left\{\begin{smallmatrix} \bullet & \bullet \\ v_2 & v_3 \end{smallmatrix}\right\},$$

$$\left\{\begin{smallmatrix} \bullet & \bullet & \bullet \\ v_1 & v_2 & v_3 \end{smallmatrix}\right\}, \left\{\begin{smallmatrix} \bullet\!\!-\!\!\bullet \\ v_1 & v_2 \end{smallmatrix}\right\}, G\}.$$

Clearly o(S(G)) = 10.

Now we define special lattice subgraphs of a graph G is a lattice drawn using all the subgraphs of G together with G and the empty set.

***Example 4.3:*** Let G be the graph $\left\{\begin{smallmatrix} \bullet\!\!-\!\!\bullet & \bullet\!\!-\!\!\bullet \\ v_1 & v_2 & v_3 & v_4 \end{smallmatrix}\right\}$

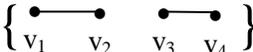

Figure 4.11

The subgraphs of G are $\{\phi, \left\{\begin{smallmatrix} \bullet \\ v_1 \end{smallmatrix}\right\} \ \left\{\begin{smallmatrix} \bullet \\ v_2 \end{smallmatrix}\right\} \ \left\{\begin{smallmatrix} \bullet \\ v_3 \end{smallmatrix}\right\}, \left\{\begin{smallmatrix} \bullet \\ v_4 \end{smallmatrix}\right\},$

$$\left\{ \begin{matrix} \bullet & \bullet \\ v_1 & v_2 \end{matrix} \right\}, \left\{ \begin{matrix} \bullet & \bullet \\ v_1 & v_3 \end{matrix} \right\}, \left\{ \begin{matrix} \bullet & \bullet \\ v_1 & v_4 \end{matrix} \right\}, \left\{ \begin{matrix} \bullet & \bullet \\ v_2 & v_3 \end{matrix} \right\},$$

$$\left\{ \begin{matrix} \bullet & \bullet \\ v_2 & v_4 \end{matrix} \right\}, \left\{ \begin{matrix} \bullet & \bullet \\ v_3 & v_4 \end{matrix} \right\}, \left\{ \begin{matrix} \bullet & \bullet & \bullet \\ v_1 & v_2 & v_3 \end{matrix} \right\}, \left\{ \begin{matrix} \bullet & \bullet & \bullet \\ v_1 & v_2 & v_4 \end{matrix} \right\},$$

$$\left\{ \begin{matrix} \bullet & \bullet & \bullet \\ v_1 & v_3 & v_4 \end{matrix} \right\}, \left\{ \begin{matrix} \bullet & \bullet & \bullet \\ v_2 & v_3 & v_4 \end{matrix} \right\}, \left\{ \begin{matrix} \bullet & \bullet & \bullet & \bullet \\ v_1 & v_2 & v_3 & v_4 \end{matrix} \right\},$$

$$\left\{ \begin{matrix} \bullet\!\!-\!\!\bullet \\ v_1 \quad v_2 \end{matrix} \right\}, \left\{ \begin{matrix} \bullet\!\!-\!\!\bullet \\ v_3 \quad v_4 \end{matrix} \right\}, \left\{ \begin{matrix} \bullet\!\!-\!\!\bullet & \bullet \\ v_1 \quad v_2 & v_3 \end{matrix} \right\}, \left\{ \begin{matrix} \bullet\!\!-\!\!\bullet & \bullet \\ v_1 \quad v_2 & v_4 \end{matrix} \right\},$$

$$\left\{ \begin{matrix} \bullet\!\!-\!\!\bullet & \bullet & \bullet \\ v_1 \quad v_2 & v_3 & v_4 \end{matrix} \right\}, \left\{ \begin{matrix} \bullet\!\!-\!\!\bullet & \bullet \\ v_3 \quad v_4 & v_1 \end{matrix} \right\}, \left\{ \begin{matrix} \bullet\!\!-\!\!\bullet & \bullet \\ v_3 \quad v_4 & v_2 \end{matrix} \right\},$$

$$\left\{ \begin{matrix} \bullet\!\!-\!\!\bullet & \bullet & \bullet \\ v_3 \quad v_4 & v_1 & v_2 \end{matrix} \right\}, \left\{ \begin{matrix} \bullet\!\!-\!\!\bullet & \bullet\!\!-\!\!\bullet \\ v_1 \quad v_2 & v_3 \quad v_4 \end{matrix} \right\} \right\}.$$

Clearly  o(S(G)) = 25.

We now give the lattice graph of G.

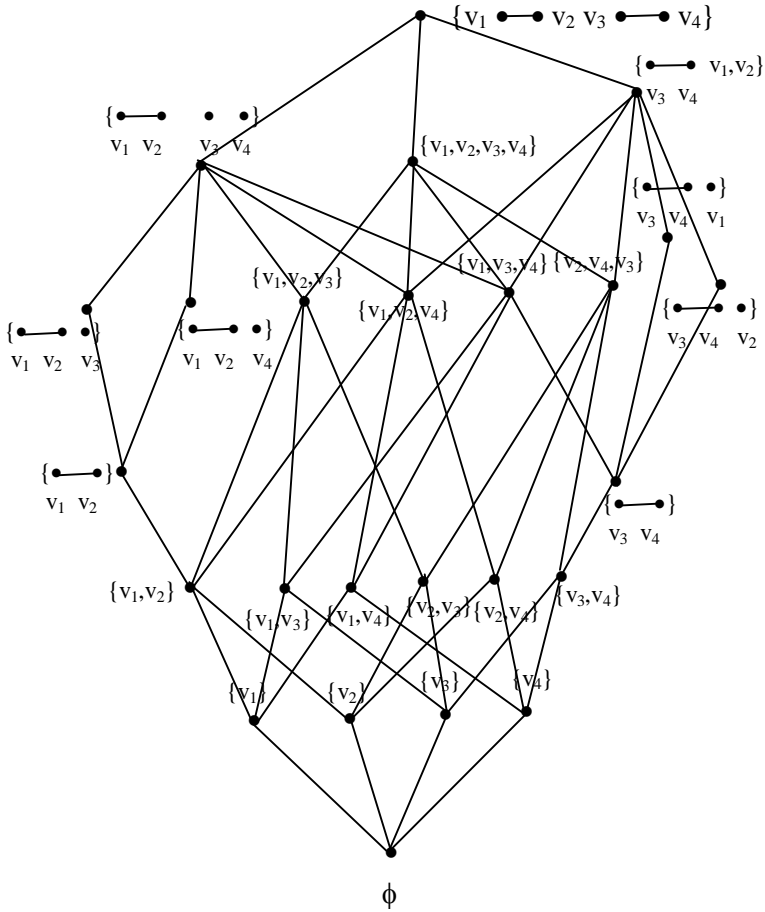

Figure 4.12

Now we give the special lattice graph of S(G) where

G is  •——•  •  given in example.
     $v_1$  $v_2$  $v_3$

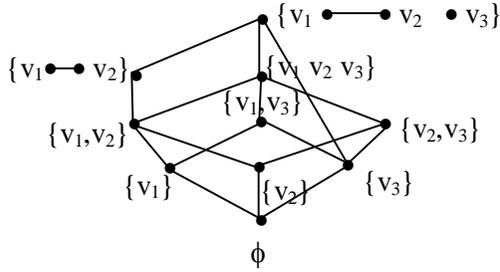

Figure 4.13

***Example 4.4:*** Now we give the special lattice graph of

S(G) where G is 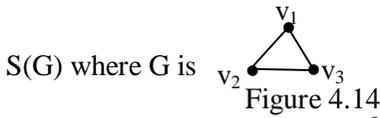

Figure 4.14

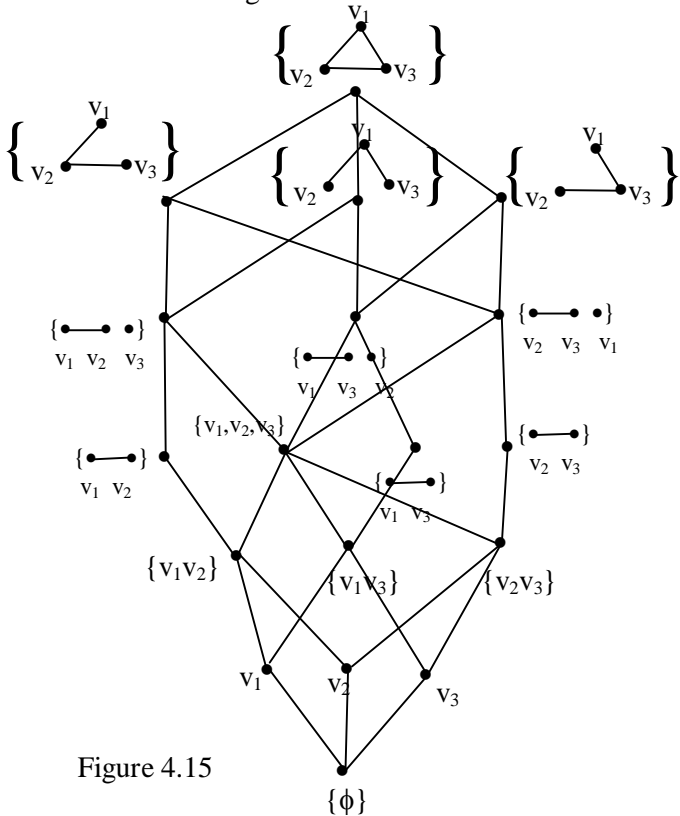

Figure 4.15

described in example.

Clearly the subgraphs do not form a Boolean algebra. It is pertinent to mention the following interesting problems are discussed by the following examples.

***Example 4.5:*** Let G be a graph $v_1 \bullet \quad \bullet v_2$.

<div align="center">Figure 4.16</div>

$$S(G) = \{ \phi, \{ \bullet v_1 \} \quad \{ \bullet v_2 \} \left\{ \begin{matrix} \bullet & \bullet \\ v_1 & v_2 \end{matrix} \right\} \}.$$

The special lattice graph is

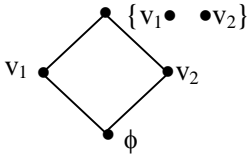

<div align="center">Figure 4.17</div>

is a Boolean algebra of order four.

***Example 4.6:*** Let G be the single vertex graph $v_1$.

$$S(G) = \{ \phi, \left\{ \begin{matrix} \bullet \\ v_1 \end{matrix} \right\} \}$$

The special lattice graph is a Boolean algebra of order two.

***Example 4.7:*** Let G be the graph with no edges just three vertices

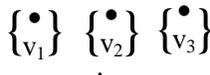

Then $S(G) = \{ \phi, G, \left\{ \begin{matrix} \bullet \\ v_1 \end{matrix} \right\} \quad \left\{ \begin{matrix} \bullet \\ v_2 \end{matrix} \right\} \quad \left\{ \begin{matrix} \bullet \\ v_3 \end{matrix} \right\}, \left\{ \begin{matrix} \bullet & \bullet \\ v_1 & v_2 \end{matrix} \right\},$

$$\left\{ \begin{matrix} \bullet & \bullet \\ v_1 & v_3 \end{matrix} \right\}, \left\{ \begin{matrix} \bullet & \bullet \\ v_2 & v_3 \end{matrix} \right\} \right\}.$$

The special lattice subgraph is as follows:

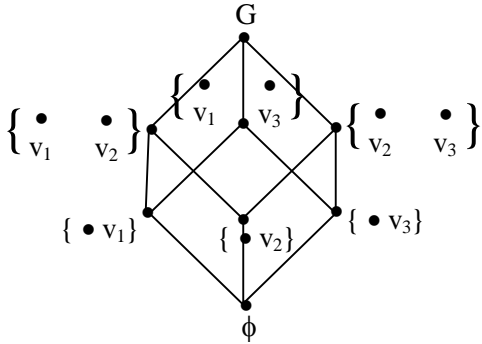

Figure 4.18

Clearly the special lattice subgraph is a Boolean algebra of order 8.

In view of this we state the following theorem the proof of which is left as an exercise to the reader.

**THEOREM 4.1:** *Let $G = \{v_1, v_2, \ldots, v_n\}$ ; $\bullet v_1 \bullet v_2 \ldots \bullet v_n$ be a n vertex graph with no edges. Then $S(G)$ the collection of all subgraphs of G with G and $\phi$ is a Boolean algebra of order $2^n$.*

Now we see in general the special lattice subgraph of a graph G is not a Boolean algebra.

***Example 4.8:*** Let G be the graph 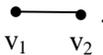 .

Figure 4.19

The subgraphs of G are

$$S(G) = \{\phi, G, \begin{Bmatrix} \bullet \\ v_1 \end{Bmatrix} \begin{Bmatrix} \bullet \\ v_2 \end{Bmatrix} \quad \{\overset{\bullet}{v_1} \rule[0.3em]{1.2em}{0.4pt} \overset{\bullet}{v_2}\} \begin{Bmatrix} \bullet & \bullet \\ v_1 & v_2 \end{Bmatrix}\}.$$

The special lattice of subgraph is as follows:

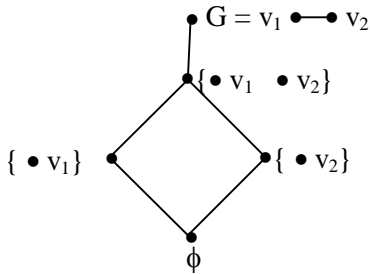

Figure 4.20

Clearly S(G) is not a Boolean algebra but has a Boolean algebra of order 4 as a sublattice of the special lattice of subgraphs.

In view of this we give the following.

**THEOREM 4.2:** *Let G be a graph which has at least one vertex S(G) the collection of subgraphs of G together with $\phi$ and G. The special lattice subgraph of G is not a Boolean algebra.*

The proof is obvious hence left as an exercise to the reader.

Next we give yet another theorem which guarantees of a sublattice of a special graph which is a Boolean algebra.

**THEOREM 4.3:** *Let G be a graph with n vertices and atleast an edge. S(G) the subgraphs of G. The special lattice of S(G) has a sublattice which a Boolean algebra of order $2^n$.*

This proof is also direct and hence left as an exercise to the reader.

We will give some more examples.

***Example 4.9:*** Let G be a graph which is as follows.

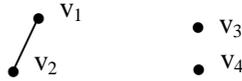

Figure 4.21

Let S(G) = {collection of all subgraph of G together with G and φ}

$$= \{\phi, \left\{\begin{smallmatrix}\bullet\\v_1\end{smallmatrix}\right\}, \left\{\begin{smallmatrix}\bullet\\v_2\end{smallmatrix}\right\}, \left\{\begin{smallmatrix}\bullet\\v_3\end{smallmatrix}\right\}, \left\{\begin{smallmatrix}\bullet\\v_4\end{smallmatrix}\right\}, \left\{\begin{smallmatrix}\bullet & \bullet\\v_1 & v_2\end{smallmatrix}\right\},$$

$$\left\{\begin{smallmatrix}\bullet & \bullet\\v_1 & v_3\end{smallmatrix}\right\}, \left\{\begin{smallmatrix}\bullet & \bullet\\v_1 & v_4\end{smallmatrix}\right\}, \left\{\begin{smallmatrix}\bullet & \bullet\\v_2 & v_3\end{smallmatrix}\right\}, \left\{\begin{smallmatrix}\bullet & \bullet\\v_2 & v_4\end{smallmatrix}\right\},$$

$$\left\{\begin{smallmatrix}\bullet & \bullet\\v_3 & v_4\end{smallmatrix}\right\}, \left\{\begin{smallmatrix}\bullet\!\!-\!\!\bullet\\v_1 & v_2\end{smallmatrix}\right\}, \left\{\begin{smallmatrix}\bullet & \bullet & \bullet\\v_1 & v_2 & v_3\end{smallmatrix}\right\}, \left\{\begin{smallmatrix}\bullet & \bullet & \bullet\\v_1 & v_2 & v_4\end{smallmatrix}\right\},$$

$$\left\{\begin{smallmatrix}\bullet\!\!-\!\!\bullet & \bullet\\v_1 & v_2 & v_3\end{smallmatrix}\right\}, \left\{\begin{smallmatrix}\bullet\!\!-\!\!\bullet & \bullet\\v_1 & v_2 & v_4\end{smallmatrix}\right\}, \left\{\begin{smallmatrix}\bullet & \bullet & \bullet\\v_1 & v_3 & v_4\end{smallmatrix}\right\},$$

$$\left\{\begin{smallmatrix}\bullet & \bullet & \bullet\\v_2 & v_3 & v_4\end{smallmatrix}\right\}, \left\{\begin{smallmatrix}\bullet & \bullet & \bullet & \bullet\\v_1 & v_2 & v_3 & v_4\end{smallmatrix}\right\}, \left\{\begin{smallmatrix}\bullet\!\!-\!\!\bullet & \bullet & \bullet\\v_1 & v_2 & v_3 & v_4\end{smallmatrix}\right\}\}$$

$$= G.$$

The special lattice subgraph of G is as follows.

Clearly o(S(G))= 20.

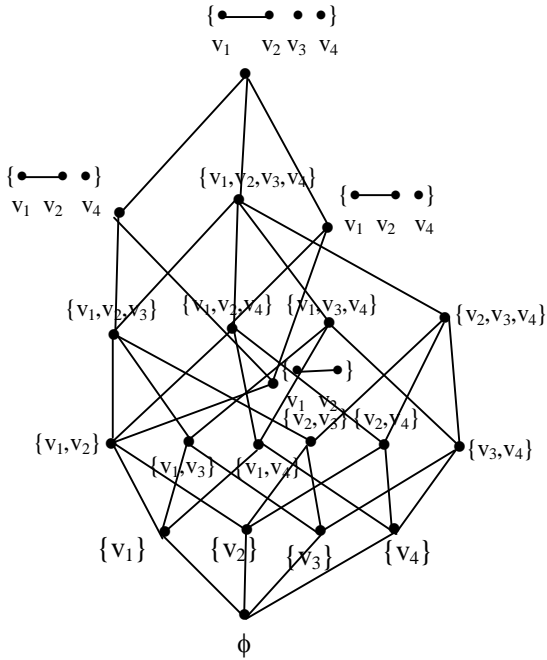

Figure 4.22

We see this has a sublattice which is a Boolean algebra of order $2^4$ given by

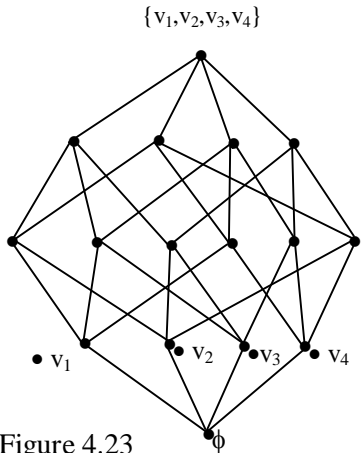

Figure 4.23

**DEFINITION 4.1:** *Let G be a graph with n vertices and m edges. S(G) = {collection of all subgraphs of G and $\phi$}; we define S(G) as the power subgraph set of G.*

We have seen several examples of the power subgraph set of G.

We give an important result related with this power subgraph set.

**THEOREM 4.4:** *Let G be a graph with n vertices and m edges (m $\geq$ 1). S(G) the power subgraph set of G. S(G) has a lattice structure with o(S(G)) points and the lattice of S(G) denoted by $L_{S(G)}$ contains a Boolean algebra of order $2^n$ or if X = {$v_1$, …, $v_n$} then P(X) $\subseteq$ S(G).*

This proof is also direct hence left as an exercise to the reader.

Now we see this power subgraph set has several interesting properties which we enlist in the following as results.

S(G) denotes the power subgraph set of a graph G.

**Result 4.1:** G is the greatest element of S(G) and $\phi$ is the least element of G.

**Result 4.2:** If G is a graph which has atleast one edge S(G) is not a Boolean algebra.

**Result 4.3:** On S(G) we have both operations '$\cup$' and '$\cap$' defined which makes S(G) a lattice.

**Result 4.4:** S(G) is a partially ordered set.

**Result 4.5:** If p and q are two subgraphs of G which is in S(G). We know p $\cap$ q and p $\cup$ q are in S(G).

**Result 4.6:** S(G) can also in a way said to be all minors of G.

**Result 4.7:** $\phi$ is the complement of G.

The following questions are of interest and are listed as open problems.

**Problem 4.1:** Given G is a graph with n vertices and p edges. Find the cardinality of S(G).

**Problem 4.2:** Can the lattice associated with S(G) be distributive?

***Example 4.10:*** Let G be a graph given below.

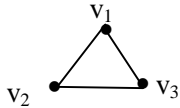

Figure 4.24

To find S(G); S(G) = {$\phi$, G, $\left\{\begin{smallmatrix}\bullet\\v_1\end{smallmatrix}\right\}$, $\left\{\begin{smallmatrix}\bullet\\v_2\end{smallmatrix}\right\}$, $\left\{\begin{smallmatrix}\bullet\\v_3\end{smallmatrix}\right\}$,

$\left\{\begin{smallmatrix}\bullet&&\bullet\\v_1&&v_2\end{smallmatrix}\right\}$, $\left\{\begin{smallmatrix}\bullet&&\bullet\\v_1&&v_3\end{smallmatrix}\right\}$, $\left\{\begin{smallmatrix}\bullet&&\bullet\\v_2&&v_3\end{smallmatrix}\right\}$, $\left\{\begin{smallmatrix}\bullet&&\bullet&&\bullet\\v_1&&v_2&&v_3\end{smallmatrix}\right\}$,

$\left\{\begin{smallmatrix}\bullet\!\!-\!\!\bullet&&\bullet\\v_1\ \ \ v_2&&v_3\end{smallmatrix}\right\}$, $\left\{\begin{smallmatrix}\bullet\!\!-\!\!\bullet&&\bullet\\v_2\ \ \ v_3&&v_1\end{smallmatrix}\right\}$, $\left\{\begin{smallmatrix}\bullet\!\!-\!\!\bullet&&\bullet\\v_1\ \ \ v_3&&v_2\end{smallmatrix}\right\}$,

$\left\{\begin{smallmatrix}\bullet\!\!-\!\!\bullet\\v_1\ \ \ v_2\end{smallmatrix}\right\}$, $\left\{\begin{smallmatrix}\bullet\!\!-\!\!\bullet\\v_1\ \ \ v_3\end{smallmatrix}\right\}$, $\left\{\begin{smallmatrix}\bullet\!\!-\!\!\bullet\\v_2\ \ \ v_3\end{smallmatrix}\right\}$ } } be the power subgraph set of G.  o(S(G)) = 13.

The lattice associated with S(G) is as follows:

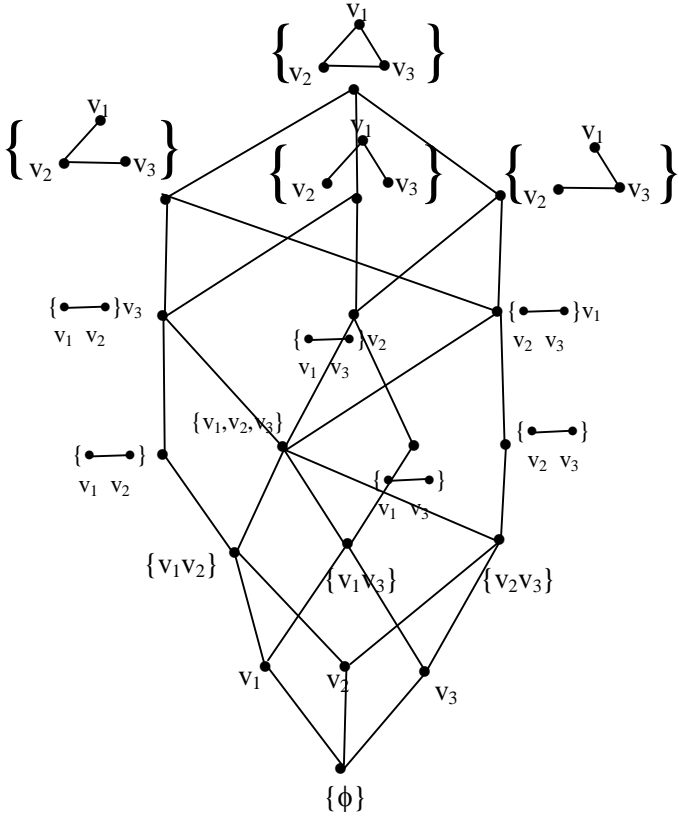

Figure 4.25

**Example 4.11:** Let G be the graph

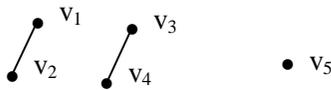

Figure 4.26

S(G) denote the collection of all subgraphs of G.

If $H_1 =$  is a subgraph of G

Figure 4.27

we see $S(H_1) \subseteq S(G)$. $H_2 =$ 

Figure 4.28

be a subgraph of G.

$S(H_2) \subseteq S(G)$ and clearly $S(H_1)$ is not isomorphic to $S(H_2)$.

Consider $H_3 =$ 

Figure 4.29

a subgraph of G.

$S(H_3) \subseteq S(G)$ and we see as special lattice subgraphs we see $S(H_3) \cong S(H_1)$.

We have several properties about subgraphs as special sublattice subgraphs.

***Example 4.12:*** Let G = 

Figure 4.30

be a graph.

S(G) = {Collection of all subgraphs of G together with φ and G}. One can have the special lattice subgraphs of S(G).

H = • • is a subgraph of G.
     $v_1$   $v_2$

Now S(H) = {Collection of all subgraphs of H with H and φ}

$$= \{\phi, H, \left\{\begin{matrix} \bullet \\ v_1 \end{matrix}\right\}, \left\{\begin{matrix} \bullet \\ v_2 \end{matrix}\right\}, \left\{\begin{matrix} \bullet & \bullet \\ v_1 & v_2 \end{matrix}\right\}\} \subseteq S(G).$$

The special sublattice subgraph S(H) of S(G) is the special lattice of subgraphs.

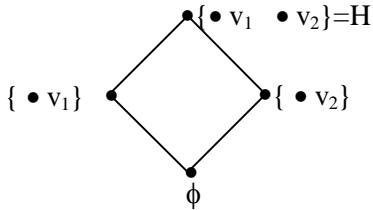

Figure 4.31

S(H) is a Boolean algebra of order four.

***Example 4.13:*** Let G =

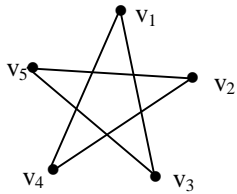

Figure 4.32

be a graph. To find the collection of subgraphs of G with G and φ.

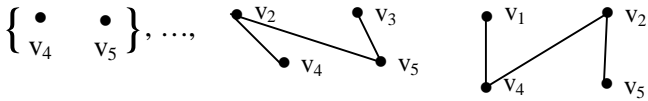

and so on}.

Now we see S(G) is closed under '$\cup$' and '$\cap$'.

Now S(G) can be given a topological we call S(G) as the special topology subgraphs of a graph.

Given any graph G we define S(G) as the special topological subgraphs. Thus even if vertices are same and edges are different we get distinct graphs.

We get for a given set of n-vertices we have many special topological subgraphs depending on the edges.

Study of these new class of topologies using subgraphs happens to be one of the innovations both on graph theory and topologies.

It is left as an open problem which is as follows.

**Problem 4.3:** Let G be a graph with n vertices. If G has p edges what is S(G)? If G has m edges what is S(G) (p>m)?

Compare them.

***Example 4.14:*** Let G be the graph

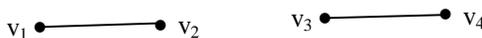

Figure 4.33

and H be the graph.

$v_1$ •————————• $v_2$    $v_3$ •          • $v_4$

Figure 4.34

To find S(G) and S(H).

$$S(H)= \{\phi, \left\{\begin{smallmatrix}\bullet\\v_1\end{smallmatrix}\right\}, \left\{\begin{smallmatrix}\bullet\\v_2\end{smallmatrix}\right\}, \left\{\begin{smallmatrix}\bullet\\v_3\end{smallmatrix}\right\}, \left\{\begin{smallmatrix}\bullet\\v_4\end{smallmatrix}\right\}, \left\{\begin{smallmatrix}\bullet & \bullet\\v_1 & v_2\end{smallmatrix}\right\},$$

$$\left\{\begin{smallmatrix}\bullet & \bullet\\v_1 & v_3\end{smallmatrix}\right\}, \left\{\begin{smallmatrix}\bullet & \bullet\\v_1 & v_4\end{smallmatrix}\right\}, \left\{\begin{smallmatrix}\bullet & \bullet\\v_2 & v_3\end{smallmatrix}\right\}, \left\{\begin{smallmatrix}\bullet & \bullet\\v_2 & v_4\end{smallmatrix}\right\},$$

$$\left\{\begin{smallmatrix}\bullet & \bullet\\v_3 & v_4\end{smallmatrix}\right\}, \left\{\begin{smallmatrix}\bullet\!-\!\!-\!\bullet\\v_1 \quad v_2\end{smallmatrix}\right\}, \left\{\begin{smallmatrix}\bullet & \bullet & \bullet\\v_1 & v_2 & v_3\end{smallmatrix}\right\}, \left\{\begin{smallmatrix}\bullet & \bullet & \bullet\\v_1 & v_2 & v_4\end{smallmatrix}\right\},$$

$$\left\{\begin{smallmatrix}\bullet\!-\!\!-\!\bullet & \bullet\\v_1 \quad v_2 & v_3\end{smallmatrix}\right\}, \left\{\begin{smallmatrix}\bullet\!-\!\!-\!\bullet & \bullet\\v_1 \quad v_2 & v_4\end{smallmatrix}\right\}, \left\{\begin{smallmatrix}\bullet & \bullet & \bullet\\v_1 & v_3 & v_4\end{smallmatrix}\right\},$$

$$\left\{\begin{smallmatrix}\bullet & \bullet & \bullet\\v_2 & v_3 & v_4\end{smallmatrix}\right\}, \left\{\begin{smallmatrix}\bullet & \bullet & \bullet & \bullet\\v_1 & v_2 & v_3 & v_4\end{smallmatrix}\right\}, \left\{\begin{smallmatrix}\bullet\!-\!\!-\!\bullet & \bullet & \bullet\\v_1 \quad v_2 & v_3 & v_4\end{smallmatrix}\right\}$$

= H}.  o(S(H)) = 20.

$$S(G) = \{\phi, \left\{\begin{smallmatrix}\bullet\\v_1\end{smallmatrix}\right\}, \left\{\begin{smallmatrix}\bullet\\v_2\end{smallmatrix}\right\}, \left\{\begin{smallmatrix}\bullet\\v_3\end{smallmatrix}\right\}, \left\{\begin{smallmatrix}\bullet\\v_4\end{smallmatrix}\right\}, \left\{\begin{smallmatrix}\bullet & \bullet\\v_1 & v_2\end{smallmatrix}\right\},$$

$$\left\{\begin{smallmatrix}\bullet & \bullet\\v_1 & v_3\end{smallmatrix}\right\}, \left\{\begin{smallmatrix}\bullet & \bullet\\v_1 & v_4\end{smallmatrix}\right\}, \left\{\begin{smallmatrix}\bullet & \bullet\\v_2 & v_3\end{smallmatrix}\right\}, \left\{\begin{smallmatrix}\bullet & \bullet\\v_2 & v_4\end{smallmatrix}\right\},$$

$$\left\{\begin{smallmatrix}\bullet & \bullet\\v_3 & v_4\end{smallmatrix}\right\}, \left\{\begin{smallmatrix}\bullet\!-\!\!-\!\bullet\\v_1 \quad v_2\end{smallmatrix}\right\}, \left\{\begin{smallmatrix}\bullet\!-\!\!-\!\bullet\\v_3 \quad v_4\end{smallmatrix}\right\}, \left\{\begin{smallmatrix}\bullet & \bullet & \bullet\\v_1 & v_2 & v_3\end{smallmatrix}\right\},$$

$$\left\{\begin{smallmatrix}\bullet & \bullet & \bullet\\v_1 & v_2 & v_4\end{smallmatrix}\right\}, \left\{\begin{smallmatrix}\bullet & \bullet & \bullet\\v_1 & v_3 & v_4\end{smallmatrix}\right\}, \left\{\begin{smallmatrix}\bullet & \bullet & \bullet\\v_2 & v_3 & v_4\end{smallmatrix}\right\},$$

$$\left\{\begin{smallmatrix}\bullet\!-\!\!-\!\bullet & \bullet\\v_1 \quad v_2 & v_3\end{smallmatrix}\right\}, \left\{\begin{smallmatrix}\bullet\!-\!\!-\!\bullet & \bullet\\v_1 \quad v_2 & v_4\end{smallmatrix}\right\}, \left\{\begin{smallmatrix}\bullet & \bullet\!-\!\!-\!\bullet\\v_1 & v_3 \quad v_4\end{smallmatrix}\right\},$$

$$\left\{ \begin{array}{cccc} \bullet & \bullet & \bullet & \bullet \\ v_1 & v_2 & v_3 & v_4 \end{array} \right\}, \left\{ \begin{array}{cccc} \bullet\!\!-\!\!\bullet & & \bullet & \bullet \\ v_1 & v_2 & v_3 & v_4 \end{array} \right\}$$

$$\left\{ \begin{array}{cccc} \bullet & \bullet & \bullet\!\!-\!\!\bullet \\ v_1 & v_2 & v_3 & v_4 \end{array} \right\}, \left\{ \begin{array}{cccc} \bullet\!\!-\!\!\bullet & & \bullet\!\!-\!\!\bullet \\ v_1 & v_2 & v_3 & v_4 \end{array} \right\}. \text{ o(SG)} = 25.$$

We see S(H) ⊆ S(G). Thus a graph with 4 vertices and two edges has 25 subgraphs and a graph H with 4 vertices and one edge has 20 subgraphs.

Thus addition of edges gives more subgraphs. Further if we have only 4 vertices graph K the S(K) will have $2^4 = 16$ subgraphs.

We see the special topological subgraphs depend on the number of edges of the graph.

**THEOREM 4.5:** *All special lattice of graphs G are Smarandache lattices where G has more then or equal to two vertices.*

Proof follows from the simple fact if G as a graph with n vertices.

The special lattice of subgraphs associated with S(G) has a sublattice which is a Boolean algebra of order $2^n$ ($n \geq 2$). Hence these lattices are Smarandache lattices.

It is important to notice that each subgraph of the graph G can be associated with a special graph topological space.

By this way we get a class of new special topological subgraph spaces which is very different from the usual topological spaces on subsets of a set.

Now if G has finite number of vertices certainly the special subgraph topological spaces are finite.

Further depending on the graph the special topological subgraph spaces will be different.

It is left for the reader to study the special topological subgraph spaces of a connected graph, tree, Turan graph, Wagner graph, k-partile graph and stars.

However we illustrate for a few of them the rest is left as an exercise to the reader.

**Example 4.15:** Let G be the Turan graph.

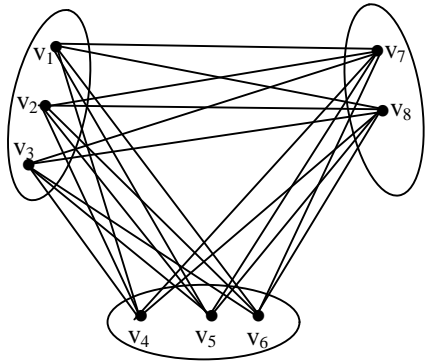

Figure 4.35

S(G) = {Collection of all subgraph of G with G and $\phi$}.

The interested reader is expected to find o(S(G)).

We call the special topological subgraph space as the Turan special topological subgraph space or special Turan topological subgraph spaces.

***Example 4.16:*** Let G be a graph

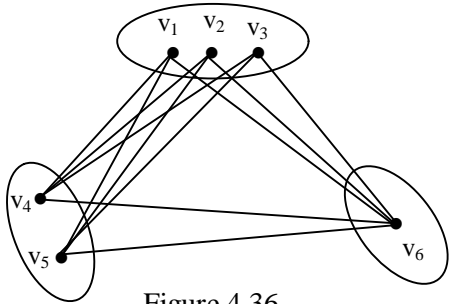

Figure 4.36

The subgraphs of G denoted by S(G) = {$\phi$, G, $\left\{ \begin{smallmatrix} \bullet \\ v_1 \end{smallmatrix} \right\}$, $\left\{ \begin{smallmatrix} \bullet \\ v_2 \end{smallmatrix} \right\}$,

$\left\{ \begin{smallmatrix} \bullet \\ v_3 \end{smallmatrix} \right\}$, $\left\{ \begin{smallmatrix} \bullet \\ v_4 \end{smallmatrix} \right\}$, $\left\{ \begin{smallmatrix} \bullet \\ v_5 \end{smallmatrix} \right\}$, $\left\{ \begin{smallmatrix} \bullet \\ v_6 \end{smallmatrix} \right\}$, … }.

Interested reader can find o(S(G)) and the special topological subgraph of G.

***Example 4.17:*** Let G be a graph given in the following.

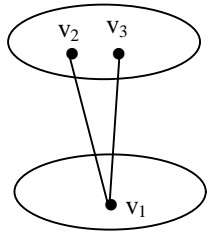

Figure 4.37

We see S(G) = {$\phi$, $\left\{ \begin{smallmatrix} \bullet \\ v_1 \end{smallmatrix} \right\}$, $\left\{ \begin{smallmatrix} \bullet \\ v_2 \end{smallmatrix} \right\}$, $\left\{ \begin{smallmatrix} \bullet \\ v_3 \end{smallmatrix} \right\}$, $\left\{ \begin{smallmatrix} \bullet & \bullet \\ v_1 & v_2 \end{smallmatrix} \right\}$,

$\left\{ \begin{smallmatrix} \bullet & \bullet \\ v_1 & v_3 \end{smallmatrix} \right\}$, $\left\{ \begin{smallmatrix} \bullet & \bullet \\ v_3 & v_2 \end{smallmatrix} \right\}$, $\left\{ \begin{smallmatrix} \bullet & \bullet & \bullet \\ v_1 & v_2 & v_3 \end{smallmatrix} \right\}$, $\left\{ \begin{smallmatrix} \bullet\!\!-\!\!\bullet \\ v_1 \ \ v_2 \end{smallmatrix} \right\}$,

$$\left\{ \underset{v_1 \ \ v_3}{\bullet\!\!-\!\!\bullet} \right\}, \left\{ \underset{v_1 \quad v_2}{\bullet\!\!-\!\!\!-\!\!\bullet} \quad \underset{v_3}{\bullet} \right\}, \left\{ \underset{v_1 \quad v_3}{\bullet\!\!-\!\!\!-\!\!\bullet} \quad \underset{v_2}{\bullet} \right\}, G \right\}$$

$S(G) = 13.$

***Example 4.18:*** Consider the graph

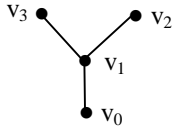

Figure 4.38

To find $S(G)$ and the special topological subgraph associated with G.

$$S(G) = \{\phi, \left\{\underset{v_0}{\bullet}\right\}, \left\{\underset{v_1}{\bullet}\right\}, \left\{\underset{v_2}{\bullet}\right\}, \left\{\underset{v_3}{\bullet}\right\}, \left\{\underset{v_0 \ v_1}{\bullet \ \bullet}\right\}, \left\{\underset{v_0 \ v_2}{\bullet \ \bullet}\right\},$$

$$\left\{\underset{v_0 \ v_3}{\bullet \ \bullet}\right\}, \left\{\underset{v_1 \ v_2}{\bullet \ \bullet}\right\}, \left\{\underset{v_1 \ v_3}{\bullet \ \bullet}\right\}, \left\{\underset{v_2 \ v_3}{\bullet \ \bullet}\right\}, \left\{\underset{v_0 \ v_1}{\bullet\!\!-\!\!\bullet}\right\},$$

$$\left\{\underset{v_1 \ v_2}{\bullet\!\!-\!\!\bullet}\right\}, \left\{\underset{v_1 \ v_3}{\bullet\!\!-\!\!\bullet}\right\}, \left\{\underset{v_0 \ v_1 \ v_2}{\bullet \ \bullet \ \bullet}\right\}, \left\{\underset{v_0 \ v_1 \ v_2}{\bullet\!\!-\!\!\bullet \ \bullet}\right\},$$

$$\left\{\underset{v_0 \ v_1 \ v_2}{\bullet\!\!-\!\!\!-\!\!\bullet}\right\}, \left\{\underset{v_1 \ v_2 \ v_3}{\bullet\!\!-\!\!\bullet \ \bullet}\right\}, \left\{\underset{v_2 \ v_1 \ v_3}{\bullet \ \bullet \ \bullet}\right\}, \left\{\underset{v_1 \ v_2 \ v_3}{\bullet\!\!-\!\!\bullet \ \bullet}\right\},$$

$$\left\{\underset{v_0 \ v_1 \ v_3}{\bullet \ \bullet \ \bullet}\right\}, \left\{\underset{v_0 \ v_1 \ v_3}{\bullet\!\!-\!\!\bullet \ \bullet}\right\}, \left\{\underset{v_0 \ v_1 \ v_3}{\bullet\!\!-\!\!\bullet\!\!-\!\!\bullet}\right\}, \left\{\underset{v_3 \ v_1 \ v_2}{\bullet\!\!-\!\!\bullet\!\!-\!\!\bullet}\right\},$$

$$\left\{\underset{v_0 \ v_2 \ v_3}{\bullet \ \bullet \ \bullet}\right\} G \}.$$

This graph is a tree and S(G) can also be termed as the special topological subgraphs space of a tree.

Now we proceed on to discuss about neutrosophic graphs G and the S(G) associated with them.

It is pertinent to mention here if G is a pure neutrosophic graph then S(G) of it is the same as that of the usual graph G where the neutrosophic edges are replaced by usual edges. The difference occurs only when some of the edges are neutrosophic and other are ordinary or usual edges.

We will proceed onto illustrate them.

Let $v_1 \bullet\!\!\!-\!\!\!-\!\!\!\bullet v_2$ = G be the usual graph and H = $\bullet\text{-}\text{-}\text{-}\bullet$
   Figure 4.39          Figure 4.40

a neutrosophic graph S(G) ≅ S(H) both as lattice graph as well as special topological subgraph.

Take G to be 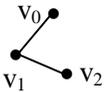 and

Figure 4.41

H to be 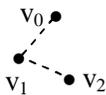

Figure 4.42

to be the usual graph and the neutrosophic graph respectively.

We see S(H) ≅ S(G) as lattice subgraph and special topological subgraph.

Let K = $\overset{v_0}{\underset{v_1 \quad v_2}{\bullet}}$    we see S(K) $\ncong$ S(G) (or S(H)) as lattice

Figure 4.43

graphs.  Also as topologies they are different.

For if we want to speak of isomorphic graphs by no means we agree to map a  •——•  usual edge to a  •- - -•

Figure 4.44

neutrosophic edge.

Hence we see we cannot treat them as identical.

***Example 4.19:***  Let us consider a graph with 3 vertices.

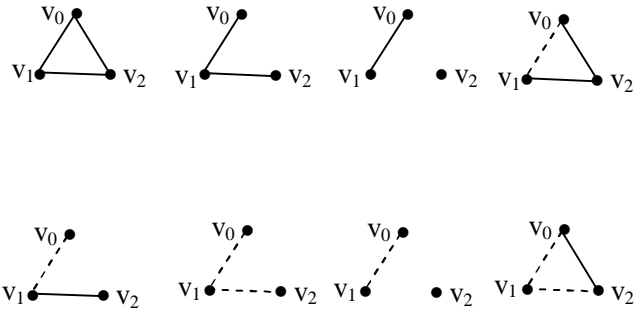

Figure 4.45

and so on.

So we see we have 2 neutrosophic complete graphs with 3 vertices and only one pure neutrosophic graph.

However,

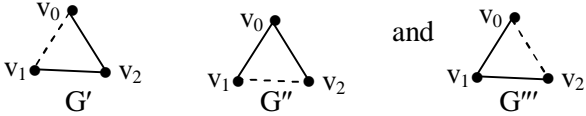

Figure 4.46

are treated as isomorphic. That is $S(G') \cong S(G'') \cong S(G''')$.

But if $K'$ is a neutrosophic graph given by we see $S(G') \not\cong S(K')$ or $S(K'')$ or $S(K'')$.

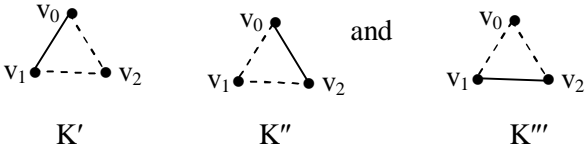

Figure 4.47

However $S(K') \cong S(K''') \cong S(K')$.

We see if T is the pure neutrosophic graph

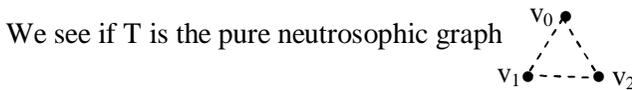

Figure 4.48

we see $S(T) \not\cong S(K')$ or $S(G')$.

Thus we have by defining neutrosophic graphs for the complete graph with three vertices we have three neutrosophic graphs all of them different.

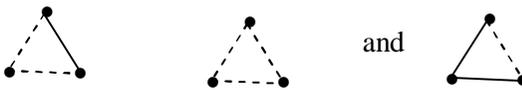

Figure 4.49

Now if we have a complete graph with four vertices planar how many distinct neutrosophic graphs do we get from them. We enlist them in the following.

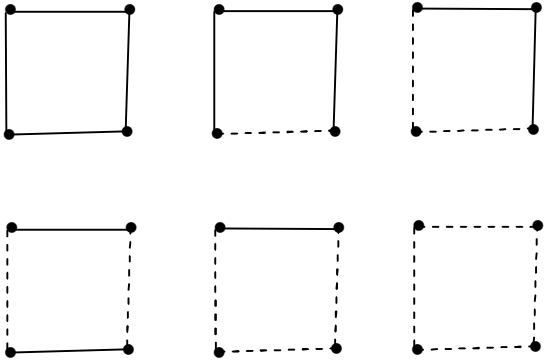

Figure 4.50

Now we can have six topological spaces and they are not isomorphic.

Let G be a non planar complete graph with four vertices.

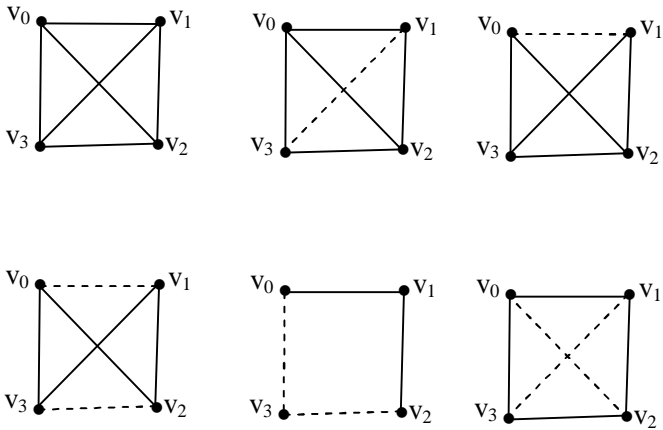

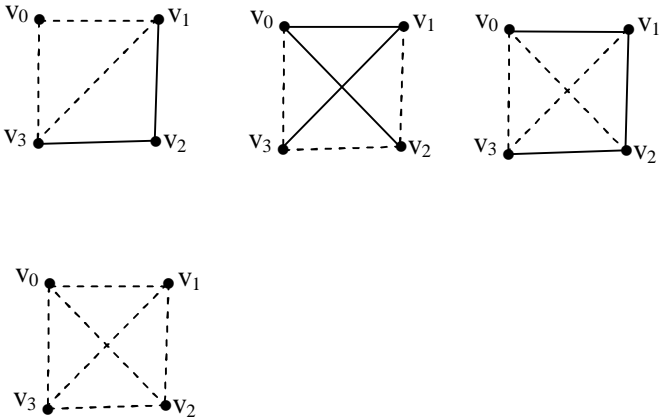

Figure 4.51

We see we have 10 distinct special topological subgraphs spaces and 10 different lattice graphs.

We call these special topological subgraph spaces as special neutrosophic topological subgraph spaces.

Now we see the number  distinct neutrosophic planar connected graphs with 5 vertices.

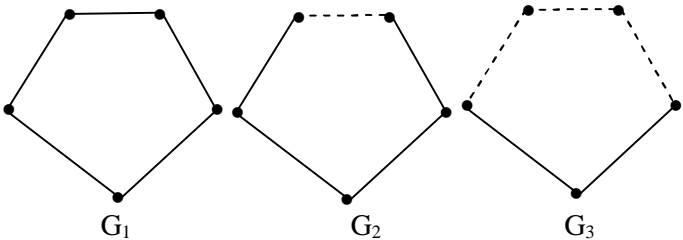

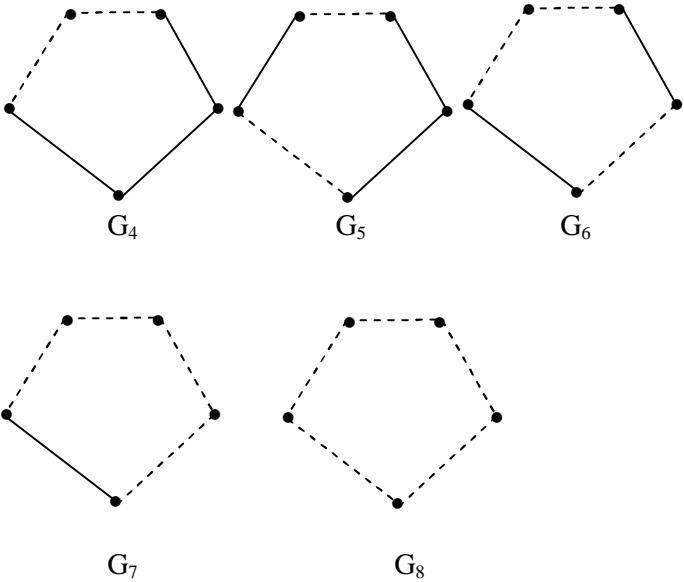

Figure 4.52

We see the neutrosophic complement of $G_1$ is $G_8$.

The neutrosophic complement of $G_2$ is $G_7$. The neutrosophic complement of $G_4$ is $G_3$.

The neutrosophic complement of $G_5$ is $G_6$.

We call $S(G_1)$ the special complement of $S(G_8)$. They $S(G_1)$ and $S(G_8)$ are not complement in usual sense.

We get 8 special topological neutrosophic subgraph spaces.

***Example 4.20:*** Let G be the graph given in the following.

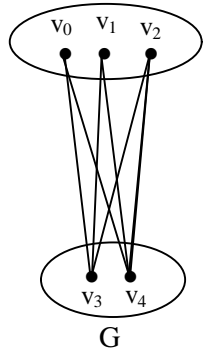

G

Figure 4.53

Then we have the following neutrosophic graphs.

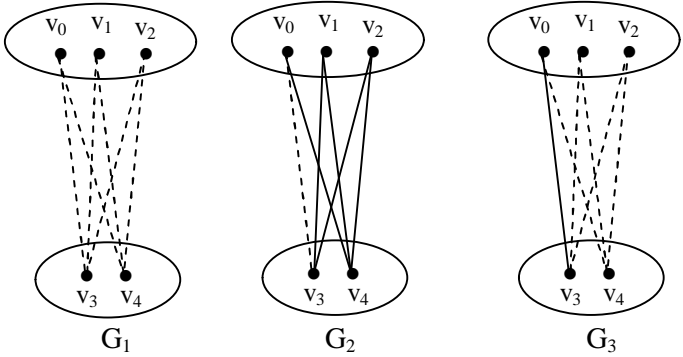

$G_1$       $G_2$       $G_3$

Figure 4.54

$G_1$ is the complement of G, $G_2$ is the complement of $G_3$.

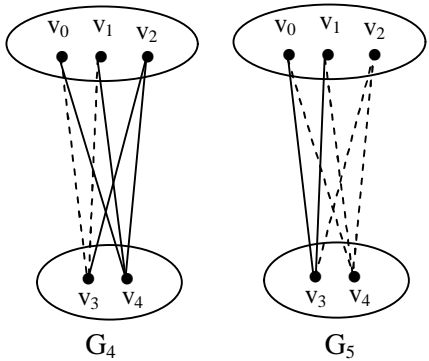

$G_4$       $G_5$

Figure 4.55

We see $G_4$ is the neutrosophic complement of $G_5$.

This is self complemented neutrosophically.

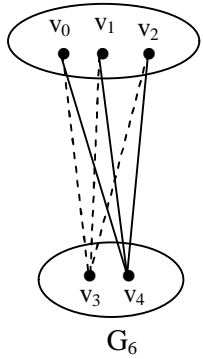

$G_6$

Figure 4.56

$G_7$ is also a neutrosophically self complemented graph.

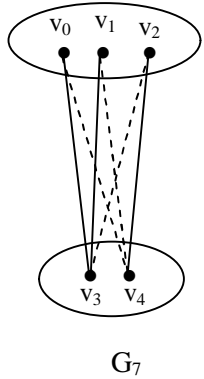

$G_7$

Figure 4.57

Consider

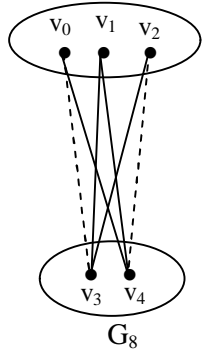

Figure 4.58

then the neutrosophic complement of $G_8$ is $G_9$.

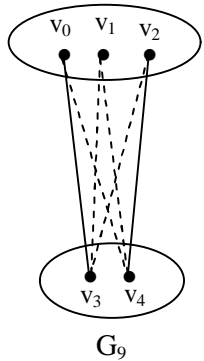

Figure 4.59

We say a neutrosophic graph is self complemented if it is symmetric about the neutrosophic edges and usual edges.

We will give examples of neutrosophic self complemented graphs.

**Example 4.21:** Let G be a neutrosophic graph.

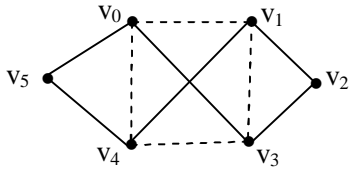

Figure 4.60

Let H be the neutrosophic graph.

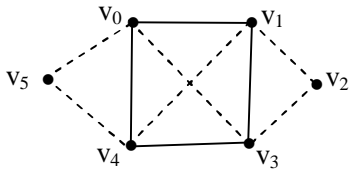

Figure 4.61
G is the neutrosophic complement of H and vice versa.

***Example 4.22:*** Let G be a neutrosophic graph.

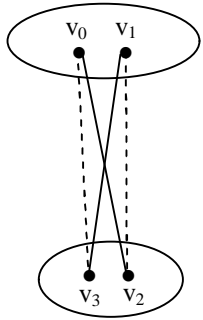

Figure 4.62

G is not a neutrosophic self complemented graph.

The neutrosophic complement graph of G is H which is as follows:

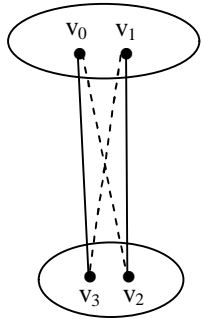

Figure 4.63

**Example 4.23:** Let G be the neutrosophic graph.

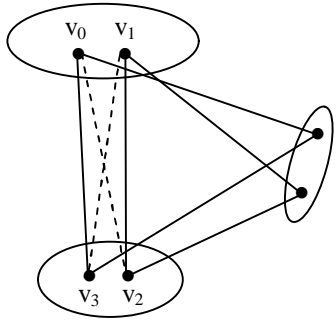

Figure 4.64

The neutrosophic complement of G is H which is as follows:

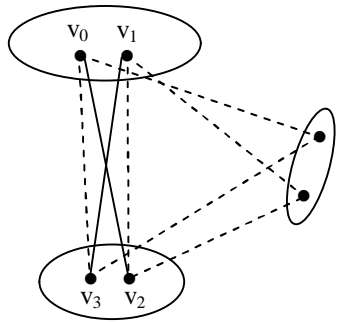

H        Figure 4.65

**Example 4.24:** Let G be the neutrosophic graph.

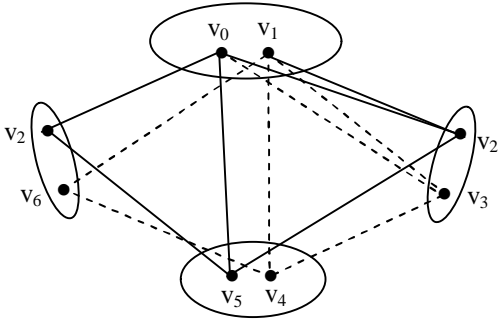

Figure 4.66

Is G a self neutrosophic graph? Justify your answer.

**Example 4.25:** Let G be a neutrosophic graph.

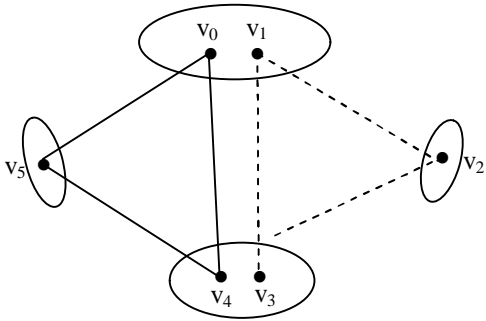

Figure 4.67

G is a self complemented neutrosophic graph.

**Example 4.26:** Let G be a neutrosophic graph.

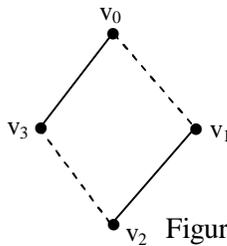

Figure 4.68

G is a self complemented neutrosophic graph.

H = 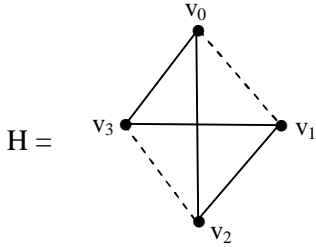

This neutrosophic graph H is not self complemented.

The complement of H is as follows:

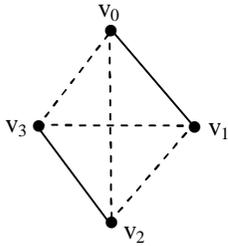

Figure 4.69

Let G be a neutrosophic graph.

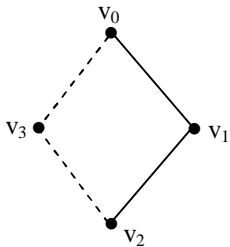

Figure 4.70

G is a self complemented graph.

Let G be the neutrosophic graph

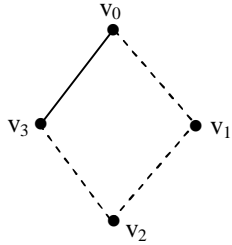

Figure 4.70

This graph is not self complemented.  The neutrosophic

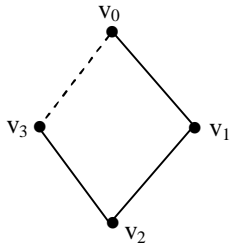

Figure 4.71

graph H is the neutrosophic complement of G.

***Example 4.27:*** Let G be the neutrosophic graph with 7 vertices 3 neutrosophic edges and 10 usual edges.

Will G be a self complemented neutrosophic graph?

We just propose some interesting problems.

**Problem 4.4:** Can a neutrosophic graph G with n neutrosophic edges and p usual edges p ≠ n be a self complemented neutrosophic graph?

**Problem 4.5:** Is it true a graph with n vertices which has equal number of usual edges and neutrosophic edges be a self complemented neutrosophic graph?

*Example 4.28:* Let G be a neutrosophic graph

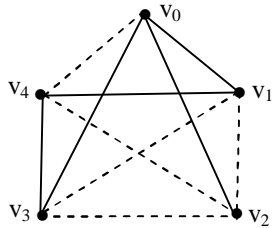

Figure 4.72

Is G a self-complemented neutrosophic graph?

Now we proceed onto study the neutrosophic graph and the adjacency matrix.

Recall from Frank Harary if G is a graph, it can also be considered as a symmetric reflexive relation on a finite set of p points.

This relation is called adjacency and it can be represented by a p × p binary matrix A = [$a_{ij}$]. In A, the i[th] row and column correspond to the i[th] point $v_i$, with $a_{ij} = 1$ if $v_i$ and $v_j$ are adjacent and $a_{ij} = 0$ otherwise.

This is called the adjacency matrix of a graph and there is clearly a one to one correspondence between labeled graphs with p points and p by p symmetric binary matrices with zero diagonal.

Now let G be a neutrosophic graph. Then the adjacency matrix associated with G will be a neutrosophic matrix if $v_i$ and $v_j$ are neutrosphically adjacent then $a_{ij} = I$

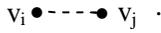

Figure 4.73

We will illustrate this first by some examples.

***Example 4.29:*** Let G be a neutrosophic graph with four vertices.

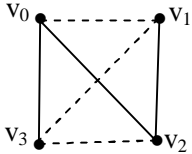

Figure 4.74

The neutrosophic adjacency matrix A associated with G is as follows:

$$A = \begin{bmatrix} 0 & I & 1 & 1 \\ I & 0 & 1 & I \\ 1 & 1 & 0 & I \\ 1 & I & I & 0 \end{bmatrix}.$$

Clearly A is a symmetric neutrosophic matrix.

***Example 4.30:*** Let G be a neutrosophic graph.

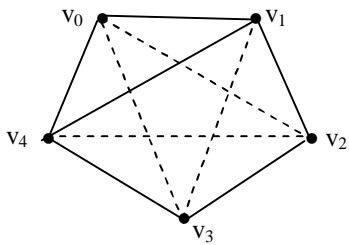

Figure 4.75

The neutrosophic adjacency matrix A of G is as follows:

$$A = \begin{bmatrix} 0 & 1 & I & I & 1 \\ 1 & 0 & 1 & I & 1 \\ I & 1 & 0 & 1 & I \\ I & I & 1 & 0 & 1 \\ 0 & 1 & I & 1 & 0 \end{bmatrix}.$$

Suppose G is the neutrosophic graph.

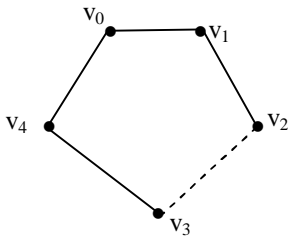

Figure 4.76

The neutrosophic adjacency matrix, A associated with is;

$$A = \begin{bmatrix} 0 & 1 & 0 & 0 & 1 \\ 1 & 0 & 1 & 0 & 0 \\ 0 & 1 & 0 & I & 0 \\ 0 & 0 & I & 0 & 1 \\ 1 & 0 & 0 & 1 & 0 \end{bmatrix}.$$

**Example 4.31:** Let G be a neutrosophic matrix which is as follows:

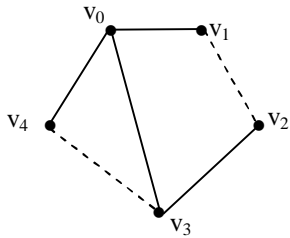

Figure 4.77

The neutrosophic adjacency matrix A associated with G is,

$$A = \begin{bmatrix} 0 & 1 & 0 & 1 & 1 \\ 1 & 0 & I & 0 & 0 \\ 0 & I & 0 & 1 & 0 \\ 1 & 0 & 1 & 0 & I \\ 1 & 0 & 0 & I & 0 \end{bmatrix}.$$

It is interesting to note if the neutrosophic matrix associated with a neutrosophic graph which has only one edge to be neutrosophic and the rest are all usual then the neutrosophic adjacency matrix A will have only two entries as I and all other entries will be 0 and 1.

Another observation is if G is a neutrosophic graph with only two neutrosophic edges then the associated neutrosophic adjacency matrix with have only 4 entries to be I's and the rest 0's or 1's.

In view of this we have the following theorem the proof of which is direct and hence left as an exercise to the reader.

**THEOREM 4.6:** *Let G be a neutrosophic graph. G has only one neutrosophic edge if and only if the neutrosophic adjacency matrix A of G has two neutrosophic entries and the rest are 0's and 1's.*

We give the generalized result in the following theorem.

**THEOREM 4.7:** *Let G be neutrosophic graph. G has s edges to be neutrosophic if and only if the adjacency neutrosophic matrix A associated with G has 2s neutrosophic entries (ie. A has totally 2s entries which are I) and the rest of the entries of A are either 0 or 1.*

This proof is also direct and hence left as an exercise to the reader.

We have the result of usual graph to be true in case of neutrosophic graphs also.

**THEOREM 4.8:** *A neutrosophic graph G is disconnected if for some labeling the neutrosophic adjacency matrix A can be partitioned into neutrosophic submatrices $A_{11}$, $A_{12}$, $A_{21}$ and $A_{22}$ where $A_{11}$ and $A_{22}$ are square neutrosophic matrices.*

$$\begin{bmatrix} A_{11} & 0 \\ 0 & A_{22} \end{bmatrix}$$

We will proceed onto give some more examples before we proceed to give analogous theorems.

***Example 4.32:*** Let G be a neutrosophic graph given in the following.

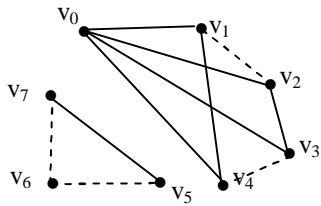

Figure 4.78

The adjacency neutrosophic matrix A associated with G is as follows:

$$A = \begin{bmatrix} 0 & 1 & 1 & 1 & 1 & 0 & 0 & 0 \\ 1 & 0 & I & 0 & 1 & 0 & 0 & 0 \\ 1 & I & 0 & 1 & 0 & 0 & 0 & 0 \\ 1 & 0 & 1 & 0 & I & 0 & 0 & 0 \\ 1 & 1 & 0 & I & 0 & 0 & 0 & 0 \\ 0 & 0 & 0 & 0 & 0 & 0 & I & 1 \\ 0 & 0 & 0 & 0 & 0 & I & 0 & I \\ 0 & 0 & 0 & 0 & 0 & 1 & I & 0 \end{bmatrix}$$

We see the neutrosophic graph is disconnected and the neutrosophic matrix takes the form.

$$\text{That is } A = \begin{bmatrix} A_{11} & 0 \\ \hline 0 & A_{22} \end{bmatrix}.$$

We call A in this form as a super symmetric square diagonal matrix.

In view of this we give some more examples before we prove a few results.

***Example 4.33:*** Let G be a neutrosophic graph which is as follows:

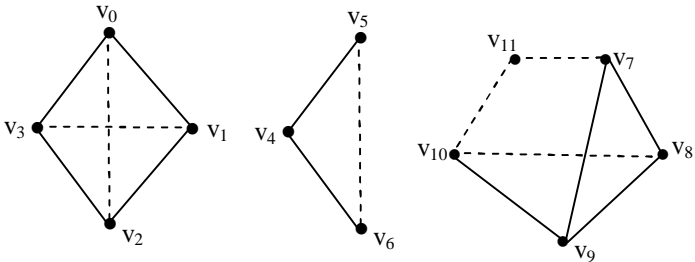

Figure 4.79

The neutrosophic adjacency matrix associated with G is as follows:

$$A = \begin{bmatrix}
0 & 1 & I & 1 & 0 & 0 & 0 & 0 & 0 & 0 & 0 & 0 \\
1 & 0 & 1 & I & 0 & 0 & 0 & 0 & 0 & 0 & 0 & 0 \\
I & 1 & 0 & 1 & 0 & 0 & 0 & 0 & 0 & 0 & 0 & 0 \\
1 & I & 1 & 0 & 0 & 0 & 0 & 0 & 0 & 0 & 0 & 0 \\
0 & 0 & 0 & 0 & 0 & 1 & 1 & 0 & 0 & 0 & 0 & 0 \\
0 & 0 & 0 & 0 & 1 & 0 & I & 0 & 0 & 0 & 0 & 0 \\
0 & 0 & 0 & 0 & 1 & I & 0 & 0 & 0 & 0 & 0 & 0 \\
0 & 0 & 0 & 0 & 0 & 0 & 0 & 0 & 1 & 1 & 0 & I \\
0 & 0 & 0 & 0 & 0 & 0 & 0 & 1 & 0 & 1 & 1 & 0 \\
0 & 0 & 0 & 0 & 0 & 0 & 0 & 1 & 1 & 0 & 1 & 0 \\
0 & 0 & 0 & 0 & 0 & 0 & 0 & 0 & 1 & 1 & 0 & I \\
0 & 0 & 0 & 0 & 0 & 0 & 0 & I & 0 & 0 & I & 0
\end{bmatrix}$$

$$= \begin{bmatrix}
A_1 & (0) & (0) \\
(0) & A_2 & (0) \\
(0) & (0) & A_3
\end{bmatrix}.$$

We see A is a super symmetric square diagonal neutrosophic matrix. The main diagonal element has three matrices.

We make the following observation the adjacency neutrosophic super matrix A of the graph G has three main diagonal elements and the graph G has 3 disjoint subgraphs.

Further this super matrix is symmetric.

So in view of this we give the following theorem.

**THEOREM 4.9:** *Let G be a neutrosophic graph. If the adjacency neutrosophic matrix is a symmetric super neutrosophic square diagonal matrix of the form*

$$A = \begin{bmatrix} A_1 & (0) & ... & ... & (0) \\ (0) & A_2 & ... & ... & (0) \\ (0) & (0) & A_3 & ... & ... \\ (0) & (0) & (0) & ... & ... \\ (0) & ... & (0) & ... & A_n \end{bmatrix}$$

*where each $A_n$ is a symmetric matrix. Then G is a neutrosophic graph with n disjoint components.*

We will illustrate this situation by an example or two.

Proof is direct and hence left as an exercise to the reader.

***Example 4.34:*** Let G be a neutrosophic graph which is as follows:

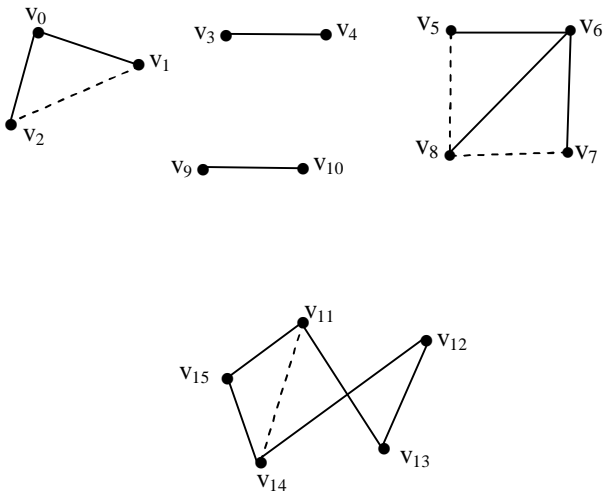

Figure 4.80

Let A be the neutrosophic adjacency matrix associated with G.

$$
A = \left[\begin{array}{ccc|cc|cccc|cc|ccccc}
0 & 1 & I & 0 & 0 & 0 & 0 & 0 & 0 & 0 & 0 & 0 & 0 & 0 & 0 & 0 \\
1 & 0 & I & 0 & 0 & 0 & 0 & 0 & 0 & 0 & 0 & 0 & 0 & 0 & 0 & 0 \\
I & I & 0 & 0 & 0 & 0 & 0 & 0 & 0 & 0 & 0 & 0 & 0 & 0 & 0 & 0 \\
\hline
0 & 0 & 0 & 0 & 1 & 0 & 0 & 0 & 0 & 0 & 0 & 0 & 0 & 0 & 0 & 0 \\
0 & 0 & 0 & 1 & 0 & 0 & 0 & 0 & 0 & 0 & 0 & 0 & 0 & 0 & 0 & 0 \\
\hline
0 & 0 & 0 & 0 & 0 & 0 & 1 & 0 & I & 0 & 0 & 0 & 0 & 0 & 0 & 0 \\
0 & 0 & 0 & 0 & 0 & 1 & 0 & 1 & 1 & 0 & 0 & 0 & 0 & 0 & 0 & 0 \\
0 & 0 & 0 & 0 & 0 & 0 & 1 & 0 & I & 0 & 0 & 0 & 0 & 0 & 0 & 0 \\
0 & 0 & 0 & 0 & 0 & I & 1 & I & 0 & 0 & 0 & 0 & 0 & 0 & 0 & 0 \\
\hline
0 & 0 & 0 & 0 & 0 & 0 & 0 & 0 & 0 & 0 & 1 & 0 & 0 & 0 & 0 & 0 \\
0 & 0 & 0 & 0 & 0 & 0 & 0 & 0 & 0 & 1 & 0 & 0 & 0 & 0 & 0 & 0 \\
\hline
0 & 0 & 0 & 0 & 0 & 0 & 0 & 0 & 0 & 0 & 0 & 0 & 0 & 1 & I & 1 \\
0 & 0 & 0 & 0 & 0 & 0 & 0 & 0 & 0 & 0 & 0 & 0 & 0 & 1 & 1 & 0 \\
0 & 0 & 0 & 0 & 0 & 0 & 0 & 0 & 0 & 0 & 0 & 1 & 1 & 0 & 0 & 0 \\
0 & 0 & 0 & 0 & 0 & 0 & 0 & 0 & 0 & 0 & 0 & I & 1 & 0 & 0 & 1 \\
0 & 0 & 0 & 0 & 0 & 0 & 0 & 0 & 0 & 0 & 0 & 1 & 0 & 0 & 1 & 0
\end{array}\right]
$$

$$
= \left[\begin{array}{c|c|c|c|c}
A_1 & (0) & (0) & (0) & (0) \\
\hline
(0) & A_2 & (0) & (0) & (0) \\
\hline
(0) & (0) & A_3 & (0) & (0) \\
\hline
(0) & (0) & (0) & A_4 & (0) \\
\hline
(0) & (0) & (0) & (0) & A_5
\end{array}\right].
$$

Clearly A is a neutrosophic symmetric super diagonal square matrix with 5 main diagonal elements.

Further the neutrosophic graph G has 5 disjoint components associated with it.

***Example 4.35:*** Let G be a neutrosophic graph.

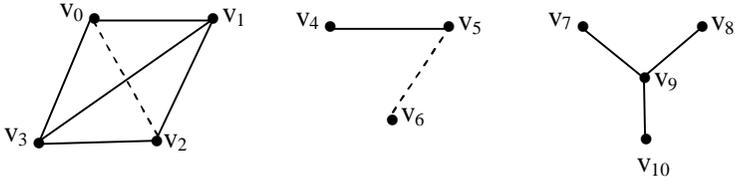

Figure 4.81

The neutrosophic adjacency matrix A associated with G is as follows:

$$A = \begin{bmatrix} 0 & 1 & I & 1 & 0 & 0 & 0 & 0 & 0 & 0 & 0 \\ 1 & 0 & 1 & 1 & 0 & 0 & 0 & 0 & 0 & 0 & 0 \\ I & 1 & 0 & 1 & 0 & 0 & 0 & 0 & 0 & 0 & 0 \\ 1 & 1 & 1 & 0 & 0 & 0 & 0 & 0 & 0 & 0 & 0 \\ \hline 0 & 0 & 0 & 0 & 0 & 1 & 0 & 0 & 0 & 0 & 0 \\ 0 & 0 & 0 & 0 & 1 & 0 & I & 0 & 0 & 0 & 0 \\ 0 & 0 & 0 & 0 & 0 & I & 0 & 0 & 0 & 0 & 0 \\ \hline 0 & 0 & 0 & 0 & 0 & 0 & 0 & 0 & 0 & 1 & 0 \\ 0 & 0 & 0 & 0 & 0 & 0 & 0 & 0 & 0 & 1 & 0 \\ 0 & 0 & 0 & 0 & 0 & 0 & 0 & 1 & 1 & 0 & 1 \\ 0 & 0 & 0 & 0 & 0 & 0 & 0 & 0 & 0 & 1 & 0 \end{bmatrix}$$

$$= \begin{bmatrix} A_1 & (0) & (0) \\ \hline (0) & A_2 & (0) \\ \hline (0) & (0) & A_3 \end{bmatrix}.$$

We see again A is a super neutrosophic symmetric square matrix.

Almost all properties associated with usual graphs are true in case of these neutrosophic graphs also.

We will just state the results the proofs can be supplied by the interested reader.

**THEOREM 4.10:** *Let G be a neutrosophic graph with neutrosophic adjacency matrix A. Then (i, j) entry of $A^n$ is the number of different walk of length n from $v_i$ to $v_j$.*

Now we illustrate these situations by some simple examples.

***Example 4.36:*** Let G be a neutrosophic graph with adjacency neutrosophic matrix A.

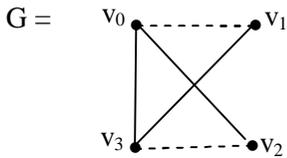

$$G =$$

Figure 4.82

$$A = . \begin{bmatrix} 0 & I & 1 & 1 \\ I & 0 & 0 & 1 \\ 1 & 0 & 0 & I \\ 1 & 1 & I & 0 \end{bmatrix}$$

Now consider the neutrosophic subgraph H

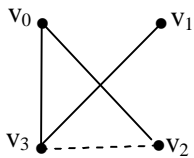

Figure 4.83

The neutrosophic adjacency matrix A′ of the subgraph H is as follows:

$$A' = \begin{bmatrix} 0 & 0 & 1 & 1 \\ 0 & 0 & 0 & 1 \\ 1 & 0 & 0 & I \\ 1 & 1 & I & 0 \end{bmatrix}.$$

Let P be the neutrosophic subgraph of G.

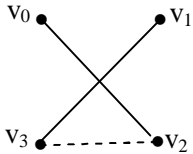

Figure 4.84

The adjacency neutrosophic matrix A″ of P is

$$A'' = \begin{bmatrix} 0 & 0 & 1 & 0 \\ 0 & 0 & 0 & 1 \\ 1 & 0 & 0 & I \\ 0 & 1 & I & 0 \end{bmatrix} \text{ and so on.}$$

Next we give the definition of a incidence matrix of a neutrosophic graph G.

Let G be a neutrosophic graph with n vertices and m edges (m edges is both neutrosophic edges as well as usual edges) and without self loop.

The incidence neutrosophic matrix A of G is a m × n matrix A = [a_{ij}] whose n rows correspond to the n vertices and the m columns correspond to m edges as

$$a_{ij} = \begin{cases} 1 \text{ if jth edge is incidence on the ith vertex} \\ I \text{ if jth edge is neutrosophic incidence on the ith vertex} \\ 0 \text{ otherwise.} \end{cases}$$

It is also called vertex edge incidence neutrosophic matrix and is denoted by A(G).

We will proceed onto give a few examples on them.

***Example 4.37:*** Let G be a neutrosophic graph.

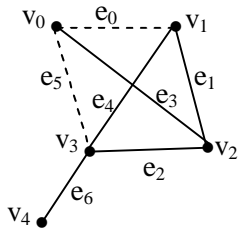

Figure 4.85

The neutrosophic incidence matrix of G is as follows:

$$A(G) = \begin{array}{c} \\ v_0 \\ v_1 \\ v_2 \\ v_3 \\ v_4 \end{array} \begin{array}{c} \begin{array}{ccccccc} e_0 & e_1 & e_2 & e_3 & e_4 & e_5 & e_6 \end{array} \\ \begin{bmatrix} I & 0 & 0 & 1 & 0 & I & 0 \\ I & 1 & 0 & 0 & 1 & 0 & 0 \\ 0 & 1 & 1 & 1 & 0 & 0 & 0 \\ 0 & 0 & 1 & 0 & 1 & I & 0 \\ 0 & 0 & 0 & 0 & 0 & 0 & 1 \end{bmatrix} \end{array}.$$

A(G) is clearly a neutrosophic matrix.

***Example 4.38:*** Let G be a neutrosophic graph.

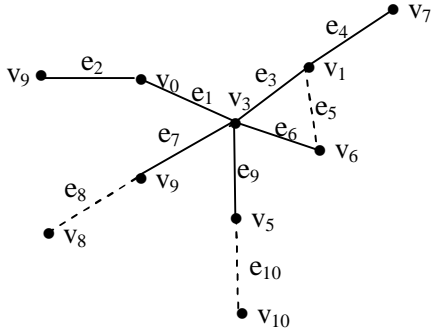

Figure 4.86

The incidence neutrosophic matrix A(G) as follows:

$$
= \begin{array}{c}
\\
v_1 \\ v_2 \\ v_3 \\ v_4 \\ v_5 \\ v_6 \\ v_7 \\ v_8 \\ v_9
\end{array}
\begin{array}{cccccccccc}
e_1 & e_2 & e_3 & e_4 & e_5 & e_6 & e_7 & e_8 & e_9 & e_{10} \\
\left[\begin{array}{cccccccccc}
1 & 1 & 0 & 0 & 0 & 0 & 0 & 0 & 0 & 0 \\
0 & 0 & 0 & 1 & I & 1 & 0 & 0 & 0 & 0 \\
1 & 0 & 1 & 0 & 0 & 1 & 1 & 0 & 1 & 0 \\
0 & 0 & 0 & 0 & 0 & 0 & 1 & I & 0 & 0 \\
0 & 0 & 0 & 0 & 0 & 0 & 0 & 0 & 1 & I \\
0 & 0 & 0 & 0 & I & 1 & 0 & 0 & 0 & 0 \\
0 & 0 & 0 & 1 & 0 & 0 & 0 & 0 & 0 & 0 \\
0 & 0 & 0 & 0 & 0 & 0 & 0 & I & 0 & 0 \\
0 & 1 & 0 & 0 & 0 & 0 & 0 & 0 & 0 & 0
\end{array}\right]
\end{array}.
$$

**Example 4.39:** Let G be a neutrosophic graph. Let A(G) be the incidence neutrosophic matrix.

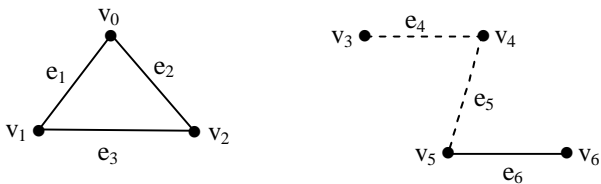

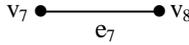

Figure 4.87

$$A(G) = \begin{array}{c} \\ v_0 \\ v_1 \\ v_2 \\ v_3 \\ v_4 \\ v_5 \\ v_6 \\ v_7 \\ v_8 \end{array} \begin{array}{c} e_1 \ e_2 \ e_3 \ e_4 \ e_5 \ e_6 \ e_7 \\ \left[ \begin{array}{ccc|ccc|c} 1 & 1 & 0 & 0 & 0 & 0 & 0 \\ 1 & 0 & 1 & 0 & 0 & 0 & 0 \\ 0 & 1 & 1 & 0 & 0 & 0 & 0 \\ \hline 0 & 0 & 0 & I & 0 & 0 & 0 \\ 0 & 0 & 0 & I & I & 0 & 0 \\ 0 & 0 & 0 & I & 0 & 1 & 0 \\ 0 & 0 & 0 & 0 & 0 & 1 & 0 \\ \hline 0 & 0 & 0 & 0 & 0 & 0 & 1 \\ 0 & 0 & 0 & 0 & 0 & 0 & 1 \end{array} \right] \end{array}.$$

We see if the neutrosophic graph is disjoint the incidence matrix is a neutrosophic diagonal matrix.

$$A(G) = \begin{bmatrix} A_1 & (0) & (0) \\ \hline (0) & A_2 & (0) \\ \hline (0) & (0) & A_3 \end{bmatrix}.$$

Clearly this super neutrosophic incidence matrix is not symmetric or square only diagonal in general.

Even if we have a disjoint graph G usual not necessarily be neutrosophic then also the adjacency matrix is a super diagonal symmetric square matrix and the incidence matrix is a super diagonal matrix non necessarily square or symmetric.

We will illustrate this situation by some examples.

***Example 4.40:*** Let G be a graph which is as follows:

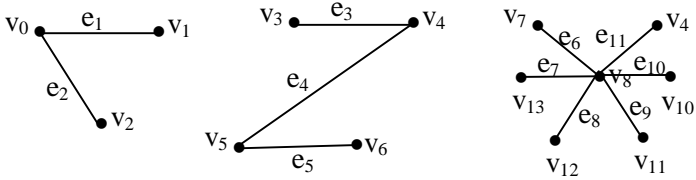

Figure 4.88

The adjacency matrix A of G is as follows:

$$
A = \begin{array}{c} \\ v_0 \\ v_1 \\ v_2 \\ v_3 \\ v_4 \\ v_5 \\ v_6 \\ v_7 \\ v_8 \\ v_9 \\ v_{10} \\ v_{11} \\ v_{12} \\ v_{13} \end{array}
\begin{array}{c}
\begin{array}{ccccccccccccccc} v_0 & v_1 & v_2 & v_3 & v_4 & v_5 & v_6 & v_7 & v_8 & v_9 & v_{10} & v_{11} & v_{12} & v_{13} \end{array} \\
\left[ \begin{array}{ccc|cccc|ccccccc}
0 & 1 & 1 & 0 & 0 & 0 & 0 & 0 & 0 & 0 & 0 & 0 & 0 & 0 \\
1 & 0 & 0 & 0 & 0 & 0 & 0 & 0 & 0 & 0 & 0 & 0 & 0 & 0 \\
1 & 0 & 0 & 0 & 0 & 0 & 0 & 0 & 0 & 0 & 0 & 0 & 0 & 0 \\
0 & 0 & 0 & 0 & 1 & 0 & 0 & 0 & 0 & 0 & 0 & 0 & 0 & 0 \\
0 & 0 & 0 & 1 & 0 & 1 & 0 & 0 & 0 & 0 & 0 & 0 & 0 & 0 \\
0 & 0 & 0 & 0 & 1 & 0 & 1 & 0 & 0 & 0 & 0 & 0 & 0 & 0 \\
0 & 0 & 0 & 0 & 0 & 1 & 0 & 0 & 0 & 0 & 0 & 0 & 0 & 0 \\
0 & 0 & 0 & 0 & 0 & 0 & 0 & 0 & 1 & 0 & 0 & 0 & 0 & 0 \\
0 & 0 & 0 & 0 & 0 & 0 & 0 & 1 & 0 & 1 & 1 & 1 & 1 & 1 \\
0 & 0 & 0 & 0 & 0 & 0 & 0 & 0 & 1 & 0 & 0 & 0 & 0 & 0 \\
0 & 0 & 0 & 0 & 0 & 0 & 0 & 0 & 1 & 0 & 0 & 0 & 0 & 0 \\
0 & 0 & 0 & 0 & 0 & 0 & 0 & 0 & 1 & 0 & 0 & 0 & 0 & 0 \\
0 & 0 & 0 & 0 & 0 & 0 & 0 & 0 & 1 & 0 & 0 & 0 & 0 & 0 \\
0 & 0 & 0 & 0 & 0 & 0 & 0 & 0 & 1 & 0 & 0 & 0 & 0 & 0 \\
\end{array} \right]
\end{array}
$$

$$= \begin{bmatrix} A_1 & (0) & (0) \\ \hline (0) & A_2 & (0) \\ \hline (0) & (0) & A_3 \end{bmatrix}.$$

It is clear A is a symmetric super diagonal square matrix.

Now we find the incidence matrix A(G) of G.

$$A(G) = \begin{array}{c} \\ v_0 \\ v_1 \\ v_2 \\ v_3 \\ v_4 \\ v_5 \\ v_6 \\ v_7 \\ v_8 \\ v_9 \\ v_{10} \\ v_{11} \\ v_{12} \\ v_{13} \end{array} \begin{array}{c} e_1 \ e_2 \ \ e_3 \ e_4 \ e_5 \ \ e_6 \ e_7 \ e_8 \ \ e_9 \ e_{10} \ e_{11} \\ \begin{bmatrix} 1 & 1 & 0 & 0 & 0 & 0 & 0 & 0 & 0 & 0 & 0 \\ 1 & 0 & 0 & 0 & 0 & 0 & 0 & 0 & 0 & 0 & 0 \\ 0 & 1 & 0 & 0 & 0 & 0 & 0 & 0 & 0 & 0 & 0 \\ 0 & 0 & 1 & 0 & 0 & 0 & 0 & 0 & 0 & 0 & 0 \\ 0 & 0 & 1 & 1 & 0 & 0 & 0 & 0 & 0 & 0 & 0 \\ 0 & 0 & 0 & 1 & 1 & 0 & 0 & 0 & 0 & 0 & 0 \\ 0 & 0 & 0 & 0 & 1 & 0 & 0 & 0 & 0 & 0 & 0 \\ 0 & 0 & 0 & 0 & 0 & 1 & 0 & 0 & 0 & 0 & 0 \\ 0 & 0 & 0 & 0 & 0 & 1 & 1 & 1 & 1 & 1 & 1 \\ 0 & 0 & 0 & 0 & 0 & 0 & 0 & 0 & 0 & 0 & 1 \\ 0 & 0 & 0 & 0 & 0 & 0 & 0 & 0 & 0 & 1 & 0 \\ 0 & 0 & 0 & 0 & 0 & 0 & 0 & 0 & 1 & 0 & 0 \\ 0 & 0 & 0 & 0 & 0 & 0 & 0 & 1 & 0 & 0 & 0 \\ 0 & 0 & 0 & 0 & 0 & 0 & 1 & 0 & 0 & 0 & 0 \end{bmatrix} \end{array}$$

A(G) is a diagonal super matrix which is not square.

$$\text{Thus } A(G) = \begin{bmatrix} A_1 & (0) & (0) \\ \hline (0) & A_2 & (0) \\ \hline (0) & (0) & A_3 \end{bmatrix}$$

Now we give yet another examples.

***Example 4.41:*** Let G be a graph which is as follows:

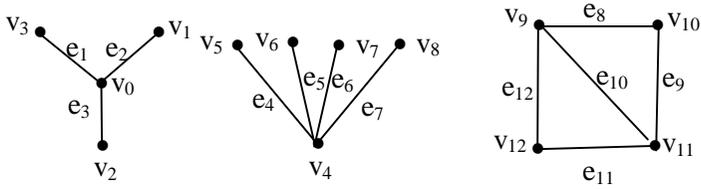

Figure 4.89

The incidence matrix A(G) associated with G is as follows:

$$A(G) = \begin{array}{c} \\ v_0 \\ v_1 \\ v_2 \\ v_3 \\ v_4 \\ v_5 \\ v_6 \\ v_7 \\ v_8 \\ v_9 \\ v_{10} \\ v_{11} \\ v_{12} \end{array} \begin{array}{cccccccccccc} e_1 & e_2 & e_3 & e_4 & e_5 & e_6 & e_7 & e_8 & e_9 & e_{10} & e_{11} & e_{12} \\ \left[\begin{array}{ccc|cccc|ccccc} 1 & 1 & 1 & 0 & 0 & 0 & 0 & 0 & 0 & 0 & 0 & 0 \\ 0 & 1 & 0 & 0 & 0 & 0 & 0 & 0 & 0 & 0 & 0 & 0 \\ 0 & 0 & 1 & 0 & 0 & 0 & 0 & 0 & 0 & 0 & 0 & 0 \\ 1 & 0 & 0 & 0 & 0 & 0 & 0 & 0 & 0 & 0 & 0 & 0 \\ \hline 0 & 0 & 0 & 1 & 1 & 1 & 1 & 0 & 0 & 0 & 0 & 0 \\ 0 & 0 & 0 & 1 & 0 & 0 & 0 & 0 & 0 & 0 & 0 & 0 \\ 0 & 0 & 0 & 0 & 1 & 0 & 0 & 0 & 0 & 0 & 0 & 0 \\ 0 & 0 & 0 & 0 & 0 & 1 & 0 & 0 & 0 & 0 & 0 & 0 \\ 0 & 0 & 0 & 0 & 0 & 0 & 1 & 0 & 0 & 0 & 0 & 0 \\ \hline 0 & 0 & 0 & 0 & 0 & 0 & 0 & 1 & 0 & 1 & 0 & 1 \\ 0 & 0 & 0 & 0 & 0 & 0 & 0 & 1 & 1 & 0 & 0 & 0 \\ 0 & 0 & 0 & 0 & 0 & 0 & 0 & 0 & 1 & 1 & 1 & 0 \\ 0 & 0 & 0 & 0 & 0 & 0 & 0 & 0 & 0 & 0 & 1 & 1 \end{array}\right] \end{array}$$

$$= \begin{bmatrix} A_1 & (0) & (0) \\ \hline (0) & A_2 & (0) \\ \hline (0) & (0) & A_3 \end{bmatrix}.$$

Clearly A(G) is a super diagonal matrix which is not square or symmetric.

Now we give the adjacency matrix A of G.

$$A = \begin{array}{c} \\ v_0 \\ v_1 \\ v_2 \\ v_3 \\ v_4 \\ v_5 \\ v_6 \\ v_7 \\ v_8 \\ v_9 \\ v_{10} \\ v_{11} \\ v_{12} \end{array} \begin{array}{cccc|ccccc|cccc} v_0 & v_1 & v_2 & v_3 & v_4 & v_5 & v_6 & v_7 & v_8 & v_9 & v_{10} & v_{11} & v_{12} \\ \hline 0 & 1 & 1 & 1 & 0 & 0 & 0 & 0 & 0 & 0 & 0 & 0 & 0 \\ 1 & 0 & 0 & 0 & 0 & 0 & 0 & 0 & 0 & 0 & 0 & 0 & 0 \\ 1 & 0 & 0 & 0 & 0 & 0 & 0 & 0 & 0 & 0 & 0 & 0 & 0 \\ 1 & 0 & 0 & 0 & 0 & 0 & 0 & 0 & 0 & 0 & 0 & 0 & 0 \\ 0 & 0 & 0 & 0 & 0 & 1 & 1 & 1 & 1 & 0 & 0 & 0 & 0 \\ 0 & 0 & 0 & 0 & 1 & 0 & 0 & 0 & 0 & 0 & 0 & 0 & 0 \\ 0 & 0 & 0 & 0 & 1 & 0 & 0 & 0 & 0 & 0 & 0 & 0 & 0 \\ 0 & 0 & 0 & 0 & 1 & 0 & 0 & 0 & 0 & 0 & 0 & 0 & 0 \\ 0 & 0 & 0 & 0 & 1 & 0 & 0 & 0 & 0 & 0 & 0 & 0 & 0 \\ 0 & 0 & 0 & 0 & 0 & 0 & 0 & 0 & 0 & 0 & 1 & 1 & 1 \\ 0 & 0 & 0 & 0 & 0 & 0 & 0 & 0 & 0 & 1 & 0 & 1 & 0 \\ 0 & 0 & 0 & 0 & 0 & 0 & 0 & 0 & 0 & 1 & 1 & 0 & 1 \\ 0 & 0 & 0 & 0 & 0 & 0 & 0 & 0 & 0 & 1 & 0 & 1 & 0 \end{array}$$

$$= \begin{bmatrix} A_1 & (0) & (0) \\ \hline (0) & A_2 & (0) \\ \hline (0) & (0) & A_3 \end{bmatrix}$$

is a super symmetric square diagonal matrix.

We have the following results the proof of which is left as an exercise to the reader.

**THEOREM 4.11:** *Let G be a graph which has m disjoint graphs.*
    *(i) Then the adjacency matrix A of G is a super symmetric diagonal square matrix.*

*(ii) The incidence matrix A(G) of G is a super diagonal matrix.*

*We see the number of diagonal matrices (elements) A(G) and A are m in number that is*

$$A(G) = \begin{bmatrix} A_1 & (0) & \dots & (0) \\ (0) & A_2 & (0) & \\ \vdots & \vdots & \ddots & \vdots \\ (0) & (0) & (0) & A_m \end{bmatrix}$$

*and*

$$A = \begin{bmatrix} A'_1 & (0) & (0) & \dots & (0) \\ (0) & A'_2 & (0) & & (0) \\ \vdots & \vdots & \vdots & \ddots & \vdots \\ (0) & (0) & (0) & \dots & A'_m \end{bmatrix}$$

*where $A_j$'s are rectangular matrices where as $A'_j$ are square symmetric diagonal matrices $1 \le j \le m$.*

Proof is direct hence left as an exercise to the reader.

We has been examples we see the same is true or neutrosophic graphs with m disjoint neutrosophic graphs.

We will now show by examples the power of an adjacency neutrosophic graph G.

***Example 4.42:*** Let G be the neutrosophic graph

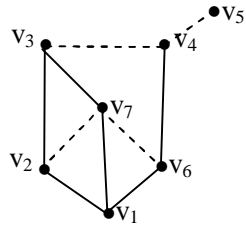

Figure 4.90

The adjacency matrix A associated with G is as follows:

$$
A = \begin{array}{c c}
& \begin{array}{c c c c c c c}
v_1 & v_2 & v_3 & v_4 & v_5 & v_6 & v_7
\end{array} \\
\begin{array}{c}
v_1 \\ v_2 \\ v_3 \\ v_4 \\ v_5 \\ v_6 \\ v_7
\end{array} &
\begin{bmatrix}
0 & 1 & 0 & 0 & 0 & 1 & 1 \\
1 & 0 & 1 & 0 & 0 & 0 & I \\
0 & 1 & 0 & I & 0 & 0 & 1 \\
0 & 0 & I & 0 & I & 1 & 0 \\
0 & 0 & 0 & I & 0 & 0 & 0 \\
1 & 0 & 0 & 1 & 0 & 0 & I \\
1 & I & 1 & 0 & 0 & I & 0
\end{bmatrix}
\end{array} ;
$$

A is a $7 \times 7$ neutrosophic symmetric matrix.

$$
A^2 = \begin{bmatrix}
3 & I & 1+I & 1 & 0 & I & 2I \\
I & 2+I & I & I & 0 & 1+I & 1+I \\
1 & I & 2+I & 0 & I & 2I & 0 \\
1 & 0 & 0 & 1+I & 0 & 0 & 2I \\
0 & 0 & I & 0 & I & I & 0 \\
I & 1+I & I & 0 & I & 1+2I & 1 \\
2I & 1 & I & 2I & 0 & 1 & 1+2I
\end{bmatrix} .
$$

The product of the adjacency matrix. Each of the diagonal entries of $A^2$ equals the degree of the corresponding vertex, if the graph has no self loops.

We have all the results in case of usual graphs to be true in case of neutrosophic graphs.

The only advantage is we can several neutrosophic graphs but only one usual graph with the number of edges and vertices remaining fixed.

We will illustrate this situation by some examples.

***Example 4.43:*** Let G be the neutrosophic graph which is as follows:

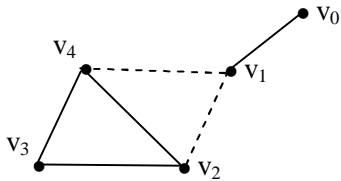

Figure 4.91

The adjacency matrix A of G is as follows:

$$A = \begin{array}{c} \\ v_0 \\ v_1 \\ v_2 \\ v_3 \\ v_4 \end{array} \begin{array}{c} \begin{array}{ccccc} v_0 & v_1 & v_2 & v_3 & v_4 \end{array} \\ \left[\begin{array}{ccccc} 0 & 1 & 0 & 0 & 0 \\ 1 & 0 & I & 0 & I \\ 0 & I & 0 & 1 & 1 \\ 0 & 0 & 1 & 0 & 1 \\ 0 & I & 1 & 1 & 0 \end{array}\right], \end{array}$$

$$A^2 = \begin{bmatrix} 1 & 0 & I & 0 & I \\ 0 & 1+2I & I & 2I & I \\ I & I & 2+I & 1 & 1+I \\ 0 & 2I & 1 & 2 & 1 \\ I & 1 & 1+I & 1 & 2+I \end{bmatrix}.$$

Clearly $v_0$ has only one edge passing through it.

The vertex $v_1$ has two neutrosophic edges and one usual edge that is why the second diagonal term is $1 + 2I$.

The vertex $v_2$ has 2 ordinary edges and one neutrosophic edge passing thro' it that is why the term is $2+I$.

The edges thro' the vertex are two ordinary edges.

Finally the edges thro' $v_4$ are two ordinary edges and one neutrosophic edge.

***Example 4.44:*** Let G be a neutrosophic graph given as follows:

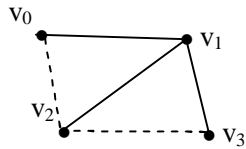

Figure 4.92

The adjacency neutrosophic matrix A associated with G is as follows:

$$\begin{array}{cccc} & v_0 & v_1 & v_2 & v_3 \end{array}$$

$$A = \begin{array}{c} v_0 \\ v_1 \\ v_2 \\ v_3 \end{array} \begin{bmatrix} 0 & 1 & I & 0 \\ 1 & 0 & 1 & 1 \\ I & 1 & 0 & I \\ 0 & 1 & I & 0 \end{bmatrix}.$$

We find $A^2$;

$$A^2 = \begin{bmatrix} 1+I & I & 1 & 1+I \\ I & 3 & 2I & I \\ 1 & 2I & 2I+1 & 1 \\ 1+I & I & 1 & 1+I \end{bmatrix}.$$

$$\text{Now } A^3 = \begin{bmatrix} 2I & 3+2I & 5I & 2I \\ 3+2I & 4I & 3+2I & 3+2I \\ 5I & 2I+3 & 4I & 5I \\ 2I & 3+2I & 5I & 2I \end{bmatrix}.$$

We see $Y = A + A^2 + A^3$

$$= \begin{bmatrix} 1+3I & 4+3I & 6I+1 & 1+3I \\ 4+3I & 3+4I & 4+4I & 4+3I \\ 1+6I & 4I+4 & 6I+1 & 6I+1 \\ 1+3I & 3I+4 & 6I+1 & 1+3I \end{bmatrix}.$$

No term in Y is zero proving the neutrosophic graph G is connected.

Now we proceed onto find the Y of a disconnected neutrosophic graph.

***Example 4.45:*** Let G be a neutrosophic graph which is as follows:

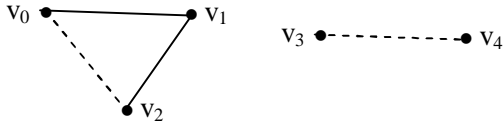

Figure 4.93

The neutrosophic adjacency matrix A of G is as follows:

$$A = \begin{array}{c} \\ v_0 \\ v_1 \\ v_2 \\ v_3 \\ v_4 \end{array} \begin{array}{c} v_0 \ v_1 \ v_2 \ v_3 \ v_4 \end{array} \begin{bmatrix} 0 & 1 & I & 0 & 0 \\ 1 & 0 & 1 & 0 & 0 \\ I & 1 & 0 & 0 & 0 \\ 0 & 0 & 0 & 0 & I \\ 0 & 0 & 0 & I & 0 \end{bmatrix}.$$

$$\text{Now } A^2 = \begin{bmatrix} 1+I & I & 1 & 0 & 0 \\ I & 2 & I & 0 & 0 \\ 1 & I & I+1 & 0 & 0 \\ 0 & 0 & 0 & I & 0 \\ 0 & 0 & 0 & 0 & I \end{bmatrix}.$$

We see from the diagonal elements of $A^2$ we see the vertex $v_0$ has one real edge and one neutrosophic edge.

The vertex $v_1$ two edges denoted by 2 is the second diagonal element.

The vertex $v_2$ has one real edge and one neutrosophic edge denoted by 1+I as the 3rd diagonal element of $A^2$.

For the vertices $v_3$ and $v_4$ we have the diagonal elements to be I confirming that only one neutrosophic edge bridges them.

We find $A^3$;

$$A^3 = \begin{bmatrix} 2I & 2+I & 3I & 0 & 0 \\ 2+I & 2I & I+2 & 0 & 0 \\ 3I & I+2 & 2I & 0 & 0 \\ 0 & 0 & 0 & 0 & I \\ 0 & 0 & 0 & I & 0 \end{bmatrix}$$

We find $A^4$;

$$A^4 = \begin{bmatrix} 2+4I & 5I & 3I+2 & 0 & 0 \\ 5I & 4+2I & 5I & 0 & 0 \\ 2+3I & 5I & 4I+2 & 0 & 0 \\ 0 & 0 & 0 & 0 & I \\ 0 & 0 & 0 & I & 0 \end{bmatrix}$$

Now $Y = A + A^2 + A^3 + A^4$

$$= \begin{bmatrix} 3+7I & 3+7I & 3+7I & 0 & 0 \\ 3+7I & 6+4I & 7I+3 & 0 & 0 \\ 3+7I & 3+7I & 3+7I & 0 & 0 \\ 0 & 0 & 0 & I & 3I \\ 0 & 0 & 0 & 3I & I \end{bmatrix}.$$

We see this neutrosophic matrix Y has zeros.

So the graphs are disjoint.

Now we just recall the following results as theorems and the proof of this is left as an exercise to the reader.

**THEOREM 4.12:** *Let G be a neutrosophic graph with n vertices. Let A be the neutrosophic adjacency matrix of the graph G.*

*Let $Y = A + A^2 + A^3 + \ldots + A^{n-1}$ then G is disconnected if and only if there exists at least one entry in the matrix Y that is zero.*

**THEOREM 4.13:** *Let G be a neutrosophic graph. A the neutrosophic adjacency matrix of G.*

*Then $A^n$ is also symmetric for any n.*

*That is to prove product of a symmetric matrix with itself is symmetric.*

*Now we introduce a few simple notions about neutrosophic graphs.*

*In the first place given the number of vertices and edges we can have several neutrosophic graphs which are different.*

For instance we have given two vertices there exists one and only one neutrosophic graph which is also pure that is

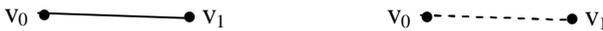

Figure 4.94

Given three vertices. We have the following neutrosophic graphs.

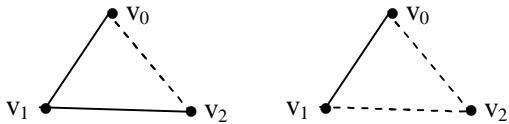

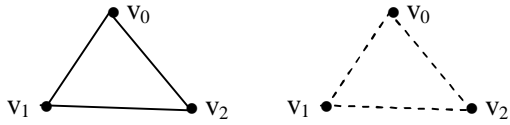

Figure 4.95

We see two neutrosophic graphs one pure neutrosophic graph and one usual graph.

It is an open problem to find for a given graph G with n vertices and p edges the total number of neutrosophic graphs with p edges and n vertices.

**Example 4.46:** Let G be a neutrosophic graph which is as follows:

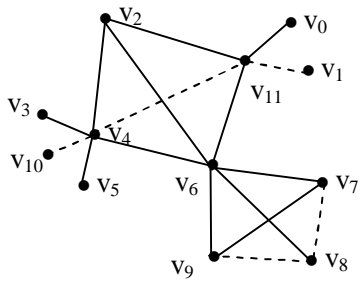

Figure 4.96

We have several subgraphs which are not neutrosophic sum subgraphs which are pure neutrosophic.

Some subgraphs which are neutrosophic.

We just give them.

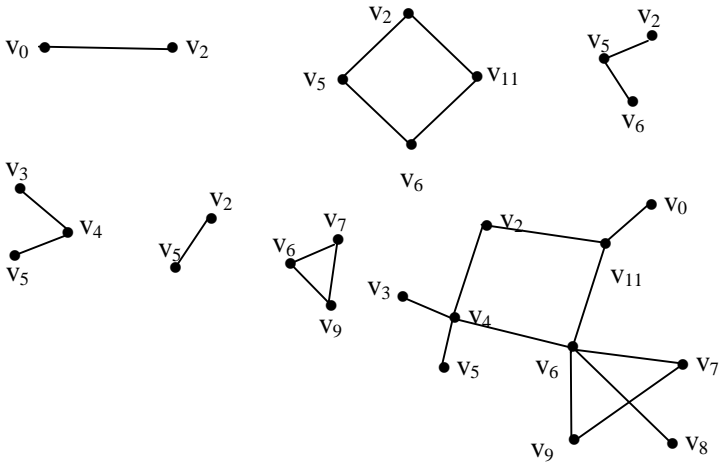

Figure 4.97

We see there is only one usual subgraphs which is large in the since all other usual subgraphs of G are only subgraphs of H we call this H as the largest usual subgraph of G.

Now we find all pure neutrosophic subgraphs of G.

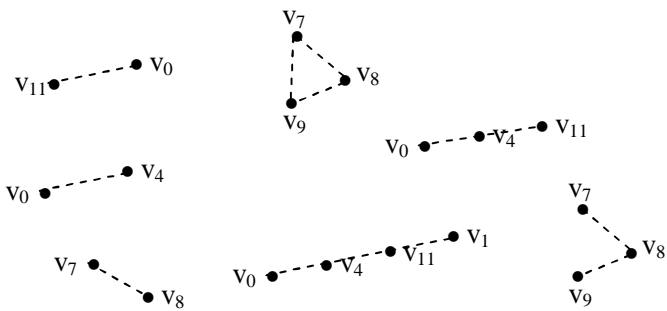

Figure 4.98

and so on.

We see P is the largest pure neutrosophic subgraph of G. All other pure neutrosophic subgraphs of G are subgraphs of P. Infact P is disjoint. However H is connected.

We call P the largest pure neutrosophic subgraph of G.

Consider the neutrosophic subgraphs of G.

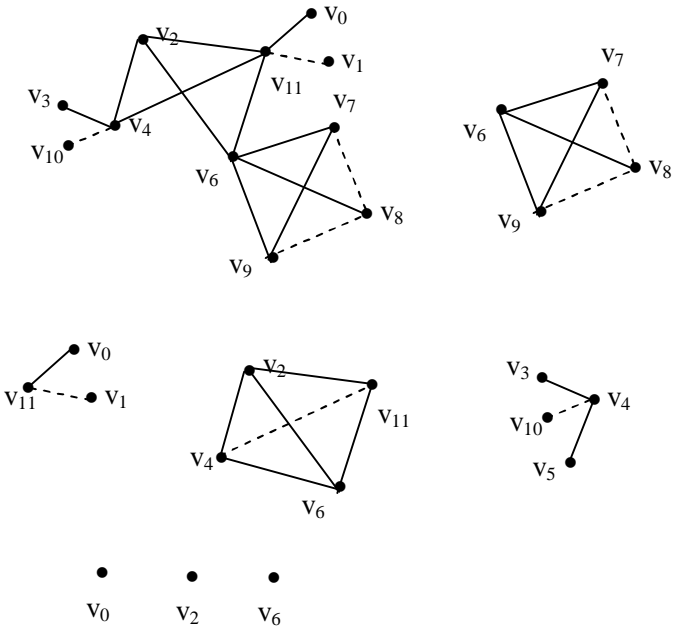

Figure 4.99

are all usual subgraphs of G.

The smallest pure neutrosophic graph is of the form •- - - - - - •

Figure 4.100

and the smallest neutrosophic graph is of the form

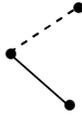

Figure 4.101

of course largest is only one in case of pure neutrosophic and usual graphs.

***Example 4.47:*** Let G be a neutrosophic graph given in the following.

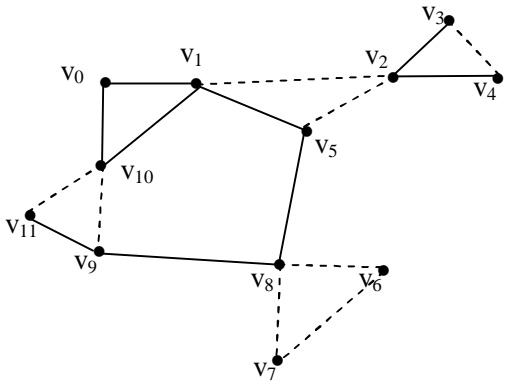

Figure 4.102

We see there are many small pure neutrosophic subgraphs.

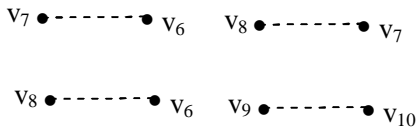

Figure 4.103

and so on.

There are many small neutrosophic graphs viz.

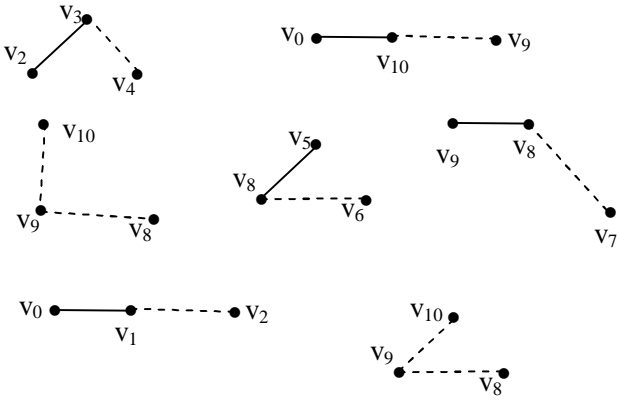

Figure 4.104

and so on.

Of course all vertices amount for small usual graphs.

Now we proceed onto present the largest pure neutrosophic graph and largest usual graph of G.

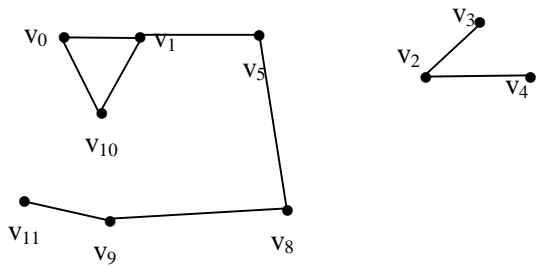

Figure 4.105

Clearly this subgraph is disjoint and it is the largest usual subgraph.

The largest pure neutrosophic subgraph is as follows:

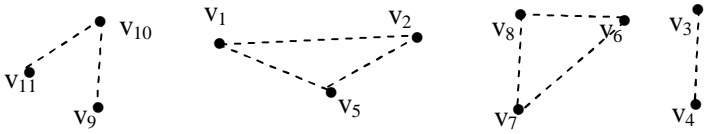

Figure 4.106

Clearly the largest pure neutrosophic subgraph is not connected and has four components. The usual largest subgraph is also not connected and has two components.

We define the following concepts.

If a connected neutrosophic graph G has both the largest pure neutrosophic subgraph as well the largest usual subgraph are disconnected we call G to be pseudo strongly disconnected largest neutrosophic graph.

If a connected neutrosophic graph G has both the largest pure neutrosophic subgraph as well as the largest usual subgraph are connected we call G to be strongly connected graph.

If in a connected graph G which is neutrosophic only one of the largest subgraph is connected other is disconnected then we call G to be just connected.

Suppose we have a neutrosophic graph G and it is a component of two graphs that is G is disconnected and one fo the subgraph is connected neutrosophic and other subgraph is connected usual then we call G to be special disconnected strong neutrosophic graph.

We will give examples of them.

***Example 4.48:*** Let G be a neutrosophic graph which is as follows:

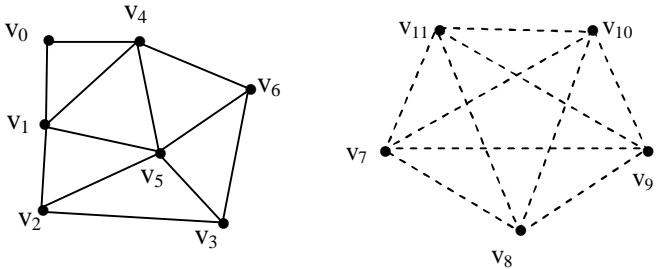

Figure 4.107

G is clearly a special disconnected strong neutrosophic graph.

Consider

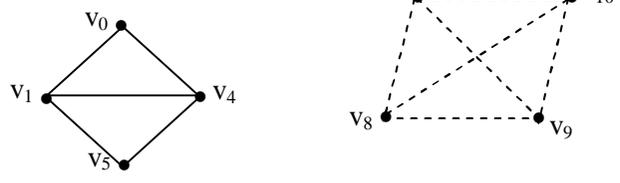

Figure 4.108

are usual subgraph of G and pure neutrosophic subgraph of G.

***Example 4.49:*** Let G be neutrosophic graph which is as follows:

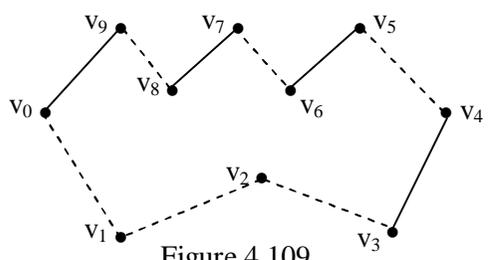

Figure 4.109

Clearly G is connected however both pure neutrosophic largest subgraph as well as largest usual subgraph of G are not connected.

***Example 4.50:*** Let G be a neutrosophic graph which is as follows:

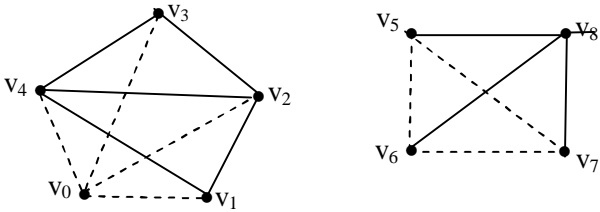

Figure 4.110

G is disconnected and both the largest pure neutrosophic subgraph as well as largest usual subgraph of G are disconnected.

**THEOREM 4.14:** *Let G be the disconnected neutrosophic graph with atleast $H_1$ and $H_2$ ($G = H_1 \cup H_2$) where both $H_1$ and $H_2$ are connected and neutrosophic. Then G has both largest pure neutrosophic subgraph as well as largest usual graph to be disconnected.*

Proof is left as an exercise to the reader.

**THEOREM 4.15:** *Let G be a neutrosophic graph. The smallest usual subgraphs of G are vertices and the smallest pure neutrosophic subgraphs of G are*

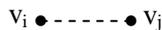

Figure 4.111

The proof is direct and hence is left as an exercise to the reader.

Find conditions on G so that the largest subgraphs are always disconnected.

Now we have spoken about also 3 types of subgraphs of a neutrosophic graph G.

***Example 4.51:*** Let G be a neutrosophic graph

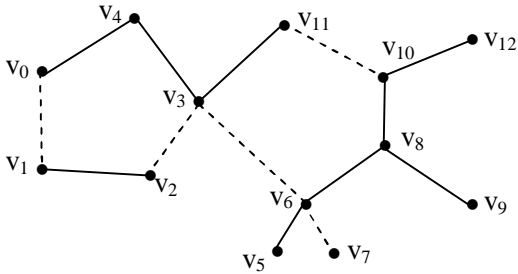

Figure 4.112

The subgraphs of G are

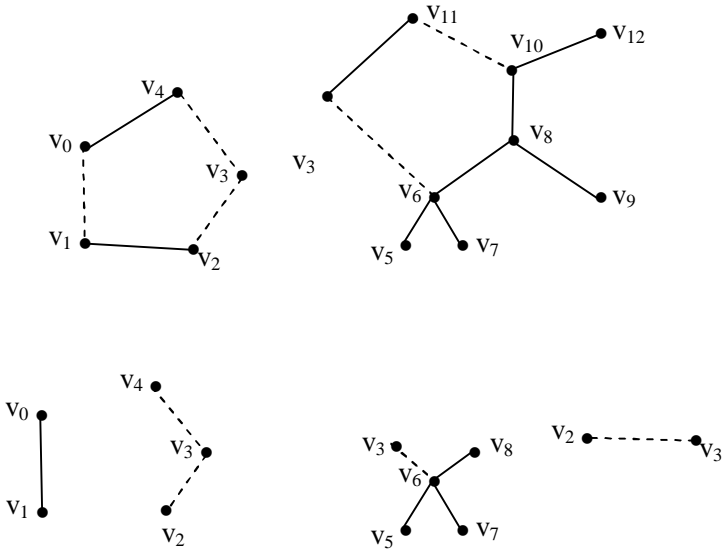

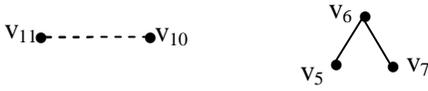

Figure 4.113

and so on.

We have seen subgraphs of neutrosophic graphs. The topological and lattice structure on these subgraphs.

We suggest some problems in the final chapter of this book.

**Chapter Five**

# SUGGESTED PROBLEMS

In this chapter we suggest the following problems for the interested reader.

Some of the problems are very difficult and some of them are at research level and some are open conjectures.

1. Give some examples of type I subset vertex graph.

2. Find the vertex subset graph of

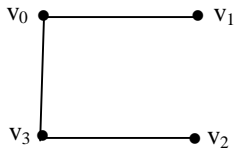

Figure 5.1

3. Let {G, V} be the graph, obtain the vertex subset graph of (G, V) given by the following figure.

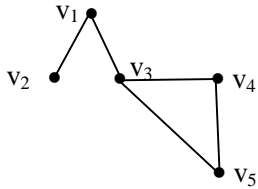

Figure 5.2

4. Let G be the graph with V = {$v_0$, $v_1$, $v_2$, $v_3$, $v_4$} as the vertex set.

   Find the number of type II graphs associated with G.

5. Distinguish between type I and type II graphs.

6. What are the advantages of using type II vertex subset graphs?

7. Is a type I vertex subset graph unique?

8. Given V = {$v_0$, $v_1$, $v_2$, $v_3$, $v_4$, $v_5$, $v_6$} as the vertex set.

   Find the number of type II vertex subset graphs associated with V.

9. Let G =

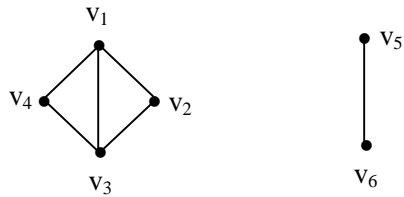

Figure 5.3

be the graph.

Find the type I vertex set graph associated with G.

10. Let

G =

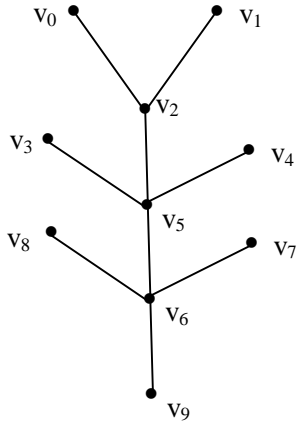

Figure 5.4

Is the associated type I subset graph of G a tree? Justify your claim.

11. Let $V = \{v_0, v_1, v_2, v_3, \ldots, v_{12}\}$ be the collection of vertices.

(i) How many type II vertex subset graphs associated with V are trees?

(ii) Can this be generalized for a set of n vertices?

(iii) How many of type II vertex subset graphs associated with V are complete graphs?

12. Study problems (11) for $V = \{v_1, v_2, \ldots, v_{24}\}$.

13. Find the type I vertex graph associated with figure 5.5.

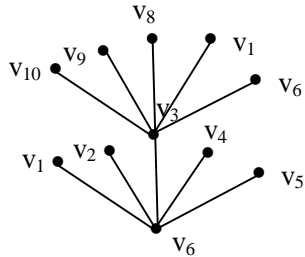

Figure 5.5

(i) Will that be a tree?
(ii) How many subgraphs of the type I vertex subset graph will be trees?
(iii) Find the number of subgraphs of type I vertex subset graphs which are complete graphs?

14. Given n-vertices as the graph G.
    (i) How many type II subset vertex graphs exist?
    (ii) How many will be trees?
    (iii) How many are n-any trees?

15. Give an example a strong neutrosophic graph with 5 vertices and 6 edges.

16. Is every subgraph of a strong neutrosophic graph G a strong neutrosophic graph?

17. Does there exist a strong neutrosophic graph all of whose subgraphs are strong neutrosophic?

18. Let G be a strong neutrosophic graph which is as follows.

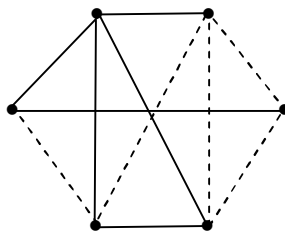

Figure 5.6

(i)  Find all subgraphs of G.
(ii)  How many subgraphs of G are strong neutrosophic?
(iii) How many subgraphs of G are neutrosophic?

19. Let G be a strong neutrosophic graph which is as follows:

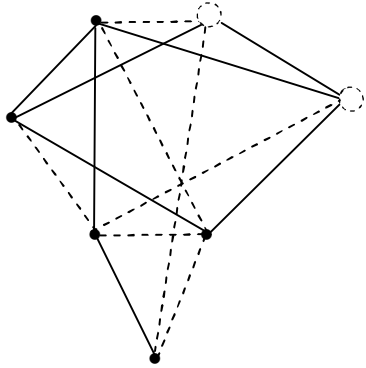

Figure 5.7

(i)  Can G have strong pure neutrosophic subgraphs?
(ii)  Can G have usual subgraphs?
(iii) Can G have neutrosophic subgraphs?
(iv) Find all subgraphs of G.

20. Does there exist a strong neutrosophic graph which has no neutrosophic subgraphs?

21. Give some interesting features enjoyed by the strong neutrosophic graphs.

22. Let G be the strong neutrosophic graph. S = {Collection of all subgraphs of G including φ and G}.

(i)  Prove S is closed under '∪' of subgraphs of G.
(ii) Prove S is closed under '∩' subgraphs of G.
(iii) Prove S is closed under difference of subgraphs.

23. Given a strong neutrosophic graph. If the number of neutrosophic vertices and usual vertices are given.

   (i)   Can one give the number of usual subgraphs of G?
   (ii)  Can we have the number of pure neutrosophic subgraphs of G?
   (iii) Find the number of neutrosophic subgraphs of G.
   (iv)  Find the total number of subgraphs of G.

24. Let G be a strong neutrosophic graph which is as follows.

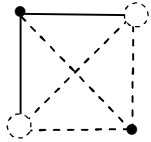

Figure 5.8

   (i)   Find all subgraphs of G.
   (ii)  Find all strong neutrosophic subgraphs of G.
   (iii) Find all usual subgraphs of G.
   (iv)  Find all neutrosophic subgraphs of G.
   (v)   Find all pure neutrosophic subgraphs of G.

25. Let $G_1$ be a strong neutrosophic graph which is as follows

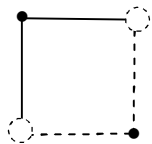

Figure 5.9

   (i)   Answer questions (i) to (v) of problems 24 for this graph.
   (ii)  Compare G of problem 24 and $G_1$.

26. Let $G_2$ be a strong neutrosophic graph given in the following figure 5.10.

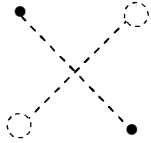

Figure 5.10

(i) Find all subgraphs of $G_2$
(ii) Compare $G_2$ with $G_1$ and G given in problems 24 and 25.

27. Let G be a strong neutrosophic graph which is as follows:

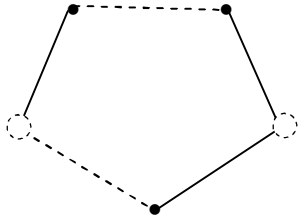

Figure 5.11

(i) Find all subgraphs of G.
(ii) Can G have pure neutrosophic subgraphs?
(iii) How many planar connected strong neutrosophic graphs exist with five vertices some of which are usual some neutrosophic.

28. Let G be a plane connected n vertices of which r are usual and n–r are neutrosophic and r edges are neutrosophic and n–r edges are ordinary.

   (i)  Find the number of usual subgraphs of G.
   (ii) Find the number of strong neutrosophic subgraphs of G.
   (iii) Find the number of neutrosophic subgraphs of G.

29. Let G be strong neutrosophic graph which is as follows:

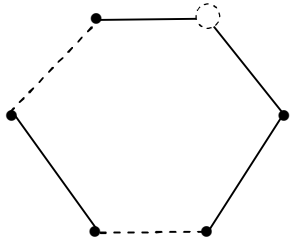

Figure 5.12

   (i)  Study question (i), (ii) and (iii) given in problem 26.

   (ii) Suppose $G_1$ is a strong neutrosophic graph which is as follows:

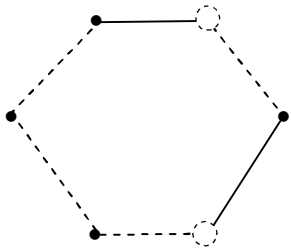

Figure 5.13

Compare G and $G_1$ and answer problem (i), (ii) and (iii) given in problem 26 in case of $G_1$.

30. Let G be a non planar strong neutrosophic graph which is as follows:

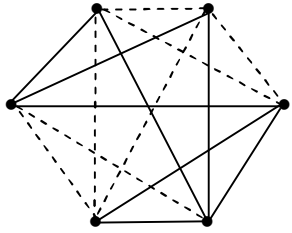

Figure 5.14

(i)   Find all subgraphs of G.
(ii)  Find the number of usual graphs of G.
(iii) Find the number of pure neutrosophic subgraphs of G.
(iv)  Find the number of strong neutrosophic subgraphs of G.

31. Let G be a complete connected strong neutrosophic non
    planar graph with n vertices.
    n = r + (n–r); r neutrosophic and n–r real and t usual edges
    rest neutrosophic edges.

    (i)   Find the number of strong neutrosophic subgraphs of G.
    (ii)  Find the number of usual subgraphs of G.
    (iii) Find the number of strong pure neutrosophic subgraphs
          of G.

32. Let G be a strong neutrosophic tree which is as follows:

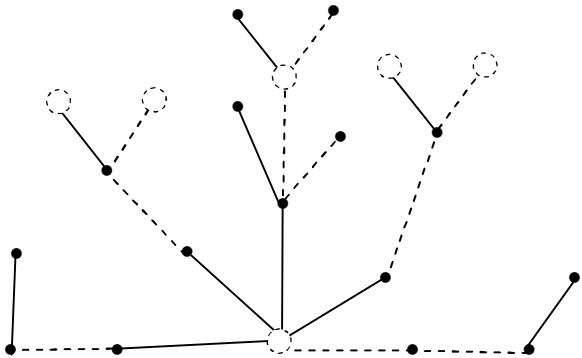

Figure 5.15

(i)   Find the total number of subgraphs of G.
(ii) Find the number of usual subgraphs of G.
(iii) How many subgraphs of G are strong neutrosophic
      subgraphs?
(iv)  Find the total number of neutrosophic subgraphs of G.

33. Let G be a strong neutrosophic graph which is as follows:

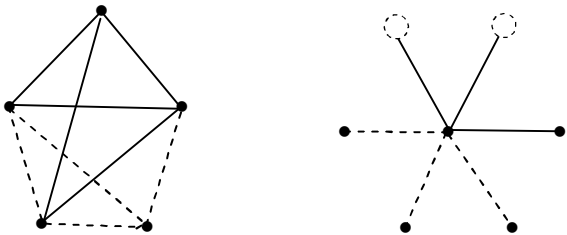

Figure 5.16

(i)   Find all subgraphs of G.
(ii)  Prove the collection of subgraphs is closed under ∪
      and ∩.

34. Give an example of a 6-partite strong neutrosophic graph.

35. Give an example of 10 partite neutrosophic graph G where
    G is a strong neutrosophic graph.

36. Let G be a strong neutrosophic planar connected graph with
    8 vertices (4 vertices are usual 4 neutrosophic) and 2 edges
    usual and the rest 6 edges neutrosophic.

(i)   Find all neutrosophic subgraphs of G.
(ii)  Find all strong neutrosophic subgraphs of G.
(iii) Find the number of usual subgraphs of G.

37. Find some interesting applications of k-partite neutrosophic strong graphs.

38. Let

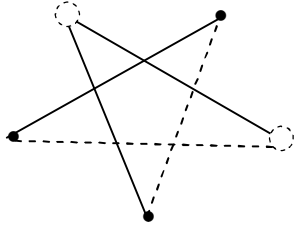

Figure 5.17
be a strong neutrosophic graph.

(i)  Find all subgraphs of G.
(ii)  How many subgraphs of G are strong neutrosophic?
(iii)  How many are usual subgraphs?

39. Let G be a strong neutrosophic graph which is as follows:

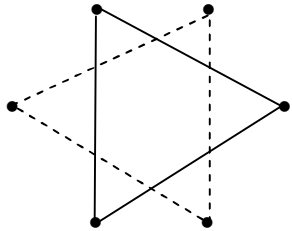

Figure 5.18

(i)  Find all subgraphs of G.

(ii)  Is it isomorphic with the strong neutrosophic graph.

H : 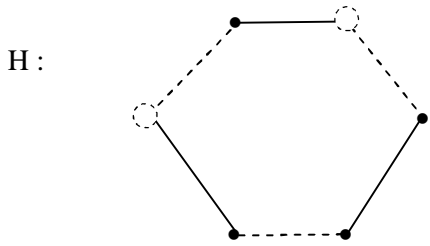

Figure 5.19

(iii) Find all usual subgraphs of G.
(iv) Find strong neutrosophic subgraphs of G.
(v) Is it possible to find pure neutrosophic subgraphs in G which is not strong?
(vi) Can H have pure neutrosophic subgraphs which are not strong?

40. Obtain all special features enjoyed by strong neutrosophic graph.

41. How are strong neutrosophic graphs different from neutrosophic graph?

    Explain this by examples.

42. Let G =

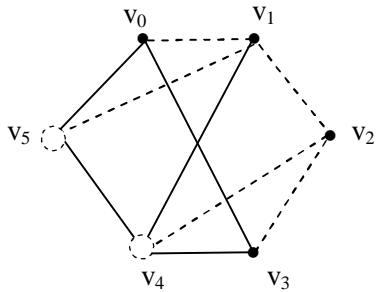

Figure 5.20
be the strong neutrosophic graph.

(i)  Find the complements of the graph G.
(ii) Obtain all subgraphs of G.
(iii) Which of the subgraphs of G ae strong neutrosophic?

43. Given the graph.

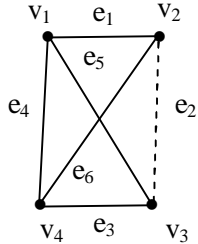

Figure 5.21

(i)  Find all neutrosophic graphs which can be obtained
     using the 4 vertices V = {$v_1$, $v_2$, $v_3$, $v_4$} and edges
     E = {$e_1$, $e_2$, …, $e_6$}.
(ii) Find all strong neutrosophic graphs obtained using E
     and V.
(iii) Find all pure neutrosophic graphs using E and V.

44. Find the proper application of strong neutrosophic graphs.

45. Find the number of 5 vertices and 9 edges strong
    neutrosophic graphs, neutrosophic graphs and super strong
    neutrosophic graphs.

46. Give examples of the four types of circuits.

47. Give example of a strong neutrosophic graphs.

48. Can a strong neutrosophic graph always contain a usual
    walk?

49. Find the power graph of G where

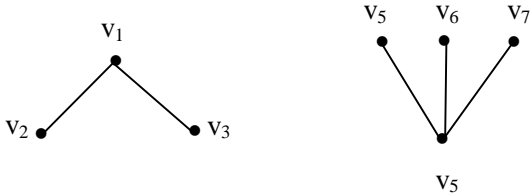

Figure 5.22

50. Let S(G) be the topological space of

G = { 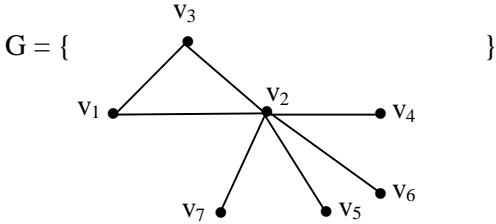 }

Figure 5.23

Study the properties associated with G in Figure 5.23.

51. What are the special features enjoyed by special subgraph topological spaces?

52. Find the power subgraphs of the neutrosophic graph.

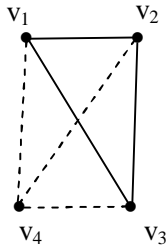

Figure 5.24

53. Can S(G) for any G has a subset which forms the Boolean algebra?

54. Can there be any S(G) which does not contain a subset collection which is a Boolean algebra?

55. Obtain all the special features enjoyed by Turan special topological subgraph spaces.

56. Prove the graph G

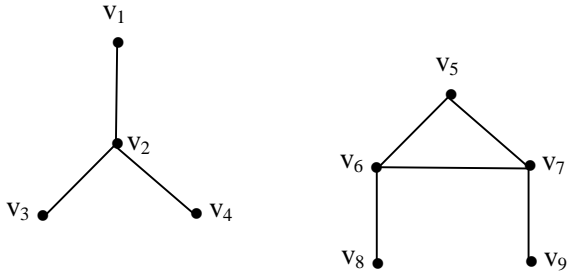

Figure 5.25

has the associated adjacency matrix to be symmetric super diagonal square matrix.

57. Can S(G) the power subgraph set have trees associated with it as substructures?

58. What are the advantages of using neutrosophic graphs in the place of usual graphs.

59. Let G =

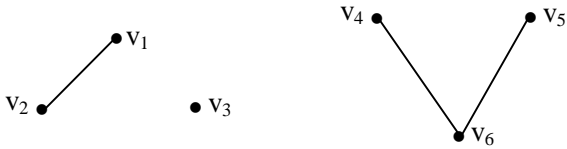

Figure 5.26

the graph.  Find S(G).

(i)   Can S(G) have trees as subgraphs?
(ii)  Can S(G) have substructure which are Boolean algebra?

60. Enumerate any other special feature associated with S(G) the collection of all subgraphs with φ.

61. Find the subgraphs of G which is as follows

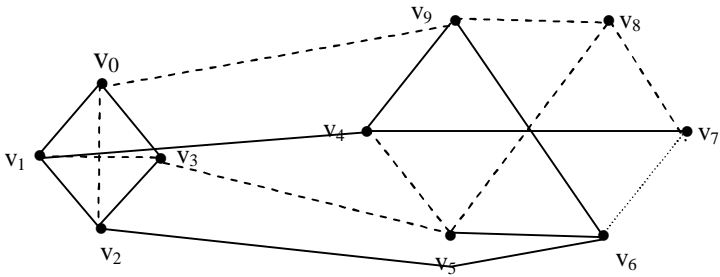

Figure 5.27

i.    Find S(G).
ii.   What is o(S(G))?
iii.  Prove S(G) is a special neutrosophic topological subgraph space.
iv.   Find at least three subspaces of this special neutrosophic topological space which are only special topological subspaces that are not neutrosophic.
v.    Find the adjacency matrix A of G.
vi.   Find A(G) the neutrosophic incidence matrix of G.

62. Let G be a neutrosophic graph which is as follows:

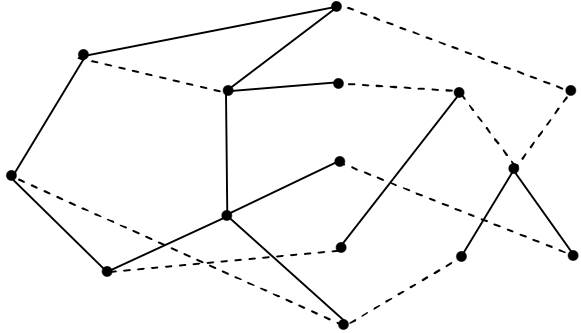

Figure 5.28

i.   Find all the pure neutrosophic subgraphs of G and show they form a special pure neutrosophic topological subgraphs space T.

ii.  Find all usual subgraphs special topological subgraph space $T_1$.

iii. Find all subgraphs S(G) of G and show S(G) is the special neutrosophic topological subgraph space S.

iv.  Show T and $T_1$ are subspaces of S.

63. Let G be the neutrosophic graph given in the following:

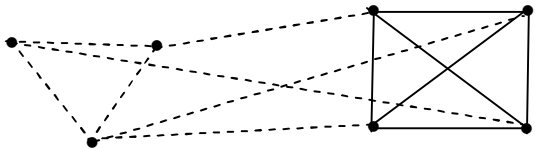

Figure 5.29

i.   Find S(G) = {The collection of all subgraphs of G including G and $\phi$}.

ii.  Find o(S(G)).

64. Let G′ be the neutrosophic graph.

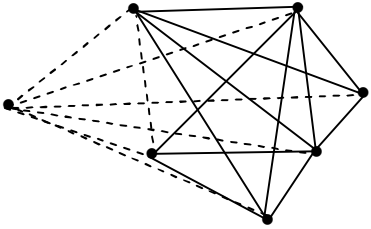

Figure 5.30

    i.   Find S(G′).
    ii.  Find o(S(G′)).
    iii. Compare S(G) of problem 36 with S(G′).

65. Let G be a neutrosophic graph.

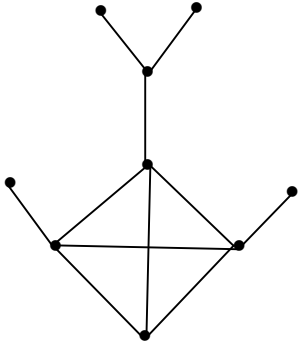

Figure 5.31

   (i)  Find S(G).

   (ii) Compare S(G) with S(H) where H is a neutrosophic
        graph which is as follows:

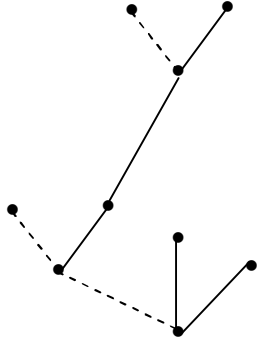

Figure 5.32

66. Let G be the graph.

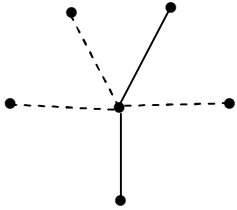

Figure 5.33

    i.   Find S(G).

   ii.   Find the special topological subgraph space associated with S(G).

  iii.   Find at least 5 special subtopological subgraph spaces associated with S(G) such that at least one of them is or special subtopological pure neutrosophic subgraph subspace of S(G).

67. Let $G_1$, $G_2$, $G_3$ and $G_4$ be four neutrosophic graphs given in the following:

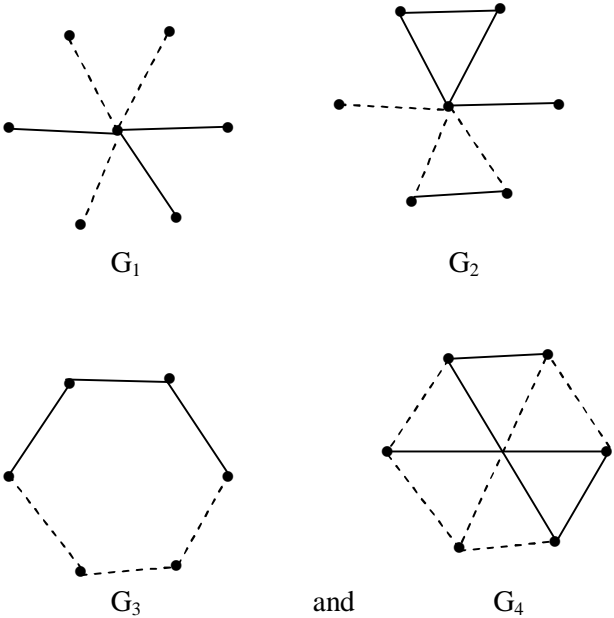

Figure 5.34

i.   Compare $S(G_1)$, $S(G_2)$, $S(G_3)$ and $S(G_4)$.
ii.  Find the special neutrosophic topological subgraph spaces and compare them.
iii. Find the special lattice of subgraphs of $S(G_1)$, $S(G_2)$, $S(G_3)$ and $S(G_4)$.

68. Let G be the neutrosophic graph which is as follows:

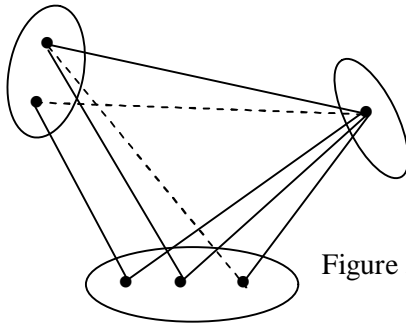

Figure 5.35

i.   Find the complement of G.
ii.  Find S(G).
iii. Find o(S(G)).
iv.  Find the special topological neutrosophic subgraph space associated with S(G).

69. Let G be a neutrosophic graph which is as follows:

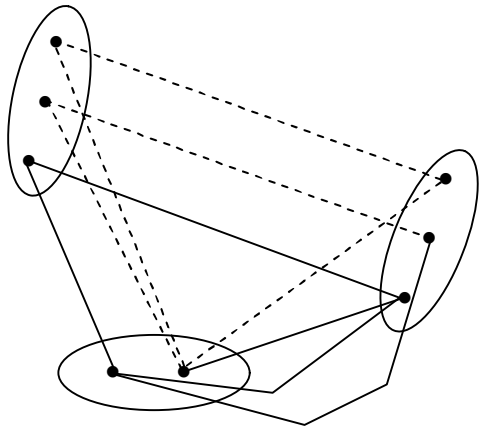

Figure 5.36

i.   Find the complement G′ of G.
ii.  Compare S(G′) and S(G).
iii. Which of the collections S(G) and S(G′) has more number of pure neutrosophic subgraphs?
iv.  Find the special topological neutrosophic subgraph spaces of S(G) and S(G′).
v.   What of the topological spaces have more number of neutrosophic topological vector subspaces.

70. Let G be a neutrosophic graph which is as follows:

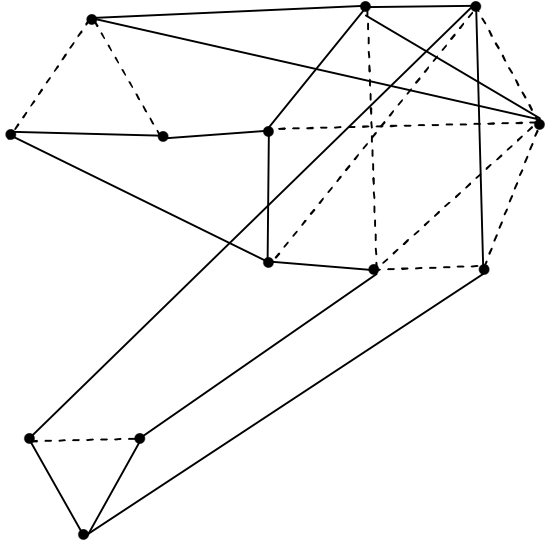

Figure 5.37

    i.   Find the complement of G.
    ii.   Find S(G).
    iii.   Find o(S(G)).

71. Let G be the neutrosophic graph which is as follows:

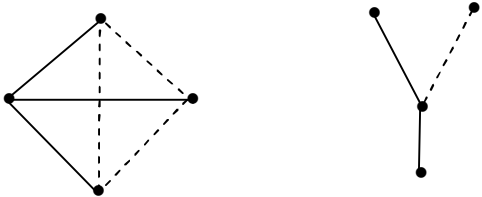

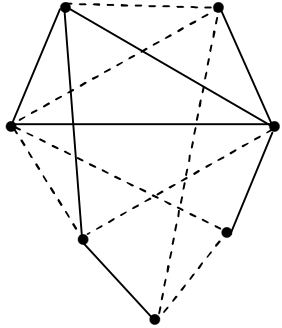

Figure 5.40

    i.   Find S(G).
    ii.  What is o(S(G))?
    iii. Find the special topological neutrosophic subgraphs of G.

72. Let G be a neutrosophic graph which is as follows:

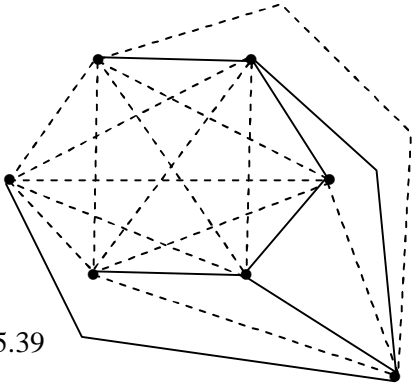

Figure 5.39

    i.   Find S(G).
    ii.  Find o(S(G)).
    iii. Does S(G) contain more number of neutrosophic subgraphs than usual subgraphs?

73. Find a characterization or method of finding neutrosophic self complemented graphs.

74. Give at least 5 examples of self complemented bipartite neutrosophic graph.

75. Obtain some interesting results about neutrosophic self complemented graph.

76. Find some interesting features enjoyed by the special lattice neutrosophic subgraphs of a neutrosophic subgraph.

77. Let G be a neutrosophic graph which is as follows:

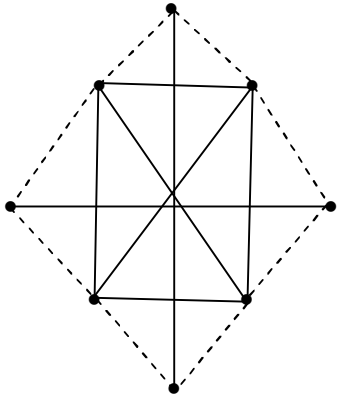

Figure 5.40

    i.   Find the special lattice neutrosophic subgraph of G.
    ii.  Find S(G) and o(S(G)).

78. Let G be a neutrosophic with 10 vertices.
    i.   Find the total number of neutrosophic graphs with 10 vertices.
    ii.  How many graphs with 10 vertices are self complemented?
    iii. How many of these neutrosophic 10 vertices graphs have equal number of neutrosophic subgraphs?

79. Let G and H be two neutrosophic graphs which are as follows:

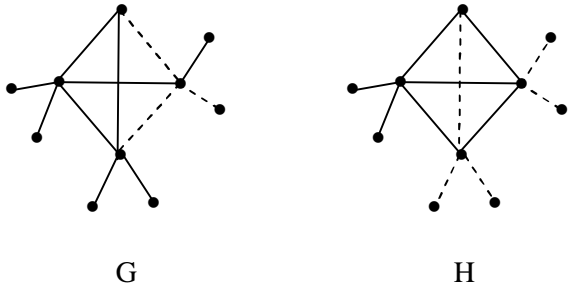

G          H

Figure 5.41

i.   Find S(H) and S(G).
ii.  Prove S(H) has more number of neutrosophic subgraphs
     than that of S(G).
iii. Find the special neutrosophic lattice subgraphs of both
     S(G) and S(H).

80. Let G be the neutrosophic graph which is as follows:

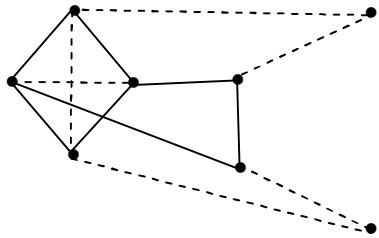

Figure 5.42

i.   Find the adjacency neutrosophic matrices A of G.

ii.  Find the incidence neutrosophic matrix A(G) of G.

81. Let G be a neutrosophic graph which is as follows:

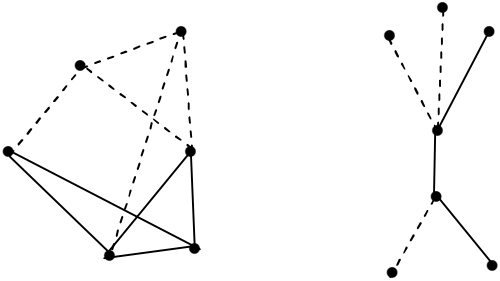

Figure 5.43

i. Find the adjacency matrix A of G. Is A   a super neutrosophic square diagonal matrix.

ii. Show $A + A^2 + \ldots + A^{12} = Y$ is such that Y has zeros.

iii. Find A(G) of G.

82. Describe a strongly connected neutrosophic graph.

83. Let G be a neutrosophic graph which is as follows:

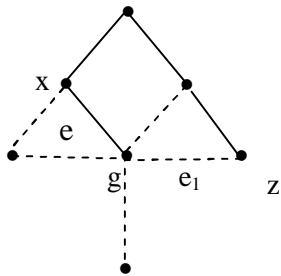

Figure 5.44

(i) Find the contracted graph G/e xy = e and find the contracted graph $G/e_1$ yz = $e_1$. Compare S(G/e) and S(G/$e_1$).

(ii) Which of them contain more number of neutrosophic subgraphs?

84. Let G be the neutrosophic graph which is as follows:

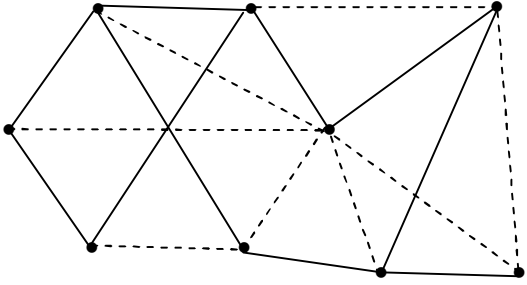

Figure 5.45

(i) Find the adjacency matrix A of G.
(ii) Prove in $A^2$ the diagonal elements correspond to the degree of the vertices or it gives the number of usual edges and neutrosophic edges at each of the vertices.

85. How many colors are needed to edge color the neutrosophic graph?

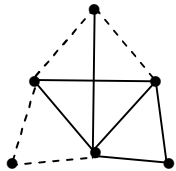

Figure 5.46

86. Find the number of colors needed to edge color the neutrosophic graph G which is as follows:

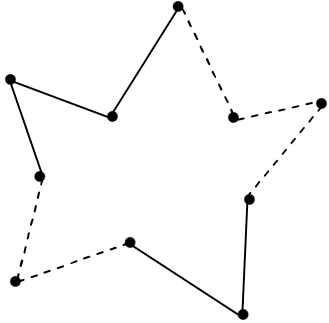

Figure 5.47

# FURTHER READING

# INDEX

## P

## Q

## S

# ABOUT THE AUTHORS

**Dr.W.B.Vasantha Kandasamy** is a Professor in the Department of Mathematics, Indian Institute of Technology Madras, Chennai. In the past decade she has guided 13 Ph.D. scholars in the different fields of non-associative algebras, algebraic coding theory, transportation theory, fuzzy groups, and applications of fuzzy theory of the problems faced in chemical industries and cement industries. She has to her credit 694 research papers. She has guided over 100 M.Sc. and M.Tech. projects. She has worked in collaboration projects with the Indian Space Research Organization and with the Tamil Nadu State AIDS Control Society. She is presently working on a research project funded by the Board of Research in Nuclear Sciences, Government of India. This is her 120[th] book.

On India's 60th Independence Day, Dr.Vasantha was conferred the Kalpana Chawla Award for Courage and Daring Enterprise by the State Government of Tamil Nadu in recognition of her sustained fight for social justice in the Indian Institute of Technology (IIT) Madras and for her contribution to mathematics. The award, instituted in the memory of Indian-American astronaut Kalpana Chawla who died aboard Space Shuttle Columbia, carried a cash prize of five lakh rupees (the highest prize-money for any Indian award) and a gold medal.
She can be contacted at vasanthakandasamy@gmail.com
Web Site: http://mat.iitm.ac.in/home/wbv/public_html/
or http://www.vasantha.in

---

**Dr. K. Ilanthenral** is Assistant Professor in the School of Computer Science and Engg, VIT University, India. She can be contacted at ilanthenral@gmail.com

---

**Dr. Florentin Smarandache** is a Professor of Mathematics at the University of New Mexico in USA. He published over 75 books and 200 articles and notes in mathematics, physics, philosophy, psychology, rebus, literature. In mathematics his research is in number theory, non-Euclidean geometry, synthetic geometry, algebraic structures, statistics, neutrosophic logic and set (generalizations of fuzzy logic and set respectively), neutrosophic probability (generalization of classical and imprecise probability). Also, small contributions to nuclear and particle physics, information fusion, neutrosophy (a generalization of dialectics), law of sensations and stimuli, etc. He got the 2010 Telesio-Galilei Academy of Science Gold Medal, Adjunct Professor (equivalent to Doctor Honoris Causa) of Beijing Jiaotong University in 2011, and 2011 Romanian Academy Award for Technical Science (the highest in the country). Dr. W. B. Vasantha Kandasamy and Dr. Florentin Smarandache got the 2012 New Mexico-Arizona and 2011 New Mexico Book Award for Algebraic Structures. He can be contacted at smarand@unm.edu